%% file: main.tex
\documentclass{zlab}

\usepackage[T1]{fontenc}
\usepackage{lmodern}

\usepackage{xspace}
\usepackage{graphicx,amsmath,amssymb,hyperref}
\usepackage[authoryear,round]{natbib}
\usepackage{xcolor}
\usepackage{etoc}
\usepackage{wrapfig}
\usepackage{listings}
\usepackage[most]{tcolorbox}
\hypersetup{
  colorlinks=true,
  urlcolor=magenta,
  linkcolor=black,
  citecolor=link,
  filecolor=black,
  pdfborder={0 0 0}
}

\definecolor{link}{HTML}{307ef1}
\providecommand{\equationname}{Equation}
\providecommand{\sectionname}{\S}

\newcommand{\figref}[1]{%
  \figurename~\hyperref[#1]{\textcolor{link}{\ref*{#1}}}%
}
\newcommand{\tabref}[1]{%
  \tablename~\hyperref[#1]{\textcolor{link}{\ref*{#1}}}%
}
\newcommand{\eqrefc}[1]{%
  \equationname~\hyperref[#1]{\textcolor{link}{\ref*{#1}}}%
}
\newcommand{\secrefc}[1]{%
  \sectionname\hyperref[#1]{\textcolor{link}{\ref*{#1}}}%
}

\makeatletter
\fancypagestyle{firststyle}{
  \fancyhead[L]{}
  \fancyhead[C]{}
  \fancyhead[R]{}
  \fancyfoot[L]{\footerfont \the\correspondingauthor}
  \fancyfoot[C]{\footerfont\raisebox{7.6pt}[0pt][0pt]{\thepage}}
  \fancyfoot[R]{}
}
\makeatother

\makeatletter
\DeclareRobustCommand\bfseries{%
  \not@math@alphabet\bfseries\mathbf
  \fontseries\bfdefault\selectfont
  \sffamily
}
\makeatother
\DeclareTextFontCommand{\textbf}{\bfseries\sffamily}
\usepackage{subcaption}
\definecolor{roleHeader}{HTML}{F1F4F8}
\definecolor{swarmHeader}{HTML}{F4F0FA}
\definecolor{promptHeader}{HTML}{EFF6EF}
\definecolor{evalHeader}{HTML}{FBF1E7}
\definecolor{computeHeader}{HTML}{fceef1}
\definecolor{envHeader}{HTML}{f8f5fe}
\definecolor{qualHeader}{HTML}{eef4fc}
\definecolor{threeUseCard}{HTML}{F4F5F6}   % three-uses figure cards (teaser panel grey)
\definecolor{datasetHeader}{HTML}{f6f2ff}
\definecolor{roleReproducer}{HTML}{EEF5FF}
\definecolor{roleDesigner}{HTML}{EEF8F0}
\definecolor{rolePackager}{HTML}{FFF7E8}
\definecolor{roleVerifier}{HTML}{F8EEF7}
\definecolor{milestoneNotPossible}{HTML}{FCECEC}
\definecolor{milestoneNone}{HTML}{F2F4F7}
\definecolor{milestoneMethod}{HTML}{EEF5FF}
\definecolor{milestoneCore}{HTML}{EFF9F1}
\definecolor{milestoneCorePlus}{HTML}{E7F6EA}
\definecolor{milestoneSecondary}{HTML}{DCEFE1}
\definecolor{milestoneMajority}{HTML}{D1EBD6}
\definecolor{milestoneNear}{HTML}{C5E6CB}
\definecolor{milestoneFull}{HTML}{B9E0BF}

\definecolor{LightGray}{HTML}{F4F5F7}
\definecolor{dcBlue}{HTML}{3A86FF}
\definecolor{dcGreen}{HTML}{2DC48D}
\definecolor{dcPurple}{HTML}{8338EC}
\definecolor{dcAmber}{HTML}{FFBE0B}
\definecolor{dcTeal}{HTML}{FB5607}
\definecolor{dcRose}{HTML}{FF006E}
\definecolor{dcIndigo}{HTML}{4B5C8A}

\definecolor{axScalable}{HTML}{2D64F0}
\definecolor{axEvolving}{HTML}{E5308C}
\definecolor{axExecution}{HTML}{1FA88B}
\newcommand{\prop}[2]{{\color{#1}\sffamily\bfseries\fontsize{9.3}{12}\selectfont\textls[-20]{#2}}}
\newtcolorbox{designbox}[1]{%
  before skip=16pt, after skip=8pt,
  enhanced jigsaw,
  colback=#1!1,
  colframe=#1!70,
  drop fuzzy shadow southeast={fill=#1!10},
  boxrule=0.9pt,
  boxsep=0.1pt,
  left=10pt, right=10pt, top=5pt, bottom=4pt,
}

\definecolor{ymlBg}{HTML}{F5F6F8}
\definecolor{ymlText}{HTML}{3B4252}
\definecolor{ymlKey}{HTML}{2E5A88}
\definecolor{ymlLit}{HTML}{A8617A}
\definecolor{ymlStr}{HTML}{5E8B6A}
\lstdefinestyle{yaml}{
  basicstyle=\small\ttfamily\color{ymlText},
  backgroundcolor=\color{ymlBg},
  frame=single,
  framerule=0pt,
  rulecolor=\color{ymlBg},
  framesep=10pt,
  xleftmargin=14pt,
  xrightmargin=14pt,
  aboveskip=10pt,
  belowskip=2pt,
  showstringspaces=false,
  keywords=[1]{action,gpus,test,milestone,notes},
  keywordstyle=[1]\color{ymlKey},
  keywords=[2]{submit,done,true,false},
  keywordstyle=[2]\color{ymlLit},
  morestring=[b]",
  stringstyle=\color{ymlStr},
}

\usepackage{ebgaramond}               % EB Garamond (Type1) for prompt text
\usepackage{array,booktabs,tabularx}  % tables inside prompt boxes
\providecommand{\tightlist}{\setlength{\itemsep}{0pt}\setlength{\parskip}{0pt}}

\definecolor{promptC1}{HTML}{FE5353}\definecolor{promptC2}{HTML}{FF9E62}
\definecolor{promptC3}{HTML}{FFBF60}\definecolor{promptC4}{HTML}{8DDD6B}
\definecolor{promptC5}{HTML}{28C879}\definecolor{promptC6}{HTML}{4BBCFF}
\definecolor{promptC7}{HTML}{9864FF}\definecolor{promptC8}{HTML}{8088FE}
\lstdefinestyle{promptcode}{%
  basicstyle=\scriptsize\ttfamily, breaklines=true,
  breakatwhitespace=false, columns=flexible, keepspaces=true,
  breakindent=0pt, postbreak=\mbox{\textcolor{gray}{$\hookrightarrow$}\space},
  aboveskip=4pt, belowskip=2pt, xleftmargin=2pt}
\newtcolorbox{promptbox}[2]{%
  enhanced, breakable,
  colback=white, colframe=#1,
  boxrule=0.9pt, arc=6pt, outer arc=6pt,
  left=14pt, right=14pt, top=13pt, bottom=12pt,
  fonttitle=\sffamily\bfseries\normalsize, coltitle=black,
  attach boxed title to top left={xshift=10pt, yshift=-\tcboxedtitleheight/2},
  boxed title style={colback=white, frame hidden, boxrule=0pt, boxsep=3pt},
  before upper={\colorlet{promptsec}{#1!85!black}\ebgaramond\small
    \setlength{\emergencystretch}{2em}},
  title={#2},
}
\newcommand{\psec}[1]{\par\smallskip{\color{promptsec}\ebgaramond\normalsize #1}\par\nopagebreak\smallskip}
\newcommand{\psecsub}[1]{\par\smallskip{\color{promptsec}\ebgaramond\bfseries #1}\par\nopagebreak\smallskip}
\newcommand{\promptellipsis}{\par\smallskip\centerline{\color{gray}\small$[\,\cdots\,]$}\par\smallskip}

\tcbuselibrary{listings}
\definecolor{fileBody}{HTML}{FCFDFE}
\definecolor{codeKw}{HTML}{7C3AED}\definecolor{codeStr}{HTML}{0FA968}\definecolor{codeCom}{HTML}{99A1AD}
\definecolor{accBlue}{HTML}{3A86FF}\definecolor{accPurple}{HTML}{8B5CF6}
\definecolor{accTeal}{HTML}{12B5A5}\definecolor{accAmber}{HTML}{F08C00}
\definecolor{accGreen}{HTML}{1FAF6B}\definecolor{accRose}{HTML}{EF476F}
\definecolor{accIndigo}{HTML}{6366F1}
\lstdefinestyle{filecode}{%
  basicstyle=\scriptsize\ttfamily\color{ymlText},
  breaklines=true, breakatwhitespace=false, columns=fixed,
  breakindent=0pt, showstringspaces=false,
  postbreak=\mbox{\textcolor{codeCom}{$\hookrightarrow$}\space},
  keywords={class,def,return,for,if,in,import,from,as,super,self,None,True,False,not,is,and,or,with,while,elif,else},
  keywordstyle=\color{codeKw}\bfseries,
  comment=[l]{\#}, commentstyle=\color{codeCom}\itshape,
  morestring=[b]", stringstyle=\color{codeStr}}
\lstdefinestyle{filetext}{%
  basicstyle=\scriptsize\ttfamily\color{ymlText},
  breaklines=true, breakatwhitespace=false, columns=fixed,
  breakindent=0pt, showstringspaces=false,
  postbreak=\mbox{\textcolor{codeCom}{$\hookrightarrow$}\space},
  comment=[l]{\#}, commentstyle=\color{codeKw}\bfseries}
\newtcblisting{filebox}[2]{%
  enhanced, breakable, listing only, listing style=filecode,
  colback=fileBody, colframe=#1, boxrule=0.8pt, arc=3pt, outer arc=3pt,
  top=3pt, bottom=4pt, left=8pt, right=6pt,
  title={\,#2}, fonttitle=\ttfamily\footnotesize\bfseries, coltitle=#1!72!black,
  colbacktitle=#1!13, toptitle=2pt, bottomtitle=2pt, titlerule=0pt,
  before skip=6pt, after skip=8pt}
\newtcblisting{mdbox}[2]{%
  enhanced, breakable, listing only, listing style=filetext,
  colback=fileBody, colframe=#1, boxrule=0.8pt, arc=3pt, outer arc=3pt,
  top=3pt, bottom=4pt, left=8pt, right=6pt,
  title={\,#2}, fonttitle=\ttfamily\footnotesize\bfseries, coltitle=#1!72!black,
  colbacktitle=#1!13, toptitle=2pt, bottomtitle=2pt, titlerule=0pt,
  before skip=6pt, after skip=8pt}

\newlength{\softTableWidth}
\newlength{\softTableContentWidth}
\newlength{\softTableBodyHeight}
\newcommand{\softTableRoleCell}[1]{%
  \begin{minipage}[c][\softTableBodyHeight][c]{\linewidth}\centering #1\end{minipage}%
}
\newcommand{\softTableTextCell}[1]{%
  \begin{minipage}[c][\softTableBodyHeight][c]{\linewidth}\raggedright #1\end{minipage}%
}
\newenvironment{softcolortable}[1][1]{%
  \begingroup
  \small
  \setlength{\softTableWidth}{#1\linewidth}%
  \setlength{\tabcolsep}{3pt}%
  \setlength{\softTableContentWidth}{\dimexpr\softTableWidth-6\tabcolsep\relax}%
  \setlength{\extrarowheight}{0pt}%
  \renewcommand{\arraystretch}{1.22}%
  \setlength{\aboverulesep}{0pt}%
  \setlength{\belowrulesep}{0pt}%
}{%
  \endgroup
}
\definecolor{swSlate}{HTML}{5582B5}
\definecolor{swTeal}{HTML}{66A0B5}
\definecolor{swSage}{HTML}{76B098}
\definecolor{swOchre}{HTML}{B5A07A}
\definecolor{swMauve}{HTML}{B585A5}

\newcommand{\swcolor}{%
  \includegraphics[height=1.7\baselineskip]{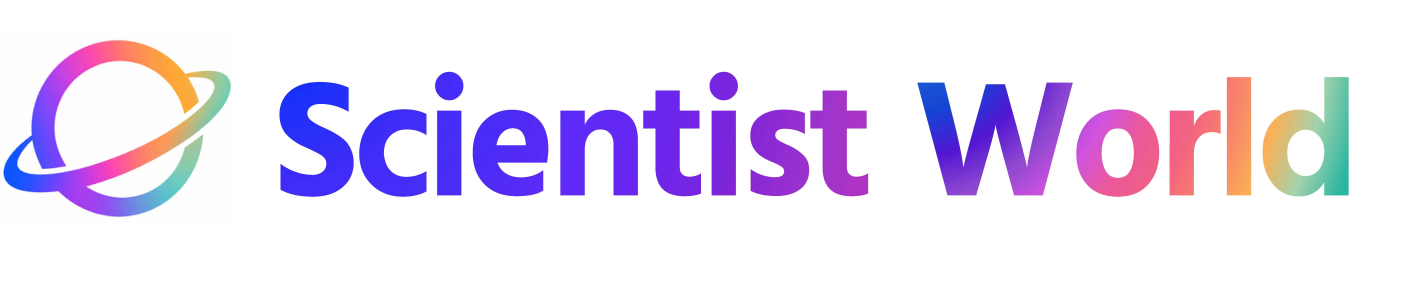}%
}
\newcommand{\sw}{\raisebox{0pt}{\includegraphics[width=5.757em]{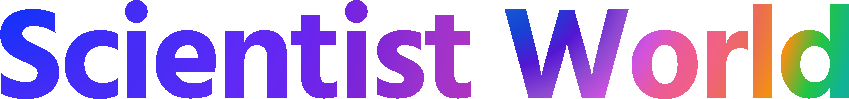}}\xspace}
\newcommand{\swplain}{Scientist World\xspace}

\newcommand{\builder}{builder\xspace}
\newcommand{\builders}{builders\xspace}
\newcommand{\Builders}{Builders\xspace}

\newcommand{\aisci}{AI-Scientist\xspace}
\newcommand{\aiscis}{AI-Scientists\xspace}

\newcommand{\metric}{FNI\xspace}
\newcommand{\metricfull}{Frontier-Normalized Improvement\xspace}

\newcommand{\model}[1]{{\sffamily\fontsize{9.3}{12}\selectfont\textls[-20]{#1}}}

\def\siteAt{1}
\newcommand{\sitelink}{\href{https://ai-scientist.world}{\textcolor{link}{ai-scientist.world}}}

\title{%
  \centering
  \swcolor\\[-7pt]
  {\fontsize{18}{24}\selectfont\color{black!80}%
  Grow a Research World for AI-Scientists}%
  \ifnum\siteAt=5 \\[2pt]{\large\sffamily\sitelink}\fi
}

\newcommand{\namesemibold}[1]{\pdfliteral direct{2 Tr 0.35 w}#1\pdfliteral direct{0 Tr}}

\author{
    \vspace{0.095cm}
    \parbox{\textwidth}{\centering
        \fontsize{12.5}{17}\usefont{T1}{lmss}{m}{n}%    LM Sans + synthetic semibold  (current)
        \namesemibold{Taiming Lu \quad Sachin Konan \quad Zhuang Liu}
    }
    \\
    \vspace{-.28cm}
    {\fontsize{11.5}{17}\usefont{T1}{lmss}{m}{n} Princeton University\ifnum\siteAt=2 \quad\textperiodcentered\quad\sitelink\fi}
    \ifnum\siteAt=1 \\ \vspace{0.820cm} {\Large\sffamily\sitelink}\fi  % link centred between the affiliation and the teaser title (26pt each side); 0.965cm = 30pt above, 22pt below
}

\makeatletter
\renewcommand{\abscontent}{}%
\makeatother

\newenvironment{abstractblock}{%
  {\centering\large\bfseries Abstract\par}
  \vspace{-0.56em}
  \begin{list}{}{%
      \setlength{\leftmargin}{2em}
      \setlength{\rightmargin}{2em}
      \setlength{\topsep}{0pt}
      \setlength{\parsep}{0pt}
  }
  \item[]
}{%
  \end{list}
  \par\normalfont\vspace{1em}
}
\begin{document}

\begingroup
\makeatletter
\let\raggedright\centering
\makeatother

\vspace*{-0.69cm}
\maketitle
\endgroup

% Commented out: contribution footnote block (frees first-page space for the abstract)
% \setlength{\skip\footins}{1cm}
% \begingroup\renewcommand\thefootnote{}\footnotetext{\fontsize{9}{11}\selectfont\begin{tabular}[t]{@{}l@{}}$^*$\,Equal Contribution \\ $^\dag$\,Corresponding Author\end{tabular}}\addtocounter{footnote}{-1}\endgroup

% ============================================
% Links table (comment out if not needed)
% ============================================
% --- Links block (moved below teaser; see OPTION 1 in the layout section) ---
\newcommand{\linksblock}{%
\begin{center}
    \vspace{-0.8cm}
    {\Large\sffamily\href{https://ai-scientist.world}{\textcolor{link}{ai-scientist.world}}}
    % {\Large\sffamily\href{https://ai-scientist.world}{\textcolor{link}{www.ai-scientist.world}}}
    \vspace{0.6cm}
    % \vspace{0.2em}
    % \github \href{https://github.com/zlab-princeton/ScientistWorld}{\textcolor{link}{\textbf{Code}}} \quad
    % \hf \href{https://github.com/orgs/ScientistWorld}{\textcolor{link}{\textbf{Data}}}
\end{center}
}

% --- Alternative: table with icons, each link on its own line ---
% \vspace{-0.3cm}
% \begin{center}
%     \renewcommand{\arraystretch}{1.2}
%     \begin{tabular}{rll}
%         \worldwideweb & \textbf{Website} & \url{https://www.test.com/}\\
%         \github & \textbf{Code} & \url{https://github.com/test}\\
%         \hf & \textbf{Data} & \url{https://huggingface.com}
%     \end{tabular}
% \end{center}

% --- Alternative: plain text, single line ---
% \vspace{-0.3cm}
% \begin{center}
%     \textbf{Website:} \url{https://www.test.com/} \quad
%     \textbf{Code:} \url{https://github.com/test} \quad
%     \textbf{Data:} \url{https://huggingface.com}
% \end{center}

% ============================================
% Define abstract content (edit once, used in both layouts)
% ============================================
\newcommand{\abstractcontent}{%
AI-Scientists arrived, but a world for them to grow in did not. 
Human scientists take theirs for granted --- centuries of open problems and shared ideas, waiting in the scientific literature. 
To AI-Scientists, that world is readable but not runnable, so we build \swplain for them:
every research paper becomes an executable environment. 
Our autonomous pipeline extracts each paper's open-ended research question and reproduces its proposed method as the baseline; 
an AI-Scientist entering the environment is evaluated against what the human authors achieved.
% Entering an environment, \aiscis are evaluated relative to what human scientists achieved. 
% 
We release the first 1,000 verified environments. 
With the world as an \textit{evaluation}, we found that frontier AI-Scientists can already surpass the human researchers
% ' method 
74\% of the time. 
As a \textit{record of how AI does science}, it contains 6,400 hours of research traces, with every idea, experiment, failure, and success.
As a \textit{training ground}, we are able to raise ($\uparrow$158\%) a small model to match frontier models in research ideation. 
At last, \aiscis' own world has arrived, grown from and growing with science itself.
% AI-Scientists have arrived; a world for them to grow in has not. 
% AI-Scientists have arrived, but a world for them to grow in did not. 
% % We, human scientists, take it for granted --- centuries of shared ideas and open problems, waiting in the scientific literature. 
% Human scientists take theirs for granted --- centuries of shared ideas and open problems, waiting in the scientific literature. 
% To AI-Scientists, that world is readable but not runnable, so we introduce \sw: every research paper becomes an executable research environment. 
% % To AI-Scientists, that world is readable but not runnable, so we introduce Scientist World: every published paper becomes an executable research environment. 
% An autonomous pipeline extracts each paper's research question as the open-ended research task and reproduces its proposed method as the baseline. When an AI-Scientist enters an environment, it is judged by how far it surpasses what human scientists achieved. We release the first 1,000 verified environments. With the world as a benchmark, we found that frontier AI-Scientists can already surpass the human researchers' method in 76\% of their attempts. To understand how 
% % AI-Scientists
% AI
% does science, we release a dataset with 6,400 hours of their research traces, down to every idea, experiment, failure, and success. Using the world as a gym, we are able to train (158\% improvement) a small model to match frontier models at proposing research ideas. At last, the AI-Scientists' world has arrived, grown from and growing with science itself.
}

% ============================================
% First page layout: uncomment ONE of the three options below
% Preview styles: https://drive.google.com/drive/u/1/mobile/folders/1A1ZqEQ6d_rEeBucmZcY07C5M_dHQwZ_I
% ============================================

% --- OPTION 1: Teaser above abstract (currently active) ---
\vspace{0.006cm}
\ifnum\siteAt=1 \vspace{0.075cm}\fi  % = -0.07cm + what the link above gave up (0.145cm), so the teaser does not move
\ifnum\siteAt=5 \vspace{0.12cm}\fi

\vspace{-.474cm}

\begin{figure}[!ht]
    \centering
    % Robot + benchmark only, same size and place as the Canva teaser
    % (figures/sw_teaser.pdf; positions in its pt, page 1400.25 wide).
    % Title and Code / Dataset / Environments links are cut from the Canva
    % export (title scaled 30 -> 33 Canva pt, i.e. 1pt bigger in the paper; Code sits
    % on the title's baseline); the bar panel (figures/teaser_bench.pdf) is the Canva chart
    % redrawn side by side (improve | innovate, eight models each, scores from
    % plots/fig_leaderboards.py; built by plots/fig_teaser_bench.py, spanning slab bottom to
    % the top of Code); figures/teaser_robot.png is the drawn AI-Scientist.
    % To go back to a single Canva export, use the picture commented below.
    % background: 0 = grey card behind the leaderboard only, 1 = no grey card, 2 = one grey card behind the whole teaser,
    % 3 = no grey, each leaderboard column in its own outlined box (PANEL_BG=boxes)
    \def\teaserBG{3}%
    \def\tTitleX{0.3454}% title ink left = the leaderboard's logo column
    \def\tTitleY{0.4196}\def\tBenchY{0.0116}\def\tShift{0}%
    \def\tBenchX{0.3150}\def\tBenchW{0.6850}\def\tRobotX{0.0192}\def\tRobotY{-0.0217}\def\tRobotW{0.2849}%
    \def\tCodeX{0.0648}\def\tCodeY{0.3863}\def\tDataX{0.0648}\def\tDataY{0.3542}\def\tEnvX{0.0656}\def\tEnvY{0.3224}%
    \ifcase\teaserBG
        \def\tBench{figures/teaser_bench_mid.pdf}\def\tLinks{figures/sw_teaser.pdf}\def\tRobot{figures/teaser_robot_angle_podium_D.png}%
    \or
        \def\tBench{figures/teaser_bench_mid_nobg.pdf}\def\tLinks{figures/sw_teaser.pdf}\def\tRobot{figures/teaser_robot_angle_podium_D.png}%
    \or  % sw_teaser_nobg.pdf = sw_teaser.pdf without its white page (plots/make_sw_teaser_nobg.py); robot body filled white (plots/robot_white_fill.py)
        \def\tBench{figures/teaser_bench_mid_nobg.pdf}\def\tLinks{figures/sw_teaser_nobg.pdf}\def\tRobot{figures/teaser_robot_angle_podium_D_fill.png}%
    \or
        \def\tBench{figures/teaser_bench_mid_boxes.pdf}\def\tLinks{figures/sw_teaser.pdf}\def\tRobot{figures/teaser_robot_angle_podium_D_leg.png}\def\tTitleX{0.2993}% title 3pt left of the logo column (0.3057 = on it)
        % boxes: rows span the left block (title ink top = top of Code, last bar = bottom of the slab), box edges 8pt
        % outside; contents up 6pt (0.0127) into the gap under the affiliations, title ink 7pt above the boxes
        % robot at 0.85 (podium 0.0305 from the text margin and from the boxes), links centred on it; boxes fill the rest
        \def\tTitleY{0.4322}\def\tBenchY{-0.0012}\def\tShift{0.0127}%
        \def\tBenchX{0.2769}\def\tBenchW{0.7231}\def\tRobotX{0.0209}\def\tRobotY{-0.0205}\def\tRobotW{0.2422}% long legs (LEG=122), whole robot 4pt up so the slab sits just above the box bottom (-0.0290 = level); revert: teaser_robot_angle_podium_D.png, \tRobotY{0.0006}
        \def\tCodeX{0.0457}\def\tCodeY{0.3713}\def\tDataX{0.0457}\def\tDataY{0.3392}\def\tEnvX{0.0465}\def\tEnvY{0.3074}%
    \fi
    % very light backdrop behind the robot (boxes layout): 0 none, 1 disc, 2 blob, 3 arch, 4 duo (lavender | sky),
    % 6 archblob (arch outline, loose edge), 7 duo2 (two loose washes overlapping); 6 and 7 are lighter
    % (BG=... python3 plots/robot_backdrop.py on the white-filled robot)
    \def\robotBG{7}%
    \ifcase\robotBG\or\def\tRobot{figures/teaser_robot_bg_disc.png}\or\def\tRobot{figures/teaser_robot_bg_blob.png}%
    \or\def\tRobot{figures/teaser_robot_bg_arch.png}\or\def\tRobot{figures/teaser_robot_bg_duo.png}%
    \or\or\def\tRobot{figures/teaser_robot_bg_archblob.png}\or\def\tRobot{figures/teaser_robot_bg_duo2.png}\fi
    {\setlength{\unitlength}{\textwidth}%
    % mode 2 keeps the same layout box (page 1 has no spare height): the contents move up 0.017 (8pt) into the gap
    % under the affiliations, and the card adds 0.0212 (10pt) above the title ink and below the podium
    \ifnum\teaserBG=2 \begin{picture}(1,0.4596)(0,-0.017)\else\begin{picture}(1,0.4596)(0,-\tShift)\fi
        \ifnum\teaserBG=2 \put(0,-0.0096){\tikz\fill[threeUseCard, rounded corners=4pt] (0,0) rectangle (\textwidth,0.4836\textwidth);}\fi
        % title ink left = panel left, 0.021 above it; panel right = text right;
        % robot centred between the text margin and the panel (0.0305 each side), podium bottom = panel bottom
        % \put(0.3111,0.4196){\includegraphics[width=0.6860\unitlength]{figures/teaser_title.pdf}}% title ink left = panel left
        \put(\tTitleX,\tTitleY){\includegraphics[width=0.6860\unitlength]{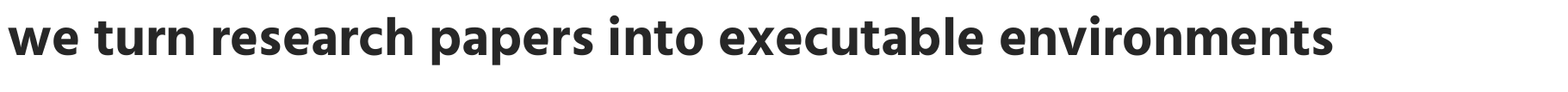}}% plots/fig_teaser_title.py; title ink left = the panel's logo column
        % \put(0.3150,0.0116){\includegraphics[width=0.6850\unitlength]{figures/teaser_bench.pdf}}   % lighter bars
    \put(\tBenchX,\tBenchY){\includegraphics[width=\tBenchW\unitlength]{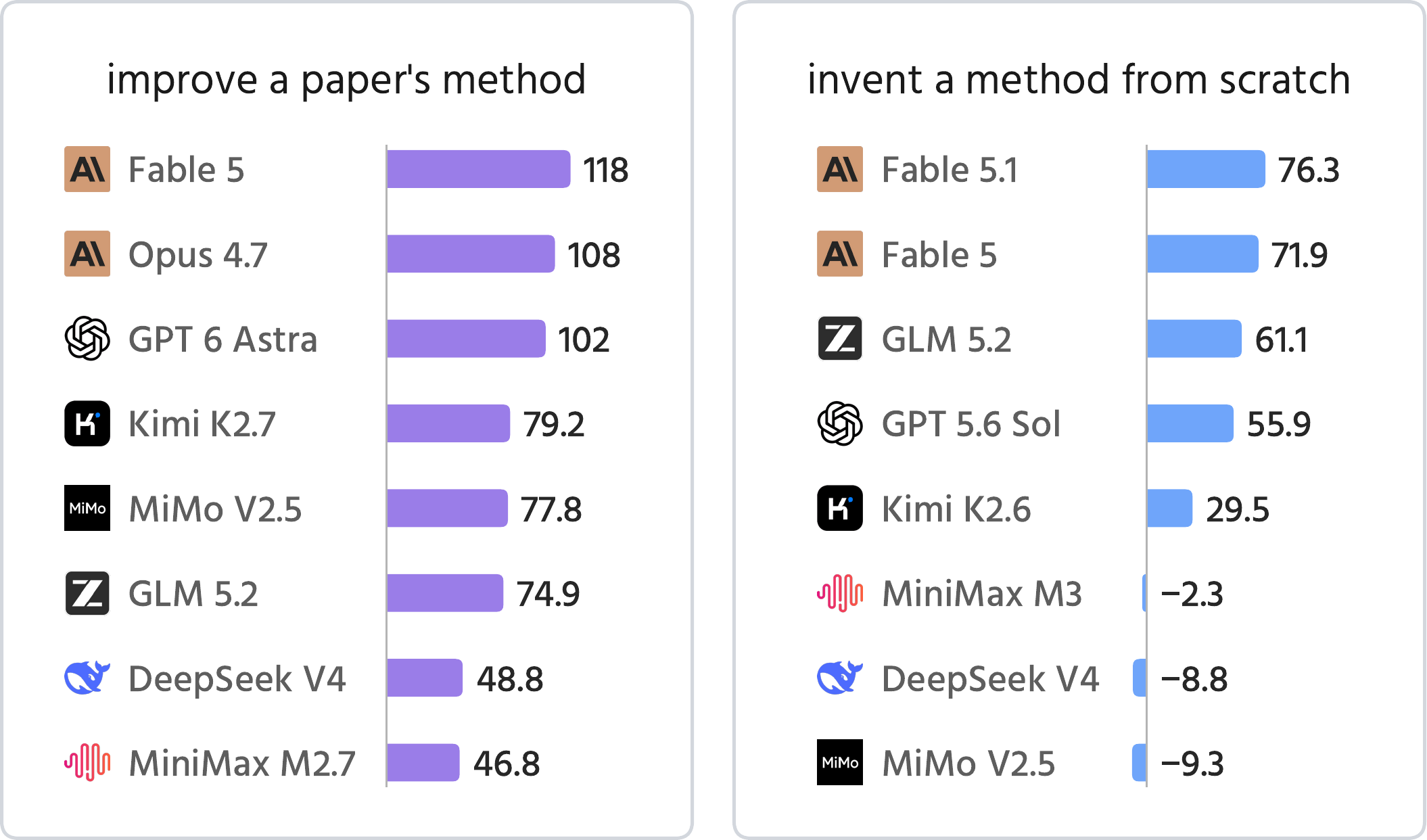}}% BARS=mid [PANEL_BG=0] python3 plots/fig_teaser_bench.py
    % \put(0.3150,0.0116){\includegraphics[width=0.6850\unitlength]{figures/teaser_bench_deep.pdf}}% BARS=deep (too dark)
        % links: Code ink top = panel top, 0.0115 ink gap between rows, centred on the robot
        \put(\tCodeX,\tCodeY){\href{https://github.com/zlab-princeton/scientist_world}{\includegraphics[trim=46 583.75 1225.25 41,clip,width=0.0921\unitlength]{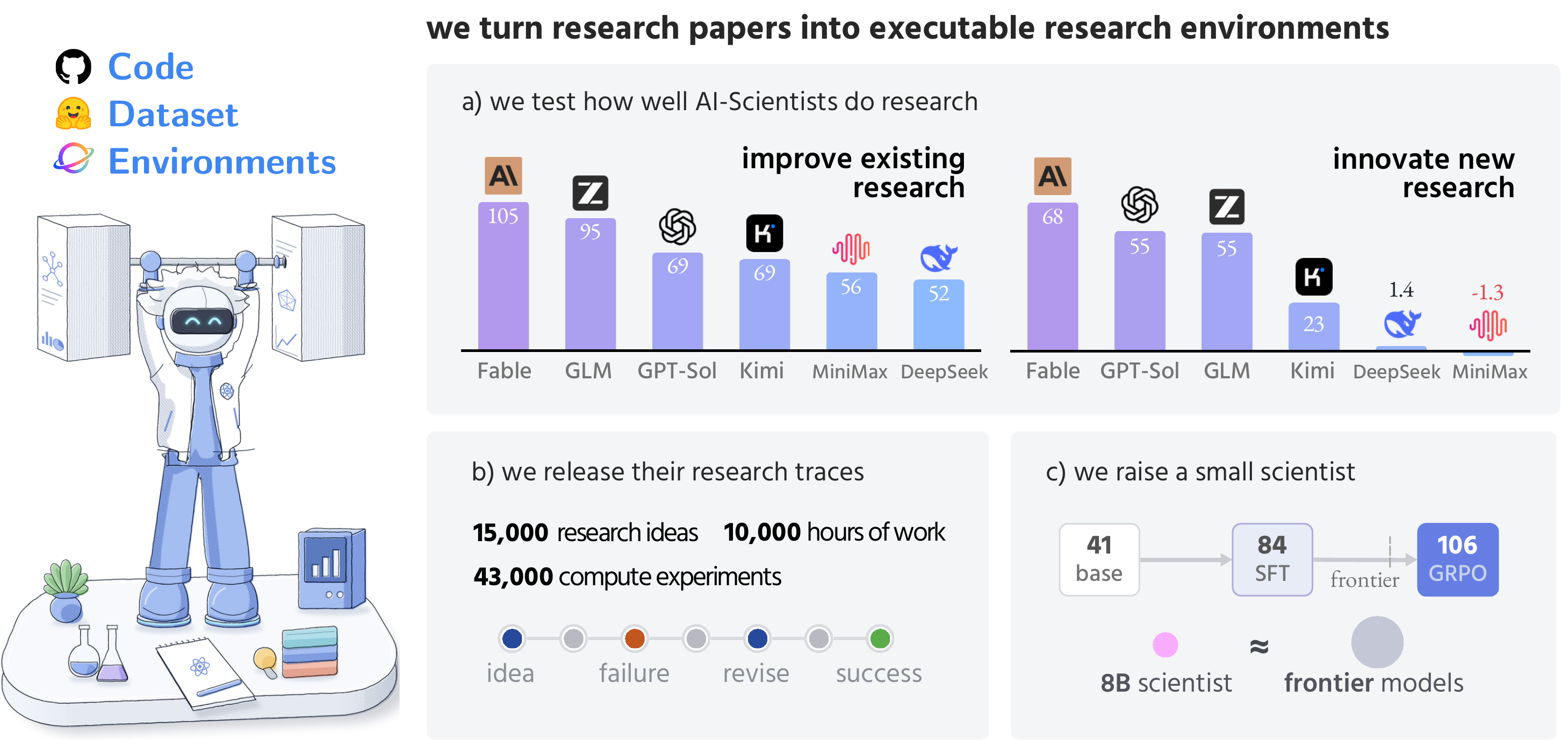}}}% Code
        \put(\tDataX,\tDataY){\href{https://huggingface.co/datasets/zlab-princeton/scientist-world-traces}{\includegraphics[trim=46 542.75 1185.25 83,clip,width=0.1207\unitlength]{\tLinks}}}% Dataset
        \put(\tEnvX,\tEnvY){\href{https://github.com/orgs/ScientistWorld/repositories}{\includegraphics[trim=45 502.75 1098.25 124,clip,width=0.1835\unitlength]{\tLinks}}}% Environments
        % \siteAt=3: a Website row one row-pitch above Code, same icon size, font (CM Sans Bold 11.1pt) and blue as the Canva rows
        \ifnum\siteAt=3 \put(0.0488,0.4080){\href{https://ai-scientist.world}{\raisebox{-1.08pt}{\includegraphics[height=10.4pt]{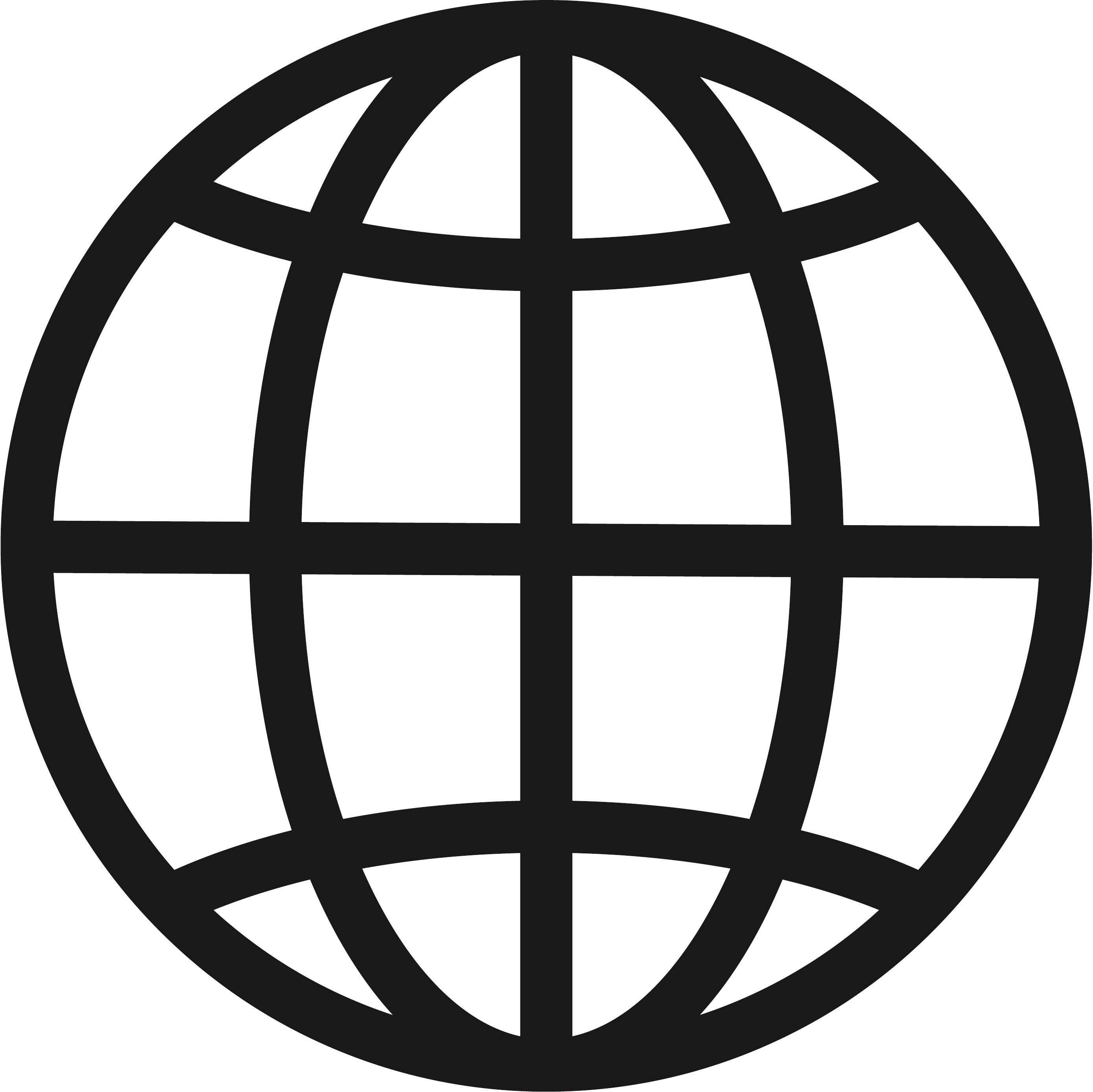}}\hspace{5.4pt}{\fontsize{11.1}{13}\usefont{T1}{lmss}{bx}{n}\color[HTML]{3E86F1}Website}}}\fi
        % robot: top 0.024 below Environments
        % \put(0.0134,-0.0272){\includegraphics[width=0.2733\unitlength]{figures/teaser_robot.png}}   % original robot (fig_teaser_robot.py)
        % \put(0.0192,-0.0217){\includegraphics[width=0.2849\unitlength]{figures/teaser_robot_angle_podium.png}}   % robot in the panel's lavender/sky only
    \put(\tRobotX,\tRobotY){\includegraphics[width=\tRobotW\unitlength]{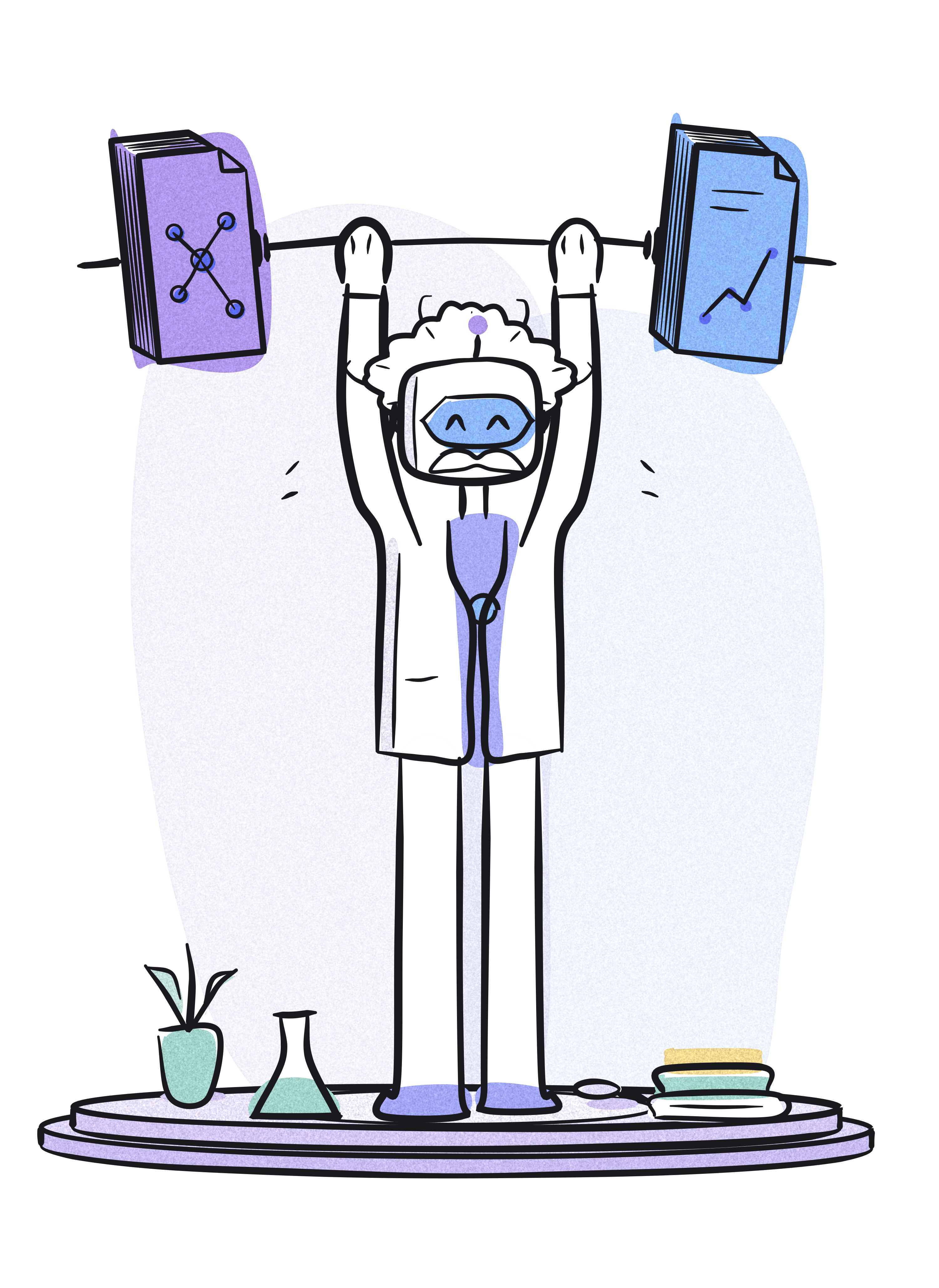}}% TONE=D python3 plots/fig_teaser_robot_angle.py figures/teaser_robot_angle_podium_D
    \end{picture}}
    % {\setlength{\unitlength}{\textwidth}%
    % \begin{picture}(1,0.51163)
    %     \put(0,0){\includegraphics[width=\textwidth]{figures/sw_teaser.pdf}}
    %     \put(0.0605,0.3940){\href{https://github.com/orgs/ScientistWorld/repositories}{\phantom{\rule{0.1438\unitlength}{0.0321\unitlength}}}}% Environments
    %     \put(0.0605,0.4288){\href{https://github.com/orgs/ScientistWorld/repositories}{\phantom{\rule{0.0832\unitlength}{0.0321\unitlength}}}}% Dataset
    %     \put(0.0605,0.4612){\href{https://github.com/orgs/ScientistWorld/repositories}{\phantom{\rule{0.0548\unitlength}{0.0321\unitlength}}}}% Code
    % \end{picture}}
    % \vspace{.003cm}
    \caption{\textbf{We introduce \swplain.} Each research paper becomes an executable environment, where an \mbox{\aisci}'s \metricfull (\metric) is above zero when its method improves on the paper's reproduced one. We test how well \aiscis improve existing research (left) and innovate new research from scratch (right). We show the best model of each lab plus the next best overall; full leaderboards are in \figref{fig:leaderboards}.}
    \label{fig:teaser}
    % \vspace{-.28cm}
\end{figure}

% teaser -> abstract: 0.3cm with the link under the caption; with the link under the affiliation (\siteAt=1) the
% teaser sits closer to the abstract and that space goes above the link instead (page 1 has no slack)
\ifnum\siteAt=1 \vspace{-0.12cm}\else\vspace{.3cm}\fi

\ifnum\siteAt=0 \linksblock\fi

% \vspace{-.474cm}

% \vspace{-0.1cm}
\vfill

\begin{abstractblock}
\abstractcontent
\end{abstractblock}
\ifnum\siteAt=4 \vspace{-0.1cm}{\centering\large\sffamily\sitelink\par}\fi

\newpage

% --- OPTION 2: Abstract above teaser ---
% \vspace{-0.3cm}
%
% \begin{abstractblock}
% \abstractcontent
% \end{abstractblock}
%
% \vspace{0.0cm}
%
% \begin{figure}[!bh]
%     \centering
%     \includegraphics[width=0.8\textwidth]{figures/teaser.pdf}
%     \caption{\textbf{Put here the caption for your figure.}}
%     \label{fig:teaser}
% \end{figure}
%
% \newpage

% --- OPTION 3: Abstract only (no teaser on first page) ---
% \begin{abstractblock}
% \abstractcontent
% \end{abstractblock}

% ============================================
% Main content
% ============================================

\section{Introduction}

\addtolength{\parskip}{3pt}
\lineskiplimit=-4pt

\noindent\hbox to\textwidth{A new kind of scientist has arrived --- for the first time, an intelligence advances the very thing it is made of.}\newline
\aiscis are the first whose discoveries improve the discoverer \citep{boiko2023autonomous, gottweis2025accelerating, mitchener2025kosmos}. They now invent new AI on their own \citep{yamada2025aiscientist, schmidgall2025agent, lu2026automation}.
This could set off a chain reaction, and every gain makes the next one faster \citep{romeraparedes2024mathematical, novikov2025alphaevolve}.
% That is a chain reaction, and every gain makes the next one faster \citep{romeraparedes2024mathematical, novikov2025alphaevolve}.
This is the power of a scientist, the power of research.

\textit{What is research?} 
It is not a single ability, and it is not a single task. 
Watch a scientist at work: 
they read until they find something no one has explained, guess at the reason, and build an experiment to test it. 
It fails. 
They eventually figure out what the failure is telling them, adjust, and run it again. 
It works.
Next month they are doing the same on a different problem, in a different part of the field. 
What makes them a scientist is not any one of these moves; 
it is the judgment that runs through all of them, and carries from one problem to the next. 
That is why research cannot live in a single task. 
Nailing one problem is a trick; research is the ability to discover again and again, problem after problem. 
We believe the same for a scientist of AI.

% But a scientist without a world is only a narrator.
% A scientist like that never stops growing, and only a world of discovery is big enough to grow in. Human researchers already live in such a world: the \textit{scientific literature}, expanding day by day. It provides two things. The first is \textit{problem definition}. Each paper poses a research question that no one has answered yet, and even once it is answered, that answer is only a checkpoint; research can almost always go further. The second is \textit{idea sharing}. No scientist starts from nothing; they build on the approaches and findings that others have already published. The literature is the shared record of everything the community has tried, what worked and what did not, so each new scientist can begin from the frontier instead of from scratch. Together, problems and ideas are what research runs on: something to work on, and something to work with. This is the world we have always worked in, and it is exactly the world a \aisci needs.

% A scientist like that never stops growing, and only a world of discovery is big enough to grow in. 
A scientist like that keeps growing, and it needs a world big enough to grow in. 
Human researchers already live in such a world: the \textit{scientific literature}.
% , expanding day by day. 
It provides two things. 
The first is \textit{problem}: each paper defines a different research question no single answer can exhaust.
% Each paper poses a research question that no one has answered yet, and even once it is answered, that answer is only a checkpoint; research can almost always go further. 
The second is \textit{idea}: each paper contributes to the field's shared memory of approaches, findings, successes, and failures.
% No scientist starts from nothing; they build on the approaches and findings that others have already published. The literature is the shared record of everything the community has tried, what worked and what did not, so each new scientist can begin from the frontier instead of from scratch. 
Together, they drive research as target and foundation. This is the world human scientists have always relied on, and the one \aiscis need.

% To this end, we introduce Scientist World. 
We introduce Scientist World
% We introduce \sw
: a living research universe built from the scientific literature itself.
% In this world, built out of the literature itself, each paper becomes an environment. 
In it, each paper becomes an executable scientific environment. 
We designed a multi-agent pipeline for this conversion, with no human in the loop.
There are four steps: 
\textit{(1)} reproduce the paper's method until it recovers the paper's result;\textit{ 
(2) }design a research task by defining what counts as scientific progress without shortcuts or gaming; 
\textit{(3)} package it so it runs on any machine; and 
\textit{(4)} verify it with independent agents. 
An \mbox{\aisci} is then scored by how much its own method improves over the human researchers' method, which we call its \metricfull (\metric). 
Building from high-quality papers, we first release 1{,}000 environments,
% and the world 
and it keeps growing with the literature (\figref{fig:teaser}).

\begin{wrapfigure}[18]{r}{0.338\textwidth}
    \vspace{-12pt}
    \centering
    \includegraphics[width=0.338\textwidth]{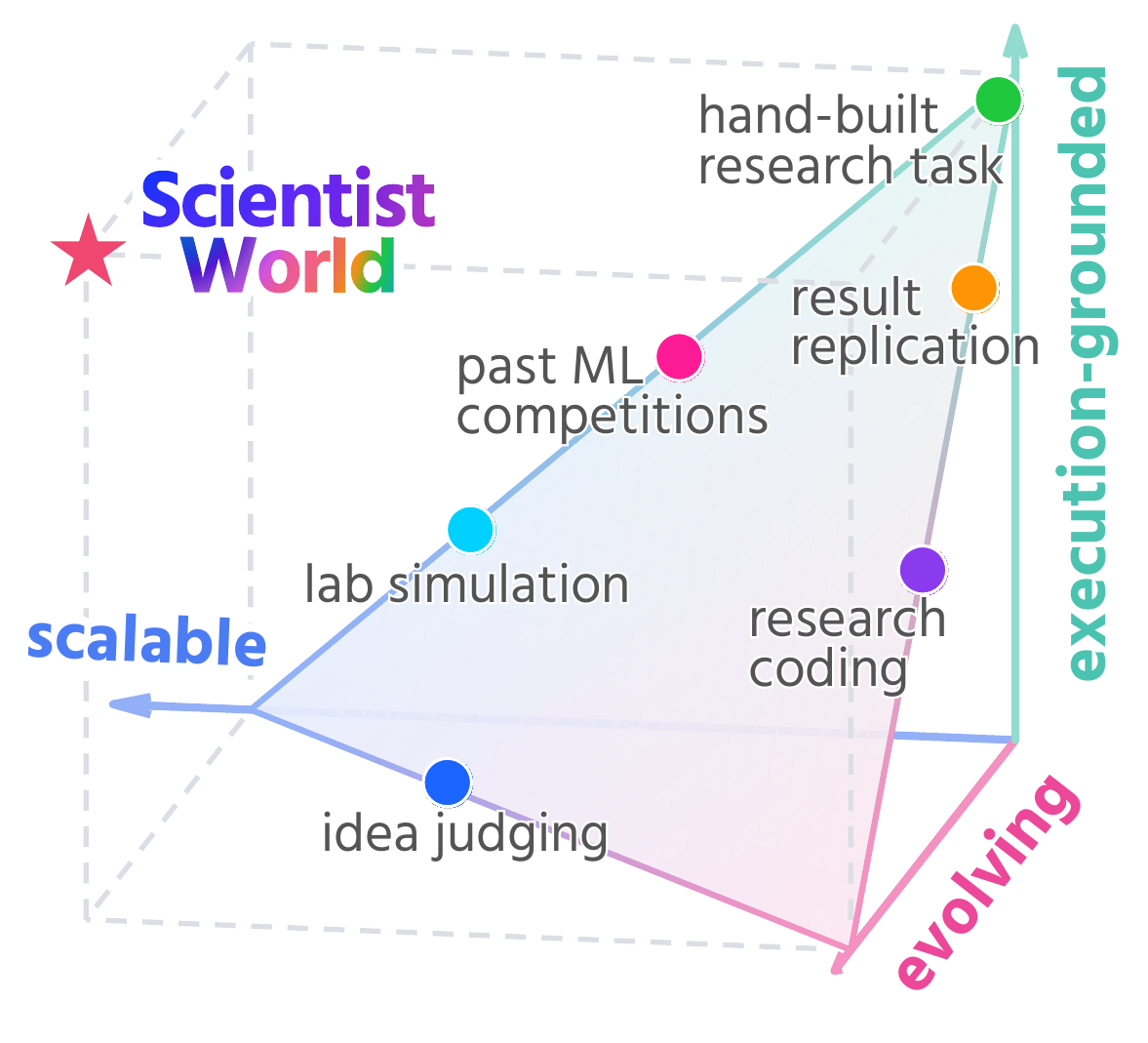}
    \vspace{-14pt}
    \caption{\textbf{Where 
    % \sw
    Scientist World
    stands.} Prior \aisci environments sit on the trade-off surface we describe; \swplain reaches the corner where scale, evolve, and execution meet.}
    \label{fig:three_axes}
    \vspace{-8pt}
\end{wrapfigure}
Because the pipeline runs with no human in the loop, \swplain is \prop{axScalable}{Scalable}: it can grow fast to cover a broad range of fields and topics. Two more properties are inherited from science itself. \prop{axEvolving}{Evolving}: science never stands still, so a fixed set measures yesterday, what a scientist has already seen; \swplain absorbs new literature every day, measuring research rather than recall. \prop{axExecution}{Execution-grounded}: a fluent \aisci can make any idea sound bold, but ideas that impress on paper might fail on execution \citep{kapoor2024aiagents, si2025ideation}; because every environment is built and measured through real experiments, research is judged by execution, not rhetoric.

Environments for research exist, but none, to our knowledge, holds the three properties together. Most are not \textit{scalable}: they recycle problems from past competitions \citep{huang2024mlagentbench, chan2024mle, nathani2025mlgym}, confine to simulated worlds \citep{jansen2024discoveryworld, chen2024scienceagentbench}, or hand-pick tasks from a few papers \citep{wijk2024rebench, zhang2025mlrcbench, garikaparthi2026researchgym}. Nearly all are not \textit{evolving}: sets are fixed at release and age into training data \citep{sainz2023nlp, xu2024benchmark}, and those that refresh cover only narrower skills such as code \citep{jain2024livecodebench, zhang2025swebenchlive}. And others are not \textit{execution-grounded}: they reproduce results the field already knows \citep{siegel2024corebench, starace2025paperbench, kon2025expbench} or judge proposed ideas as text \citep{guo2024ideabench, si2025ideation}. 
Scientist World holds three at once (\figref{fig:three_axes}).
% \sw holds all three at once (\figref{fig:three_axes}).
% \sw is the first world to hold all three at once.

We first use 
% \sw
Scientist World
to evaluate frontier \aiscis. Each scientist attempts an environment in two modes: \textit{improve}, which starts from the paper's \textit{idea} with its reproduced method; and \textit{innovate}, which starts from the \textit{problem} alone. The two abilities prove distinct, since the best improver is not the best inventor. 
% The gains are also earned: scores are uncorrelated with compute spent, yet agree closely with a blind judge of research quality. 
Every attempt leaves a trajectory, each decision paired with the measured outcome of a real experiment, and we release this corpus alongside the environments. Then we train with it: rewarded only by real outcomes, a small scientist learns to propose ideas that, implemented by a fixed frontier executor, reach the frontier cohort's level on the median held-out environment. Taken together, \swplain is three things at once:

\vspace{-.2cm}
\begin{itemize}[topsep=0pt, itemsep=2pt, parsep=0pt]
\item \textbf{A benchmark.} The standing measure of whether \aiscis can advance real research.
% \item \textbf{A benchmark.} The standing measure of whether \aiscis can advance real research: every attempt runs real experiments, scored on held-out results and ranked by its gain over the paper's own method.
\item \textbf{A dataset.} A growing corpus of research being done, with every attempt preserved in full loop.
% \item \textbf{A dataset.} A growing corpus of research being done: every scientist's attempt is preserved in full, its thoughts, its trials, and its failures, each tied to the measured outcome of a real experiment.
\item \textbf{A gym.} The same environments become training grounds, with real experimental results as rewards.
% \item \textbf{A gym.} The same environments become training grounds, where scientists practice, fail, and learn from rewards that are real experimental results, and the trajectories each one leaves behind train the next.
\end{itemize}

\vspace{-.18cm}

% \begin{center}
% \vspace{-16pt}
% \newcommand{\iconcell}[2]{\begin{minipage}[t]{0.30\textwidth}\centering
% \includegraphics[width=0.56\linewidth]{logos/icons/#1}\\[-7pt]
% {\small #2}\end{minipage}}
% \iconcell{benchmark.png}{a benchmark of real research}\hspace{6pt}\iconcell{dataset.png}{a dataset of every attempt}\hspace{6pt}\iconcell{gym.png}{a gym for future scientists}
% \vspace{-2pt}
% \end{center}
% \vspace{-.2cm}

If research is the ability to discover again and again, it needs a world that never runs out. \sw is that world, grown from science itself and open today. The chain reaction now has a place to happen.

\begin{figure}[h]
    \vspace{-.15cm}
    \centering
    % three grey cards (the original teaser's panel grey); each holds only its drawing, centred;
    % the subcaptions are ordinary LaTeX subcaptions under the cards
    % panels: plots/fig_three_use.py -> figures/three_use_v2_{a,b,c}.pdf (old ones: figures/three_use_{a,b,c}.pdf, drop the "v2_")
    \captionsetup[subfigure]{justification=centering, font=footnotesize, skip=3pt}
    \tcbset{threeusecard/.style={colback=threeUseCard, colframe=threeUseCard, boxrule=0pt, arc=4pt, boxsep=0pt,
        left=0pt, right=0pt, top=5pt, bottom=5pt, halign=center, valign=center, equal height group=threeuse}}
    \begin{subfigure}[t]{0.315\textwidth}
        \begin{tcolorbox}[threeusecard]\includegraphics{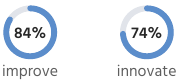}\end{tcolorbox}
        \caption{we test how well they do research}
        \label{fig:three-use-a}
    \end{subfigure}\hfill
    \begin{subfigure}[t]{0.35\textwidth}
        \begin{tcolorbox}[threeusecard]\includegraphics{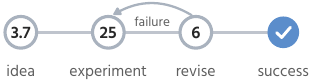}\end{tcolorbox}
        \caption{we release their research traces}
        \label{fig:three-use-b}
    \end{subfigure}\hfill
    \begin{subfigure}[t]{0.315\textwidth}
        \begin{tcolorbox}[threeusecard]\includegraphics{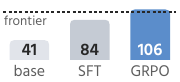}\end{tcolorbox}
        \caption{we raise a small scientist}
        \label{fig:three-use-c}
    \end{subfigure}%
    \vspace{-.3cm}
    \caption{\textbf{\swplain is three things at once.} One set of environments, grown from real papers, serves all three.}
    % \caption{\textbf{\swplain is three things at once.} \textit{(a)} the share of attempts in which the \aisci's method beats the paper's own method; \textit{(b)} every attempt kept as its full history, 6{,}400 hours in total, with the average number of idea revisions, experiments, and method revisions per attempt; \textit{(c)} the median held-out score after each training stage.}
    \label{fig:three-use}
    % \vspace{-.5cm}   % more room before Section 2
\end{figure}

\lineskiplimit=0pt
\addtolength{\parskip}{-3pt}

% \vfill

\section{From Papers to Environments}
\label{sec:method}

\vspace{-.03cm}

How well can \aiscis do research? Any answer must come from a research environment: a real problem to work on, and an objective measure of progress. A research paper is \textit{almost} such an environment. It states a problem, proposes a method, and reports what happened. Yet it is only a record of science done, not a place to do science. 
% \sw
We turns the record back into the place. It (1) rebuilds the paper's idea and experiment, (2) defines what constitutes an improvement, and (3) judges every scientist's new solution.

We design \emph{\builder} agents to perform this conversion autonomously.
% They start from the paper and a starter workspace. They first reproduce the paper's experiment and extract the problem to solve. They then design both into a self-contained environment: 
They reproduce a paper's experiment, extract the problem to solve, and design both into a self-contained environment: 
the problem becomes a task any new scientist can take up, and the reproduced result becomes the bar their solutions are scored against. 
% The rest of this section explains it.

\vspace{-.03cm}

\subsection{The Builders}
\label{sec:method-builders}

\vspace{-.18cm}

\begin{figure}[t]
    \centering
    \includegraphics[width=\textwidth]{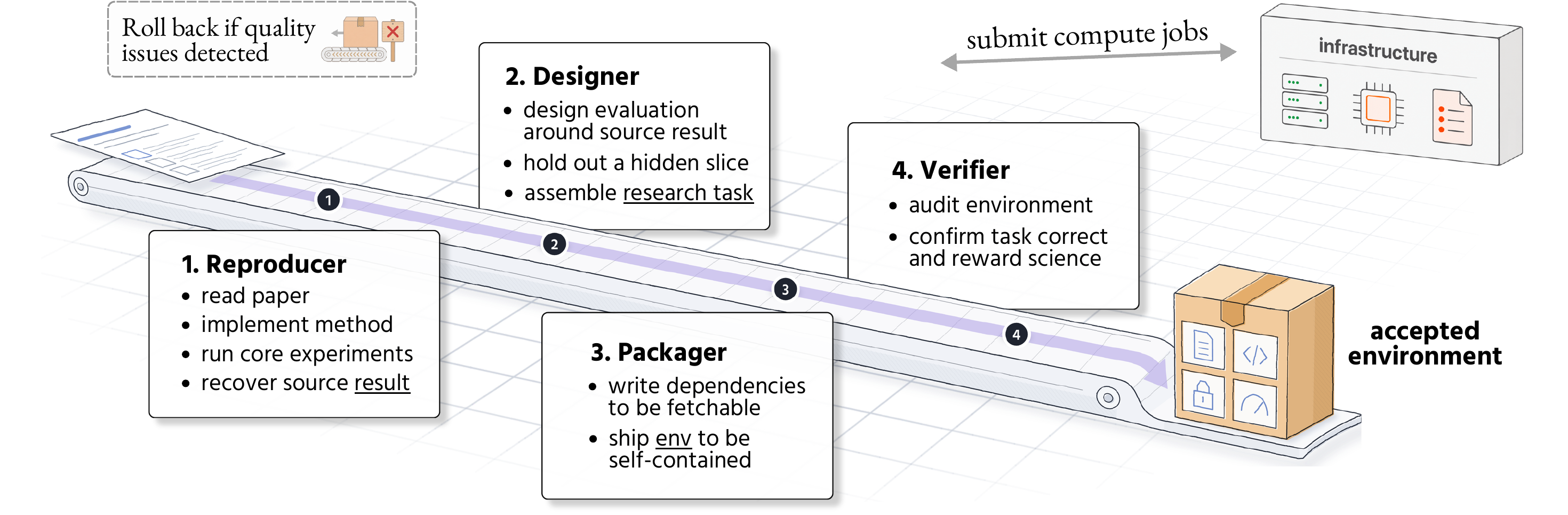}
    \vspace{-.25cm}
    \caption{\textbf{\Builders turn a paper into an environment.} Construction runs like an assembly line: a published paper enters at one end, and a reusable research environment leaves the other. Specialized \builders take turns on a single shared workspace, each refining what the previous one built and offloading heavy computation to a shared cluster. A quality gate guards the line, sending unfinished work back for another pass, so only what clears verification is accepted.}
    \label{fig:builder-iterations}
    % \vspace{-.3cm}
\end{figure}

We convert each paper through a sequence of agent iterations on one shared workspace (\figref{fig:builder-iterations}). Four roles divide the work, from reproducing the paper to verifying the finished environment. A single paper can take hundreds of unattended turns without a human in the loop. To hold the sequence together, a ladder of milestones is used to track progress. Each rung marks how much of the environment stands, from a first runnable experiment to an environment ready for new scientists. A \builder agent ends its iteration by reporting the rung it believes it reached; the next \builder begins by checking and verifying that report. At the end of the \builder iterations, the final environment is accepted or rejected based on the milestone reported.

\paragraph{Workflow.} A paper moves through \builder roles in iterations. Each agent inherits the workspace, audits the claimed milestone, and demotes whatever it cannot support. Every iteration ends at a gate: a milestone below the requirement reverts the work to an earlier iteration, or exits as failure. To be accepted, the experiment must run to support the paper's central claim. We rotate different \builder models across iterations to avoid any blind spots. What survives is a repository that any scientist can use without knowing how it was made.

% TODO: dedicated main-text figure (Canva); placeholder reuses the appendix overview.
% Spec: one paper moving through iterations on a shared workspace; role arc reproducer -> designer -> packager -> verifier; milestone ladder as the shared yardstick; gates between iterations (revert/stop arrows); ends at accepted repository.

\newpage

% \paragraph{Workflow.} A paper moves through \builder roles as a sequence of iterations. Each new agent inherits the workspace, audits the claimed milestone, and demotes whatever the workspace cannot support. Every iteration ends at a gate: a milestone below the requirement reverts the paper to an earlier iteration, or exits as failure. To be accepted, the experiment must support the paper's central claim, not merely run. We rotate different \builder models across iterations, so one model's blind spots do not persist. What survives is a repository that any later scientist can use without knowing how it was made.

\paragraph{(1) Reproducer.} The reproducer re-establishes the paper's main result. It implements both the method and the paper's original evaluation, from the authors' code if released or from scratch otherwise. Its experiments scale up from quick checks toward the full run, until the result appears. What it leaves behind we call the \emph{source method} and the \emph{source result}: the reproduced method and the numbers it achieved. 
% Everything later in the environment builds on the first and measures against the second.

\vspace{.03cm}

\paragraph{(2) Designer.} The designer turns the reproduction into a task that others can attempt. It writes the problem statement that a new scientist will read, and it fixes the rules of the task: which metric to improve, which settings must stay fixed, and which kinds of solution are allowed. It sets the \emph{source result} as the bar, so a new solution is scored by how much it improves on it. It prepares two evaluation sets: a visible one that the scientist may test against freely, and a held-out one to detect shortcuts or overfitting. 
% We place the designer after the reproducer deliberately: a task can be designed only around an experiment that already runs.

\vspace{.03cm}

\paragraph{(3) Packager.} The packager makes the designed environment run anywhere, not only on machines that built it. It replaces every local file with a script that downloads it, and declares every dependency the experiments need. The check is simple: a fresh clone of the workspace must be able to rebuild and rerun on its own.

\vspace{.03cm}

\paragraph{(4) Verifier.} The verifier audits the packaged environment before it is accepted. It rereads the workspace with a skeptic's eye, checking the code is valid, the scores match the logs, and the task follows the design rules. We keep this role separate on purpose, where the \builder that grades the work never built it.

% \paragraph{Workflow.} A paper moves through these roles as a sequence of iterations. Each new agent inherits the workspace, audits the claimed milestone, and demotes whatever the workspace cannot support. Every iteration ends at a gate: a milestone below the requirement reverts the paper to an earlier iteration, or exits as failure. To be accepted, the experiment must support the paper's central claim, not merely run. We rotate different \builder models across iterations, so one model's blind spots do not persist. What survives is a repository that any later scientist can use without knowing how it was made.

\vspace{25 pt}

\begin{wrapfigure}[8]{r}{0.433\textwidth}
    % \vspace{-.53cm}
    \vspace{-20pt}
    \centering
    \includegraphics[width=\linewidth]{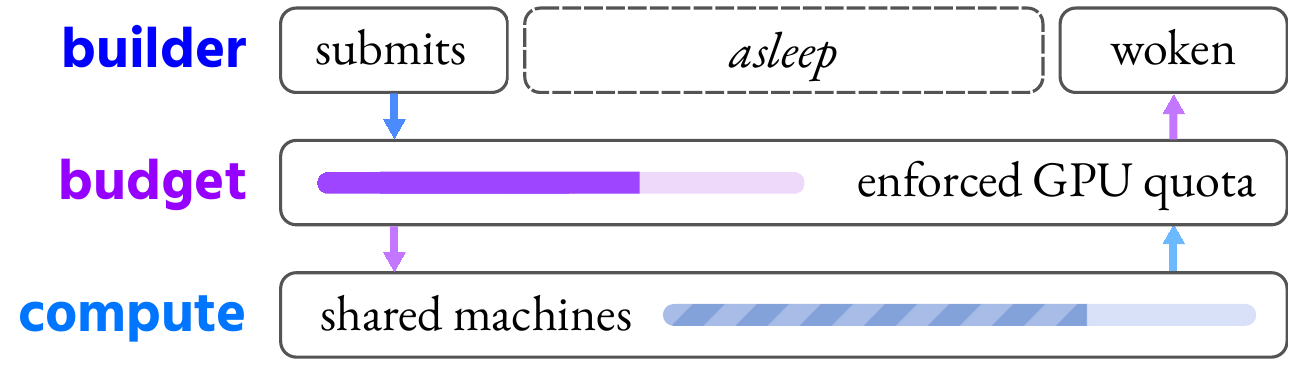}
    \vspace{-13pt}
    \caption{\textbf{Using compute.} A \builder submits a job to the shared pool, sleeps, and is woken with the logs; its GPU budget is enforced outside the loop.}
    \label{fig:infra}
    \vspace{-8pt}
\end{wrapfigure}

\paragraph{Infrastructure.} All experiments run on a shared pool of compute machines (\figref{fig:infra}), described in \secrefc{sec:infrastructure} with details. A \builder never runs jobs on its own hardware: it submits a job, waits, and is woken with the logs when the job ends. We cap each paper's GPU time a \builder can use with a budget enforced outside the \builder; when it runs out, the \builder wraps up with what it has. With thousands of papers queuing, no single one deserves unbounded compute.

\newpage

\subsection{The Environment}
\label{sec:method-conversion}

\begin{figure}[t]
    \centering
    \includegraphics[width=\linewidth]{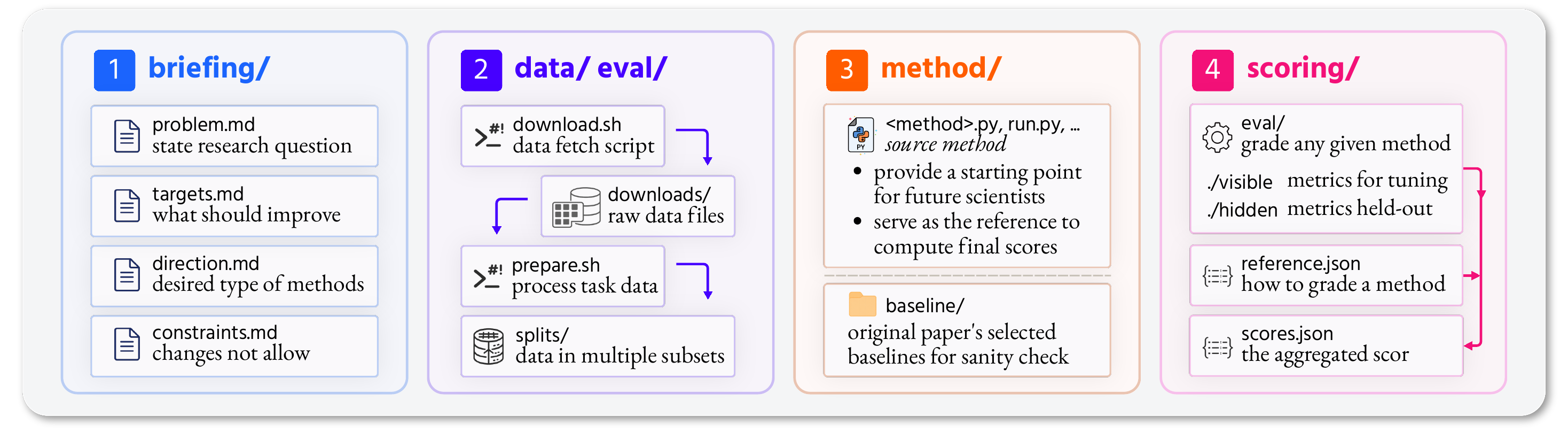}
    \caption{\textbf{Anatomy of an environment.} A produced environment is a single self-contained workspace, laid out so the problem stays separate from the solution. The \texttt{briefing}, \texttt{data}, and \texttt{scoring} fix what to solve and how it is judged without referring to any one method, which keeps the task open to future approaches and stops a score from depending on the implementation. The \texttt{method} is the paper's \textit{source method} reproduced, and the scores it reaches are the reference we compare against, so a new method's final score is the improvement it makes over them.}
    \label{fig:env-anatomy}
\end{figure}

An accepted environment is a paper refactored into an executable research setting. It does not ask a future scientist to reproduce the paper again; it gives that scientist a prepared place to work on the paper's problem. The repository contains four parts, in the order a scientist encounters them: (1) a task briefing that states the problem, target, allowed direction, and constraints; (2) data and evaluation files that define what a method must produce and how it is scored; (3) the reproduced method and baselines, which provide starting solutions (the \emph{source method} can optionally be withheld during evaluation); and (4) scripts, score files, logs, job outputs, and commits kept as the construction record. \figref{fig:env-anatomy} walks through one produced environment in full.

\paragraph{What makes it an environment.} An environment is more than reproduction code. It keeps the paper's research situation open for new methods. We design each environment around three principles. First, \builders may shrink the experiment scale to reduce compute cost, without changing the scientific question. The smaller task should still ask whether the paper's idea improves the same kind of outcome under the same kind of evaluation. Second, the task must remain non-trivial. The source method should still leave a meaningful gap over simple baselines, so future methods have something real to improve. Third, we treat the reproduced method as one solution, not as the task itself. The data interface and evaluator are written independently of the source method, so a future method can replace it by producing the same kind of artifact.

\paragraph{What makes it a fair task.} We want a future method to win by making progress on the same question, not by changing the question itself. 
% So we write the comparison rules before any scientist agent runs. 
% The task briefing splits these 
So we write research rules across four files. The problem file states the research question. The target file names what should improve, such as loss or accuracy. The direction file says what kinds of method changes still answer that question. The constraints file lists changes we disallow because they would make the result about a different problem. 
% For example, in Mamba architecture research, an agent should not improve the score by switching to a Transformer.
% The constraints
The constraints also rule out direct scale gains, for example by holding model size, data size, or compute budget fixed when those factors are not the research question.

\paragraph{How a scientist's attempt is graded.} A scientist's proposed method is graded on how far it improves over the paper's \textit{source method}, on the quantity the claim is about, such as accuracy or loss. This can be viewed as a comparison with human researchers. We score the method over multiple experiments with different metrics used in the original paper. Besides a visible set for development, we reserve a held-out slice, so a method must improve the task rather than exploit one visible number through shortcuts or overfitting (\secrefc{sec:eval-scoring}).

\paragraph{\metricfull (\metric).} We introduce \metric, scored per paper as $\text{\metric}=100\sum_c w_c\,\delta_c/(\sigma_c b)$, where $\delta_c$ is the signed gain over the source method on metric $c$, $w_c$ its weight, $\sigma_c$ a frozen frontier cohort's spread on it, and $b$ sets the cohort's mean $|\text{\metric}|$ to 100; 0 matches the paper, and negative is worse (\secrefc{sec:eval-scoring}).

\section{Scientist World}

\label{sec:world}

\sw is the artifact built from the environment-construction system: the selected paper set, the generated repositories, and the records produced while building them. This section reports how we choose a valuable, buildable 1{,}000-paper source set and how the batch conversion ran.

\begin{figure}[t]
    \noindent
    \begin{minipage}[t]{0.292\textwidth}
        \centering
        \includegraphics[width=\linewidth]{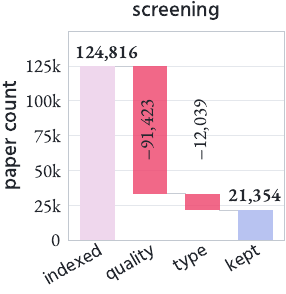}\\[-4pt]
        {\hspace{0.4cm} \small (a) screen out infeasible papers}
    \end{minipage}%
    \hspace{0.045\textwidth}%
    % \hspace{0.015\textwidth}%
    % \noindent\includegraphics[width=0.005\textwidth]{figures/segment.pdf}%
    % \hspace{0.015\textwidth}%
    \begin{minipage}[t]{0.38\textwidth}
        \centering
        \includegraphics[width=\linewidth]{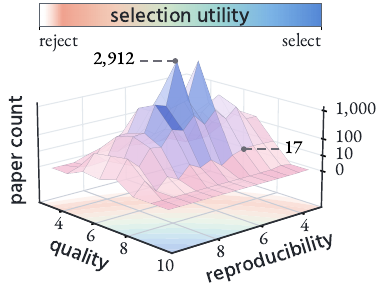}\\[-4pt]
        {\small (b) rank and keep the top 1{,}000}
    \end{minipage}%
    \hspace{0.03\textwidth}%
    \begin{minipage}[t]{0.248\textwidth}
        \centering
        \includegraphics[width=\linewidth]{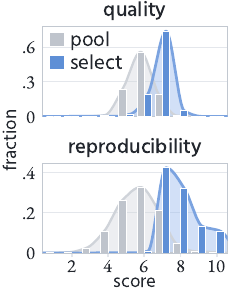}\\[-4pt]
        {\small \hspace{0.05cm} (c) selected scores higher}
    \end{minipage}
    \vspace{.15cm}
    \caption{\textbf{Selecting 1{,}000 papers from 124{,}816 indexed.} In a first pass (a), we drop papers that are \textit{low quality} or are mainly a \textit{type} of theory, a benchmark, a dataset, or a single finding, which leaves 21{,}354. In (b), we look at these more closely and favor work that is both \textit{high quality} and \textit{compute-friendly}, keeping the best 1{,}000. In (c), we check the result: the papers we picked (blue) are clearly stronger and more reproducible than the pool we started from (grey).}
    \label{fig:paper-selection}
    \vspace{-.15cm}
\end{figure}

\subsection{Building at Scale}

\label{sec:world-building}

\paragraph{Selecting papers.} Our selection pipeline keeps papers that are worth building under budget (\figref{fig:paper-selection}). We start from arXiv's core AI categories over fourteen months to obtain 124{,}816 unique papers. We (1) run two fast screens with \model{MiniMax M2.7}~\citep{minimax}. The quality screen keeps papers in roughly the top 25\% quality band. The type screen keeps empirical-method papers, whose main contribution is quantitative optimization. Together they leave 21{,}354 papers. Then (2) we run more detailed judgments for each paper on two quantities: research quality and reproduction difficulty (we prefer papers with low compute costs). 
% After removing papers below the middle quality or reproduction threshold, 8{,}987 candidates remain. 
From the remaining pool, with a geometric-mean utility over quality and feasibility, we sample the category-balanced top 1{,}000 papers.
% across eight broad areas.
% 18 fine-grained categories, which we group into eight broad areas for analysis.

\vspace{.1cm}

% Build numbers below = live 999-manifest pull (refresh before release).
% [11] caps the narrowed lines at the available text so the wrap can't leak past \newpage.
\begin{wraptable}[11]{r}{0.37\linewidth}
    \vspace{-1.295\baselineskip}
    \centering
    \begin{softcolortable}[1]
    \setlength{\tabcolsep}{4pt}%
    \setlength{\softTableContentWidth}{\dimexpr\softTableWidth-6\tabcolsep\relax}%
    \renewcommand{\arraystretch}{1.35}%
    \begin{tabular}{
        >{\leftskip=4pt \raggedright\arraybackslash}m{0.42\softTableContentWidth}
        >{\centering\arraybackslash}m{0.24\softTableContentWidth}
        >{\centering\arraybackslash}m{0.34\softTableContentWidth}
        }
    \toprule
    \rowcolor{computeHeader}
    \textbf{per paper} & \textbf{median} & \textbf{total} \\
    \midrule
    \rowcolor{gray!8}
    compute jobs & 76 & 239{,}506 \\
    \rowcolor{white}
    GPU-hours & 0.4 & 2{,}286 \\
    \rowcolor{gray!8}
    CPU-hours & 6 & 28{,}338 \\
    \rowcolor{white}
    agent turns & 730 & 947{,}056 \\
    \rowcolor{gray!8}
    input tokens & 28M & 51B \\
    \rowcolor{white}
    output tokens & 163k & 231M \\
    \bottomrule
    \end{tabular}
    \end{softcolortable}
    \caption{\textbf{Per-paper and total build cost.} Median and the total cost over 1{,}000 papers.}
    \label{tab:build-stats}
\end{wraptable}

\paragraph{Automation.} No human reviews any paper at any stage in our pipeline. The source arXiv categories add roughly 9{,}000 unique papers per month, of which about 1{,}500 survive screening and remain feasible. The current 1{,}000 are the first release; future batches extend the window through the same pipeline.

\paragraph{The build.} We build every paper with the same nine-iteration workflow: three reproducers, three designers, one packager, and two verifier passes (we repeat a role for multiple iterations for additional quality checks), with the model family rotating across \model{GPT}, \model{Claude}, \model{MiniMax}, and \model{GLM}. A paper may draw up to 70 GPU-hours, yet the median spends under half of one (\tabref{tab:build-stats}).

\newpage

\vspace{.5cm}

\subsection{The Built World}
\label{sec:built-world}

\vspace{-.18cm}

% We ran the construction pipeline over the selected papers and obtained \sw's first release: 1{,}000 environments, each a published paper refactored into a runnable, scored task. 
% This section describes what they contain and how reliable they are.

\begin{figure}[t]
    \centering
    \begin{minipage}[c]{0.46\textwidth}
        \centering
        \includegraphics[width=\linewidth]{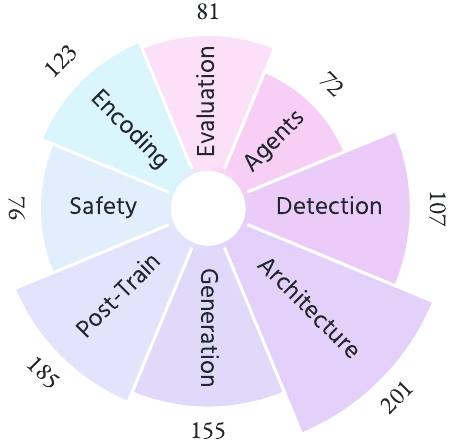}\\[4pt]
        \vspace{-.2cm}
        \captionof{figure}{\textbf{The environment composition.} The produced environments can be categorized into eight areas.}
        \label{fig:built_composition}
    \end{minipage}\hfill%
    \begin{minipage}[c]{0.45\textwidth}
        \centering

        \vspace{10pt}

        % --- Table 2: per-environment composition (spans full minipage) ---
        \begin{softcolortable}[1]
        \setlength{\tabcolsep}{4pt}%
        \setlength{\softTableContentWidth}{\dimexpr\softTableWidth-4\tabcolsep\relax}%
        \renewcommand{\arraystretch}{1.3}%
        \begin{tabular}{
            >{\leftskip=4pt \raggedright\arraybackslash}m{0.42\softTableContentWidth}
            >{\raggedleft\arraybackslash}m{0.58\softTableContentWidth}
            }
        \toprule
        \rowcolor{envHeader}
        \textbf{environment} & \textbf{average} \\
        \midrule
        \rowcolor{gray!8} \# experiments & 3.3 \\
        \rowcolor{white} \# metrics & {\footnotesize\textcolor{black!55}{(16 visible + 16 hidden)}}\,\,32 \\
        \rowcolor{gray!8} \# datasets & 1.6 \\
        \rowcolor{white} \# files & {\footnotesize\textcolor{black!55}{(3.8k lines of code)}}\,\,59 \\
        \bottomrule
        \end{tabular}
        \end{softcolortable}\\[0pt]
        \captionof{table}{\textbf{A produced environment on average.} The mean size and makeup of one environment.}
        \label{tab:env-stats}

        \vspace{9pt}

        % --- Table 3: quality and validity (spans full minipage) ---
        % Body-row height knob: \qrs is an invisible strut on each body row.
        % Raise the 17pt (height) / 6pt (depth) for taller rows. The header
        % has no \qrs, so it keeps its natural height.
        \def\qrs{\rule[-6pt]{0pt}{17pt}}%
        \begin{softcolortable}[1]
        \setlength{\tabcolsep}{4pt}%
        \setlength{\softTableContentWidth}{\dimexpr\softTableWidth-4\tabcolsep\relax}%
        \renewcommand{\arraystretch}{1.3}%
        \begin{tabular}{
            >{\leftskip=4pt \raggedright\arraybackslash}m{0.83\softTableContentWidth}
            >{\raggedleft\arraybackslash}m{0.17\softTableContentWidth}
            }
        \toprule
        \rowcolor{qualHeader}
        \textbf{across 1{,}000 envs} & \textbf{rate} \\
        \midrule
        \rowcolor{gray!8} \qrs reproduced method score within margin & 99\% \\
        \rowcolor{white} \qrs beats paper's baselines & 82\% \\
        \rowcolor{gray!8}
        \multicolumn{2}{@{\hspace{\tabcolsep}}p{\dimexpr\softTableContentWidth+2\tabcolsep\relax}@{\hspace{\tabcolsep}}}{%
            \qrs {\renewcommand{\arraystretch}{0.9}\begin{tabular}[c]{@{}l@{}}reverted iteration by a\\ later cross-check\end{tabular}}%
            \hfill
            {\footnotesize\textcolor{black!55}{(573 total, 1.6 avg)}}\,\,37\%%
        } \\
        \rowcolor{white} \qrs approved by independent verify & 100\% \\
        \bottomrule
        \end{tabular}%
        \end{softcolortable}\\[0pt]
        \captionof{table}{\textbf{Quality and validity.} How it reproduces paper method, beats baselines, and survives audit.}
        \label{tab:quality}
    \end{minipage}
\end{figure}

\paragraph{Composition.} We first release 1{,}000 environments spread across eight areas (\figref{fig:built_composition}). A scientist therefore meets problems of many kinds, not one narrow family. A typical environment is substantial, closer to a small research project than a single script (\tabref{tab:env-stats}). Because most carry several experiments rather than a single headline result, a later method has multiple quantities to improve rather than one number to chase.

\paragraph{Quality.} Every environment is built without a human in the loop, so we do not take any single \builder's word for its quality. Four checks stand behind an accepted environment, one per row of \tabref{tab:quality}:

\begin{itemize}[topsep=2pt, itemsep=6pt, parsep=0pt]
    \item \textbf{Reproduction fidelity.} The reproduced source method's numbers are checked against the paper's reported results within a small margin, across every experiment and metric used by the environment; the few exceptions are accepted only with documented justification.
    \item \textbf{Baseline separation.} Where a paper provides baselines, we reimplement them and require the reproduced source method to outperform them, preserving a real gap for future methods to close; the rate in \tabref{tab:quality} counts all 1{,}000 environments, including those without a reimplementable baseline.
     \item \textbf{Iterative cross-checking.} Each builder audits the previous milestone claim and demotes whatever the workspace cannot support. A demotion rolls the workflow back to the previous iteration; construction then resumes from that checkpoint until it passes the later checks without further reversion.
    \item \textbf{Independent verification.} The final environment must pass a verifier that did not participate in construction. Every released environment satisfies this final check.
\end{itemize}

% Because the evaluator never imports the method, a score reflects what a method produces rather than which method produced it (\secrefc{sec:eval-scoring}).

\begin{figure}[t]
    \vspace{-0.3cm}% gap to the (radial + tables) block above; more negative = smaller gap
    \centering
    \includegraphics[width=\linewidth]{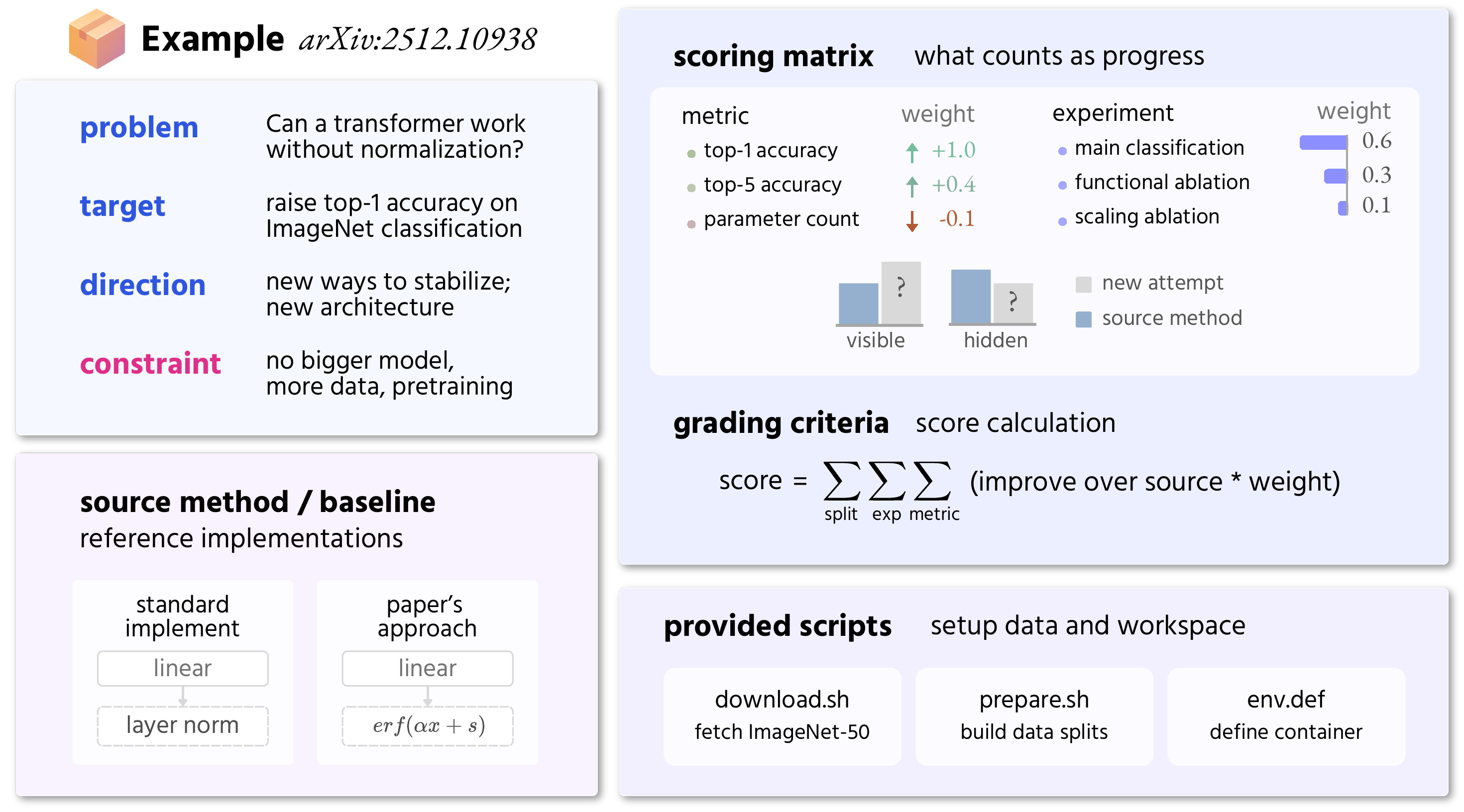}
    \vspace{-.55cm}
    \caption{\textbf{An example environment.} This environment turns a Transformer architecture paper into a constrained improvement task: raise ImageNet accuracy without changing scale, pretraining, or data. The scripts make the task runnable, the source and baseline implementations anchor the comparison, and the scoring matrix defines progress as weighted improvement over the reproduced source method.}
    \label{fig:env-example}
\end{figure}

\paragraph{Example.} \figref{fig:env-example} shows one environment end to end. The paper's question survives intact, whether a transformer can train without normalization, while its experiment shrinks to ImageNet-50 so attempts fit a bounded budget. The constraint is what keeps the task honest: the same model scale, data, and pretraining as the source method, so any gain must come from the method itself. Half of every evaluation stays hidden, so the score reflects what generalizes rather than what was tuned against the visible half.

\newpage

\begin{wrapfigure}{r}{0.36\textwidth}
    \vspace{.38cm}
    \vspace{-15pt}
    \centering
    \includegraphics[width=\linewidth]{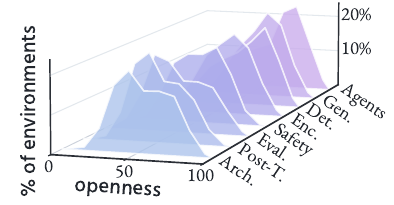}
    \vspace{-16pt}
    \caption{\textbf{Environments' openness.} From fully prescribed (0) to fully open (100).}
    \label{fig:openness}
    \vspace{-8pt}
\end{wrapfigure}
\vspace{-.1cm}
\paragraph{Openness.} Research rarely hands the scientist an approach. Deciding what to build is much of the work, and a world that always prescribes it would train executors rather than scientists. We therefore judge each task's materials and place it on a 0 to 100 scale, where 0 means the approach is fully prescribed and 100 means the task fixes only what counts as success and leaves the way there entirely to the scientist. We find that every area covers nearly the whole range (\figref{fig:openness}): the level of freedom of an environment depends on the task itself, not on where in the world it sits.

\vspace{.10cm}

\begin{figure}[!h]
\vspace{-4pt}
\begin{minipage}[c]{0.56\textwidth}
    \vspace{.5cm}
    \centering
    \includegraphics[width=\linewidth]{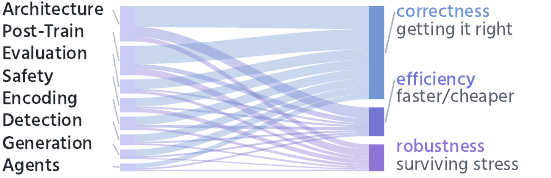}
    \captionsetup{justification=raggedright, singlelinecheck=false, margin={0pt, 26pt}}
    \captionof{figure}{\textbf{What the environments ask for.} Each task names, in its own words, what should get better.}
    \label{fig:work-flow}
\end{minipage}\hfill
\begin{minipage}[c]{0.43\textwidth}
    \paragraph{Goals.} Each environment sets its own goal. Most tasks seek correctness (for example, pushing accuracy), and the rest split mostly between efficiency and robustness, with many naming more than one (\figref{fig:work-flow}). No goal belongs to one area alone: correctness leads everywhere, efficiency appears from architecture to agents, and robustness reaches well beyond safety. The full chain from areas through goals to the work they take is shown    in \figref{fig:ask-flow}.
\end{minipage}
\end{figure}

\clearpage

\begin{figure}[t]
    \centering
    \includegraphics[width=\linewidth]{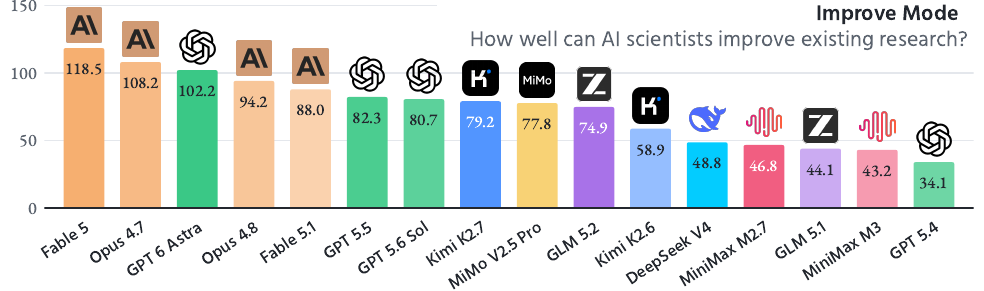}\\[-0pt]
    \includegraphics[width=\linewidth]{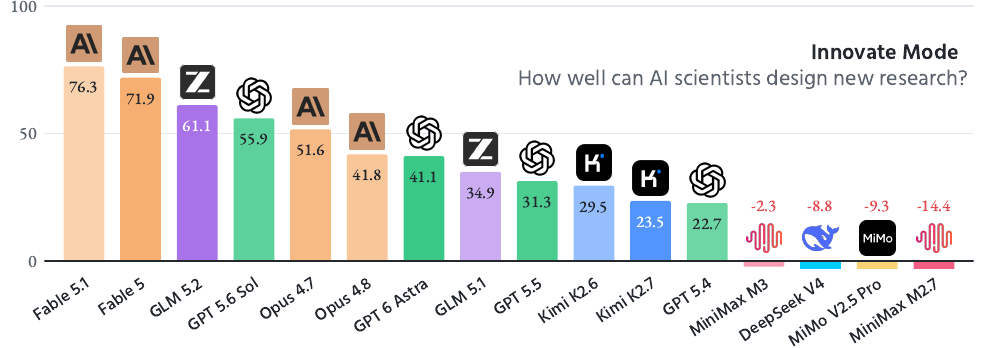}
    \vspace{-.5cm}
    \caption{\textbf{Leaderboards.} Mean \metric by model in improve mode (top) and innovate mode (bottom), averaged over visible and held-out splits. Zero matches the reproduced source method; higher is better. Improve mode gives models the source method, while innovate mode requires them to design from scratch.  We use {\small \texttt{xhigh}} effort for all models.}
    \label{fig:leaderboards}
    \vspace{-.3cm}
\end{figure}

\section{Scientists in the World}
\label{sec:scientists}
% \vspace{-.2cm}

% With the environments built, we can evaluate frontier models as scientists. Each model works on the same tasks under the same compute budget, either by improving the source method or by designing a new method from scratch. We first report their performance, then analyze where the differences come from. Finally, we use these environments to begin training stronger scientists.
% With the environments built, we put frontier \aiscis to work inside them. We score each attempt against the reproduced result, keep everything the attempt leaves behind, and train the next scientists.

% \vspace{-.2cm}

\subsection{Benchmark}
\label{sec:benchmark}

\vspace{-.2cm}

We now use the environments to evaluate \aiscis. Running every model on all 1{,}000 environments is too costly to repeat, so we evaluate on a fixed cohort of ten lower-compute environments. Each model attempts every environment under the same compute budget, in two modes:

\vspace{-.1cm}
\begin{itemize}[topsep=0pt, itemsep=2pt, parsep=0pt]
\item \textbf{Improve mode.} The model receives the reproduced source method and tries to improve its result.
\item \textbf{Innovate mode.} The model receives no source code and must design a method from scratch. We ensure there is no leakage of the paper's original solution.
\end{itemize}
\vspace{-.1cm}

Each attempt is evaluated on both the visible and held-out splits, with the final \metric score averaged across the two. Zero \metric matches the reproduced source method, while a positive \metric improves on it. Since different environments have different  scoring scales, we normalize the scores to aggregate (\secrefc{sec:eval-scoring}).

In the result, the two modes show different strengths across models (\figref{fig:leaderboards}): \model{Fable~5} is the strongest at improving existing methods, while \model{Fable~5.1} and \model{Fable~5} take the top two places at designing new methods from scratch. Among open models, \model{GLM 5.2} comes closest to the frontier, placing third at innovating. 
% The rankings also differ between the two modes, suggesting that improvement and innovation are related but distinct abilities. 
Notably, the \aiscis beat the paper's reproduced method in 84\% of improve and 74\% of innovate attempts (\secrefc{sec:success-rate}).

\clearpage

\begingroup

\setlength{\columnsep}{22pt}

% \begin{wrapfigure}[22]{l}{0.52\linewidth}
\begin{wrapfigure}[22]{l}{0.48\linewidth}
    \centering
    \vspace{.8cm}
    \vspace{-\baselineskip}
    \includegraphics[width=1.0\linewidth]{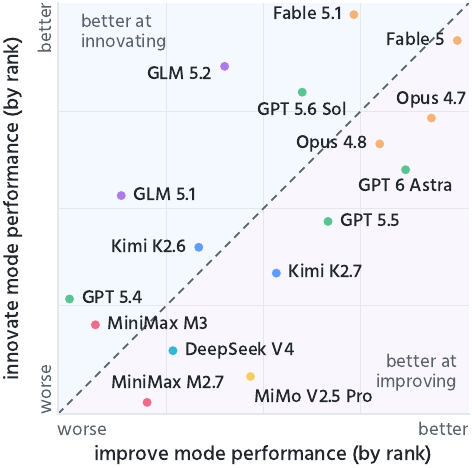}
    \caption{\textbf{The two abilities are loosely tied.} Each model's rank at improving against its rank at innovating.}
    \label{fig:innovate-improve}
\end{wrapfigure}

% The leaderboards say who wins, but not why. To understand the ranking, we look past the final score at what each attempt actually did: how much it spent, and how good the underlying research was.
The leaderboards show which models score highest in \metric score, but not what kind of ability it reflects. We first compare the two modes directly. If improving an existing method and inventing a method from scratch measured the same skill, models would keep similar ranks across the two leaderboards. Instead, the relationship is loose (\figref{fig:innovate-improve}). Some models are much stronger when building on source code, while others rank higher when forced to design from scratch. We therefore treat improve and innovate as related but distinct research abilities.

\vspace{.2cm}

\textbf{Is performance just a scaling effect?} Better results could come from scaling the attempt in two ways: using more external compute, or doing more model-side work through longer trajectories and more trial-and-error. We do not find either pattern. Compute hours and agent turns explain almost none of the variance in scores ($R^2 \leq 0.02$, \figref{fig:radar-effort}a; turns are our provider-agnostic proxy for model-side effort, since token counts may not be 
% consistently 
available). The leading models therefore do not win by simply spending more; they make better use of the same setting.

\vspace{.38cm}

\textbf{Then what drives it?} The quality of the research itself. An LLM judge reads every attempt, blind to its score, and rates it on five properties: \textit{creativity} (was the idea genuinely new, and did it pay off), \textit{engineering} (was the implementation substantial, building on the existing code), \textit{rigor} (was the reported gain real and honestly measured), \textit{significance} (how far it moved the metric), and \textit{depth} (how thoroughly the problem was explored). These ratings track the models' actual scores almost exactly ($\rho = 0.89$, \figref{fig:radar-effort}b): the models that win are the ones doing better research, not the ones trying hardest.

\begin{figure}[h]
    % \hfill
    \vspace{.38cm}
    \centering
    \begin{minipage}[c]{0.5664\linewidth}
        \includegraphics[width=\linewidth]{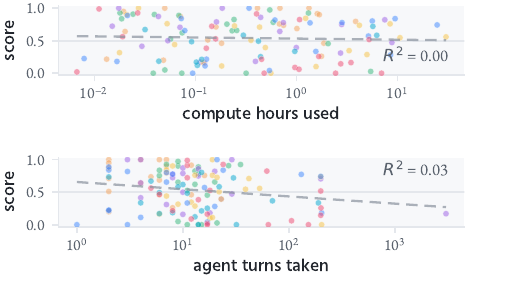}
    \end{minipage}%
    \begin{minipage}[c]{0.4336\linewidth}
        % 10pt raises panel (b); edit in place, negative lowers it
        \raisebox{5pt}[\height][\depth]{\includegraphics[width=\linewidth]{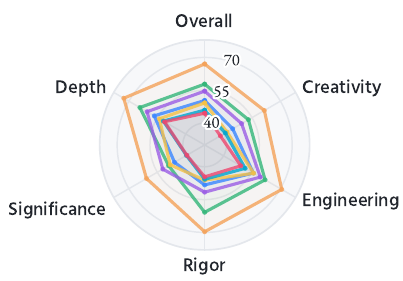}}
    \end{minipage}\\[2pt]
    \makebox[0.53\linewidth][c]{\small (a) effort versus outcome}%
    \makebox[0.47\linewidth][c]{\small (b) research quality}\\[1pt]
    \includegraphics[width=0.92\linewidth]{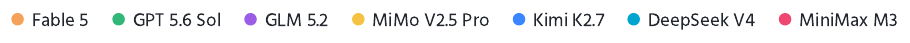}
    \vspace{-.15cm}
    \caption{\textbf{Effort versus outcome, and research quality.} (a) each run's score against the compute hours (top) and agent turns (bottom) behind it, pooling both modes; effort explains almost none of the variance in score ($R^2 = 0.00$, $0.02$). (b) research quality by dimension, judged apart from the score. Newest model per family, colored by provider.}
    \label{fig:radar-effort}
\end{figure}

\newpage

\textbf{Research is contingent.} Research is inherently random. A strong attempt does not always produce a high score: a creative idea can be implemented well and still fail empirically, and a great idea might not suit the current setting, while a simpler idea may happen to work. 
% We therefore read each score as the realized outcome of one research trajectory, not as a complete judgment of the effort behind it. 
Aggregating across many environments
% , and comparing scores with blind quality ratings, 
lets us measure progress without pretending that research has no outcome variance.

% \textbf{Why does no single model win everywhere?} Two reasons. The modes reward different strengths (\figref{fig:innovate-improve}): improving rewards engineering on top of working code, while innovating rewards the creativity to design a method from nothing, so a model tends to lead whichever mode matches its stronger property. And research is noisy: a promising idea can fail to land on a given problem, so even \model{Fable~5}, which the judge rates highest almost everywhere, does not top every board. Doing good work and getting a good result are related, but not the same.

\endgroup

\vspace{.2cm}

\subsection{Dataset}
\label{sec:dataset}

\begin{wraptable}[11]{r}{0.29\linewidth}
\vspace{-\baselineskip}
\vspace{-.43cm}
\centering
\begin{softcolortable}[1]
\setlength{\tabcolsep}{4pt}%
\setlength{\softTableContentWidth}{\dimexpr\softTableWidth-4\tabcolsep\relax}%
\renewcommand{\arraystretch}{1.3}%
\begin{tabular}{
    >{\leftskip=4pt \raggedright\arraybackslash}m{0.66\softTableContentWidth}
    >{\raggedleft\arraybackslash}m{0.34\softTableContentWidth}
    }
\toprule
\rowcolor{datasetHeader}
\textbf{record} & \textbf{average} \\
\midrule
\rowcolor{gray!8} idea revisions & 3.7 \\
\rowcolor{white} method revisions & 6 \\
\rowcolor{gray!8} experiments run & 25 \\
\rowcolor{white} commits & 35 \\
\rowcolor{gray!8} hours of work & 5.4 \\
\bottomrule
\end{tabular}
\end{softcolortable}
\captionof{table}{\textbf{Dataset statistics.} 
% Mean over completed attempts.
Each experiment is stored with its log; hours of work is the scientist's end-to-end running wall-clock time.}
\label{tab:dataset}
\end{wraptable}

We preserve every scientist's research  in full, so 
% \sw
we collect a set of research traces, 
% We store each attempt as a version-controlled repository
released per environment, model, and run mode. We record the scientist's research idea and their revisions, the method and the full turn-by-turn trajectory, and the per-experiment logs and scores (\tabref{tab:dataset}).
% on the visible and held-out splits .
We store each attempt as a version-controlled repository, 
% incrementally over each scientist's run, 
so each preserves the complete development history rather than the final state alone.
% and we keep attempts that regress on the source method alongside those that improve on it. 

\vspace{.03cm}

In total, the corpus records roughly 6{,}400 hours of research being done (\secrefc{sec:trace-corpus}).
With multiple scientists' attempts, one environment accumulates multiple independent solutions to a fixed problem, each labeled with its measured improvement over the reproduced source result. 
% We use the resulting corpus both as training data for stronger scientists and as a basis for analyzing model behavior across attempts.

\vspace{.2cm}

\subsection{Gym}
\label{sec:gym}

Finally, we raise a scientist of our own. We train \model{Qwen3-8B}~\citep{qwen3}, a scientist small enough that whatever it learns must come from the world, with Tinker~\citep{tinker}, in two stages:

\vspace{-.1cm}

\begin{wrapfigure}[12]{r}{0.48\textwidth}
    \vspace{-12pt}
    \centering
    \includegraphics[width=\linewidth]{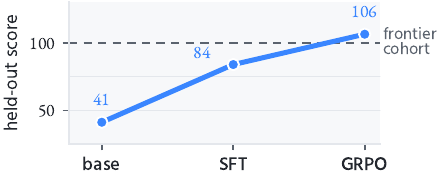}
    \vspace{-14pt}
    \caption{\textbf{Raised in the world.} Median held-out \metric of each stage's ideas, implemented identically; after GRPO it reaches the frontier cohort's anchor.}
    \label{fig:gym-training}
    \vspace{-8pt}
\end{wrapfigure}

\begin{itemize}[topsep=6pt, itemsep=5pt, parsep=0pt, leftmargin=*]
\item \textbf{Supervised fine-tuning.} 1{,}000 examples drawn from the trajectories, each pairing a task specification with an attempt's reasoning and final idea, cleaned so that no fabricated result survives.  (details in \secrefc{sec:training-details})
\item \textbf{GRPO.} We do off-policy group relative policy optimization (GRPO)~\citep{shao2024deepseekmath} on 1{,}500 already-scored ideas, rewarded by measured outcomes.
% : the held-out improvement, plus a generalization term that discounts overfit gains. 
Advantages are relative to the other ideas for the same task, and updates are clipped to a trust region around the SFT policy. 
% No judged text enters the reward; t
The scientist trains only with execution.
\end{itemize}
\vspace{.03cm}

A scientist this small cannot yet carry the whole loop: implementing and debugging a method over a long horizon is frontier work in itself. We therefore separate ideation from implementation \citep{si2026execution}. The trained scientist proposes the idea, and a single fixed frontier implementer (\model{GPT 5.6 Sol}) executes every idea under identical conditions on ten held-out papers (\secrefc{sec:idea-ablation}), so only the idea's origin varies. We prompt the implementer to follow exactly the ideator's idea.

% Scoring is the benchmark's own: zero matches the paper's reproduced method, and the frontier cohort anchors near one.

\vspace{.03cm}

The gain is monotone. The median implemented idea scores 41 \metric from the base scientist, 84 after SFT, and 106 after GRPO (\figref{fig:gym-training}); the mean rises from 51 to 90. 
% By the median environment, i
Ideas from the trained scientist, faithfully implemented, reach the anchor of the frontier cohort. What it demonstrates is the mechanism: the world supplied the data, the reward, and the held-out exam, and a scientist raised on its trajectories proposes measurably better research. The training loop works end to end.
% , and it grows with the world.

\newpage

\section{Background}

\paragraph{Autonomous scientific discovery and \aiscis.}
Automating discovery is a long-standing goal. Early systems recovered natural laws and governing equations directly from data \citep{schmidt2009distilling, cranmer2020discovering, udrescu2020ai}, and robot scientists closed the loop from hypothesis to physical experiment \citep{king2004robot, king2009automation, williams2015cheaper}. Within machine learning, automated search took over bounded parts of the pipeline: model and hyperparameter selection \citep{thornton2013autoweka, feurer2015efficient}, neural architecture design \citep{zoph2017neural, real2019regularized}, and even the discovery of faster fundamental algorithms \citep{fawzi2022discovering, mankowitz2023faster}. Program search guided by language models extended this to open mathematical and algorithmic problems \citep{romeraparedes2024mathematical, novikov2025alphaevolve}. Language models then widened the scope toward the full workflow: tool-augmented systems run their own chemical experiments \citep{boiko2023autonomous, bran2024augmenting}, and agentic pipelines generate ideas, write code, and draft papers end to end \citep{lu2024aiscientist, yamada2025aiscientist}. A newer wave organizes many agents into collaborative or self-directed discovery loops \citep{schmidgall2025agent, schmidgall2025agentrxiv, ghareeb2025robin}, and general-purpose \aiscis now attempt discovery across whole fields \citep{gottweis2025accelerating, mitchener2025kosmos}. This line builds better scientists; \sw is complementary, turning finished papers into environments where such scientists can be handed the same problem and scored on it.

\paragraph{Measuring \aiscis.}
Evaluating agents in executable settings began with general benchmarks for tool use and repository-level code editing \citep{jimenez2024swebench, liu2023agentbench, mialon2024gaia}, then moved to machine-learning experimentation and engineering \citep{huang2024mlagentbench, chan2024mle, wijk2024rebench, nathani2025mlgym}. A parallel line probes scientific reasoning and data-driven discovery \citep{jansen2024discoveryworld, majumder2024discoverybench, chen2024scienceagentbench}, extending to hypothesis formation and frontier domain research \citep{liu2025researchbench, miao2026prlbench}. A closely related group scores whether an agent can reproduce a paper's results from its description \citep{siegel2024corebench, starace2025paperbench}, reimplement its method and code \citep{xiang2025scireplicate, hua2025researchcodebench}, or take on research framed as an open-ended challenge \citep{zhang2025mlrcbench, chen2025mlrbench}. These suites are almost always fixed and hand-curated, and critiques of agent evaluation warn that they can confound cost, leak through weak holdouts, and reward shortcuts \citep{kapoor2024aiagents}. Concurrent work also turns real papers into research tasks but curates them by hand, which does not scale \citep{garikaparthi2026researchgym}. \sw instead treats construction itself as the object: it selects papers at scale and refactors each into a scored, reproduce-then-improve environment, so a method is judged by its gain over a faithfully reproduced result rather than its distance from a reported number.

\section{Conclusion}

Every published paper with code is a finished piece of research: a problem, a method, and a measured result. \sw turns that record back into a place where research happens: 1{,}000 papers become runnable, scored environments, and the collection grows with the literature rather than with annotation effort. The same environments measure today's \aiscis and train the next ones; each is a scaled-down paper and each score one contingent trajectory, but the loop they are built for, a scientist that proposes, executes, and updates inside the world, is just beginning. As the models and the literature both move, the world keeps pace, and the problems it poses stay as open as the research it was built from.

\newpage

\subsection*{Acknowledgments}
We thank Muhan Gao, Linrong Cai, Xinshuang Liu, and Haozhe Chen for helpful discussions and feedback.
The authors are pleased to acknowledge that the work reported on in this paper was substantially performed using the Princeton Research Computing resources at Princeton University which is a consortium of groups led by the Princeton Institute for Computational Science and Engineering (PICSciE) and Research Computing. 
% \commentone{Depends on your actual resource usage.}

\bibliographystyle{plainnat}
\bibliography{main}

\newpage

\section*{\LARGE Appendix}

This appendix details the \builder system that constructs \sw environments (Appendices \hyperref[sec:infrastructure]{A}--\hyperref[sec:agent-workspace]{C}), how papers are selected (Appendix \hyperref[sec:paper-selection]{D}), and evaluates scientist agents on them (Appendices \hyperref[sec:evaluation-protocol]{E}--\hyperref[sec:experimental-setup]{G}), the dataset and training details (Appendix \hyperref[sec:training-details]{H}), followed by a worked example and the agent prompts (Appendices \hyperref[sec:example-env]{I}--\hyperref[app:prompts]{J}).

\appendix

\vspace{0.9cm}
\begingroup
\setlength{\parindent}{0pt}

% --- local table-of-contents line helpers (label, number, title) ---
\newcommand{\tocsec}[3]{%
  \addvspace{9pt}%
  \noindent{\large\bfseries\hyperref[#1]{\makebox[1.8em][l]{#2}#3}}%
  \hspace{0.5em}\dotfill\hspace{0.5em}%
  {\large\bfseries\hyperref[#1]{\pageref{#1}}}\par}
\newcommand{\tocsub}[3]{%
  \addvspace{2.5pt}%
  \noindent\hspace{1.8em}\hyperref[#1]{\makebox[2.6em][l]{#2}#3}%
  \hspace{0.5em}\dotfill\hspace{0.5em}%
  \hyperref[#1]{\pageref{#1}}\par}
\newcommand{\tocsubsub}[3]{%
  \addvspace{1pt}%
  \noindent\hspace{4.4em}{\small\hyperref[#1]{\makebox[3.0em][l]{#2}#3}%
  \hspace{0.5em}\dotfill\hspace{0.5em}\hyperref[#1]{\pageref{#1}}}\par}

{\Large\bfseries Table of Contents}

\vspace{0.1cm}

\tocsec{sec:infrastructure}{A}{System Design}
\tocsub{sec:hive}{A.1}{Hive: Interactive Compute Pool}
\tocsub{sec:nexus}{A.2}{Nexus: Experiment Boundary}
\tocsub{sec:swarm}{A.3}{Swarm: Builder Coordination}
\tocsubsub{sec:swarm-milestones}{A.3.1}{Milestones}
\tocsubsub{sec:swarm-roles}{A.3.2}{Builder Roles}
\tocsubsub{sec:swarm-workflow}{A.3.3}{Workflow}
\tocsub{sec:worker}{A.4}{Worker: Agent Interface}
\tocsub{sec:layer-handoffs}{A.5}{How the Components Connect}

\tocsec{sec:agent-pipeline}{B}{Builder Design}
\tocsub{sec:agent-workflow}{B.1}{What the Builders Do}
\tocsub{sec:design-choices}{B.2}{Design Choices}

\tocsec{sec:agent-workspace}{C}{Agent Workspace}
\tocsub{sec:ws-structure}{C.1}{Structure}
\tocsub{sec:ws-scoring}{C.2}{Problem and Scoring}
\tocsub{sec:ws-execution}{C.3}{Execution}
\tocsub{sec:ws-records}{C.4}{Records and Validation}

\tocsec{sec:paper-selection}{D}{Selecting   Papers}

\tocsec{sec:evaluation-protocol}{E}{Evaluation Protocol}
\tocsub{sec:eval-modes}{E.1}{Evaluation Modes}
\tocsub{sec:eval-trial}{E.2}{Running a Trial}
\tocsub{sec:eval-scoring}{E.3}{Scoring}

\tocsec{sec:analysis}{F}{Evaluation Analysis}
\tocsub{sec:success-rate}{F.1}{How Often Scientists Beat the Paper}
\tocsub{sec:generalization}{F.2}{Do the Gains Generalize}
\tocsub{sec:effort}{F.3}{Effort and Outcomes}

\tocsec{sec:experimental-setup}{G}{Experimental Setup}
\tocsub{sec:exp-data}{G.1}{Data}
\tocsub{sec:exp-compute}{G.2}{Compute}
\tocsub{sec:exp-scaffolds}{G.3}{Agent Scaffolds}
\tocsub{sec:exp-models}{G.4}{Models}
\tocsub{sec:exp-trials}{G.5}{Evaluation Trials}
\tocsub{sec:exp-failures}{G.6}{Provider Failures}
\tocsub{sec:exp-cohort}{G.7}{Cohort and Normalization}
\tocsub{sec:exp-reproducibility}{G.8}{Reproducibility}

\tocsec{sec:training-details}{H}{Dataset and Training Details}
\tocsub{sec:trace-corpus}{H.1}{Trace Corpus}
\tocsub{sec:training-data}{H.2}{Training Data}
\tocsub{sec:training-procedure}{H.3}{Training Procedure}
\tocsub{sec:idea-ablation}{H.4}{Held-Out Evaluation}

\tocsec{sec:example-env}{I}{Example Produced Environment}

\tocsec{app:prompts}{J}{Agent Prompts}
\tocsub{app:prompts-selection}{J.1}{Prompts for Paper Grading and Selection}
\tocsub{app:prompts-swarm}{J.2}{Prompts for Swarm}
\tocsub{app:prompts-eval}{J.3}{Prompts for Evaluation}

\vfill
\endgroup
\newpage

\section{System Design}
\label{sec:infrastructure}

\begin{figure}[h]
    \centering
    \includegraphics[width=\textwidth]{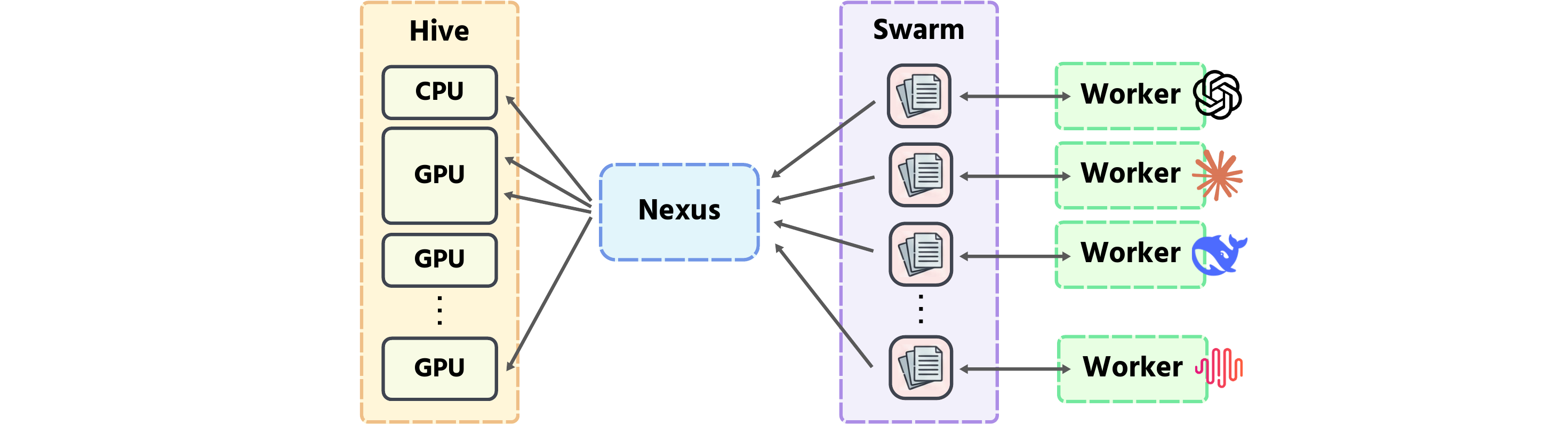}
    \caption{\textbf{System design overview.} Hive supplies execution capacity for CPU and GPU jobs; Nexus accounts for the budget and resource boundary of each attempt; Swarm maintains one paper-level trajectory across roles, restarts, and handoffs; and Workers connect different agent scaffolds through the same system interface. Together, these layers let many agents iterate on open-ended papers while preserving the evidence, budget state, and handoff context.}
    % needed  to judge the final artifact.}
    \label{fig:infra-stack}
\end{figure}

 \sw relies on agents that repeatedly alternate between reasoning and computation. This agent stack is the substrate that makes this loop practical at scale. It separates four concerns that are otherwise entangled: running compute jobs, bounding each experiment's resource use, coordinating many papers, and normalizing different agent interfaces. We separate the system into four parts: Hive provides the compute pool; Nexus makes compute accountable to a particular attempt; Swarm coordinates many runs; and Worker presents each agent scaffold through the same interaction contract. \figref{fig:infra-stack} summarizes these layers.

\subsection{Hive: Interactive Compute Pool}
\label{sec:hive}

% \paragraph{Motivation.} 
Agent-driven reproduction needs a compute layer designed for interactive iteration. An agent may run a small debugging command, inspect the result, edit the code, and then launch a longer training or evaluation job. If every such step pays the full overhead of acquiring resources and preparing execution, the research loop becomes dominated by infrastructure latency rather than by scientific reasoning. Hive keeps compute capacity available, monitored, and controlled, so these iterations remain close to an interactive workflow.

\begin{figure}[h]
    \centering
    \includegraphics[width=\textwidth]{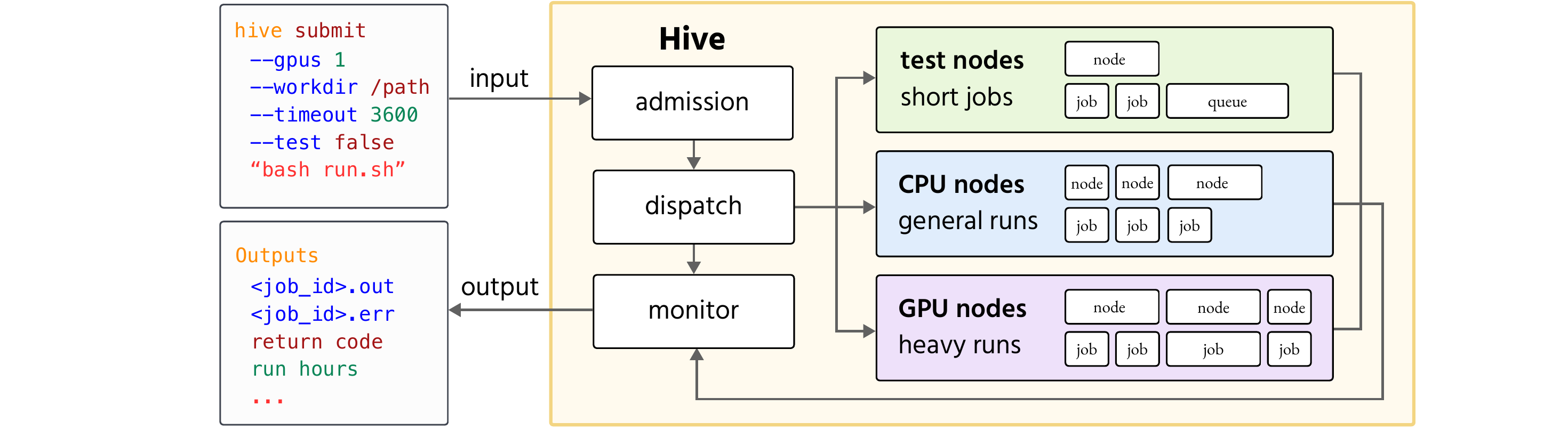}
    \caption{\textbf{Hive execution overview.} A job request enters Hive with a command, resource target, working directory, and time limit. Admission queues the request, dispatch routes it to test, CPU, or GPU capacity, and monitor records the return code, runtime, and logs that form the job record.}
    \label{fig:hive-system}
\end{figure}

Hive exposes compute as an experimental action: given a requested experiment, declared resources, and a workspace, it reports what happened. The result is not only a terminal success or failure label. It includes the execution evidence the agent needs to decide what to do next: logs, completion status, timing, and partial output when a job is still running. This matters because reproduction agents rarely need a single monolithic training run; they work through a trial and error sequence of tests, attempts, fixes, ablations, and evaluations. \figref{fig:hive-system} shows the execution path inside Hive.

\paragraph{Input.} Hive's input is a job request. In our interface, the concrete entry point is a terminal command: \texttt{hive submit}, whose parameters specify what should run. \texttt{--workdir} anchors the run to the code and files it should see, \texttt{--gpus} and \texttt{--cpus} declare the requested capacity, \texttt{--resource} targets a named pool such as test, CPU, or GPU capacity, and \texttt{--timeout-sec} bounds jobs that hang or run past their intended scale. Optional fields such as retry count and log directory make the request easier to recover and inspect. These fields are useful because they turn an open shell command into a schedulable execution unit: Hive can decide where it can run, how much capacity it needs, when it should stop, and where its evidence should be recorded.

\newpage

\paragraph{Output.} Hive returns a job identifier and maintains a job record, not an experiment-specific result file. The record directory contains \texttt{meta.json}, \texttt{stdout.log}, and \texttt{stderr.log}: metadata for status, timing, resource use, and return code, plus streamed logs for diagnosis. Any checkpoints, metrics, or plots created by the command remain in the working directory; Hive records how the execution went.

\paragraph{Elasticity.} A job asks for a category of resource (test, CPU, or GPU) and Hive places it on whatever node is free, without the agent naming a specific machine. The pool can grow or shrink as capacity becomes available, so the first worker to come online is useful immediately, and the same request format serves every experiment.

\paragraph{Node Types.} Hive exposes three node types through the same job interface. CPU nodes support CPU-only experiments, setup, preprocessing, and other jobs that do not require accelerators. GPU nodes run accelerator-heavy experiments such as training and large-scale evaluation. Test nodes are for quick code validation and small-scale testing, with jobs capped at one hour so agents do not tie up long-running allocations.

% \paragraph{Scope.} Hive is deliberately not the benchmark policy layer. It does not decide which paper should receive compute, how much budget a paper has, or whether a claimed result is sufficient. Those decisions are pushed upward to Nexus and Swarm. This separation keeps the compute layer reusable while preventing it from accumulating paper-specific semantics.

\paragraph{Reliability.} Long-running jobs are not perfectly reliable: a job can stall or be killed before it finishes, and its partial output is often still useful. Hive therefore treats an interruption as a reportable outcome rather than a silent failure: it writes the job's last state and logs into the job record, then requeues it on another valid machine, where the command restarts from the beginning; the interrupted attempt is recorded for inspection like any completed one, and its incomplete work is not charged against the budget.

\paragraph{Implementation.} We use Ray~\citep{moritz2018ray} to implement Hive. Ray starts each job on a free machine and runs it; everything else the compute pool needs for agent-driven reproduction is built on top. A resource manager tracks CPU, GPU, and test capacity; a dispatcher turns agent requests into queued jobs; an agent-facing client/API submits jobs and streams progress back; and a persistent job store records logs, status, and final outputs outside the Ray runtime. These pieces are what make the compute pool usable as research infrastructure, not just a distributed execution engine.

\newpage

\subsection{Nexus: Experiment Boundary}
\label{sec:nexus}

\begin{figure}[h]
    \centering
    \includegraphics[width=\textwidth]{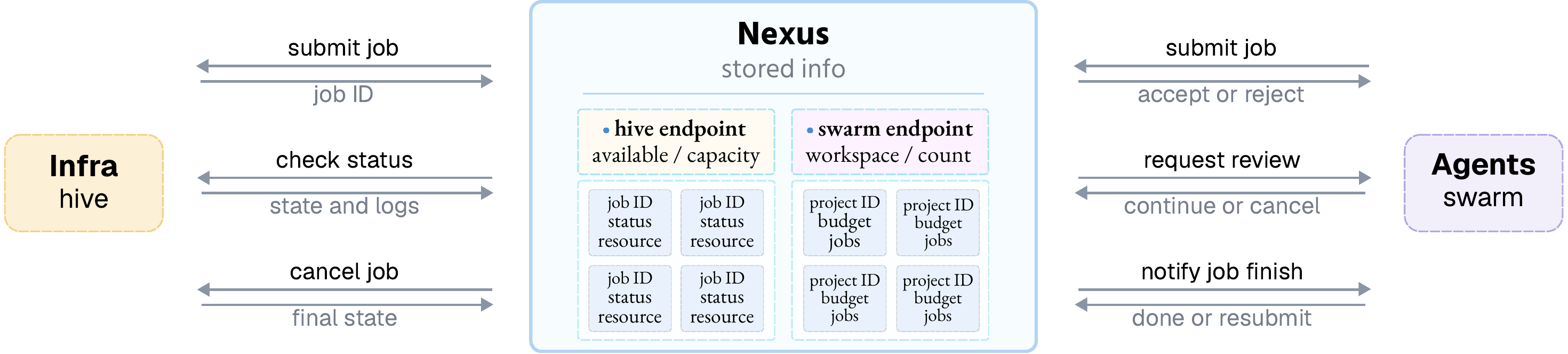}
    \caption{\textbf{Nexus operations overview.} Nexus stores the job, project, budget, and review state needed to translate agent requests into Hive actions and to translate Hive execution evidence back into agent decisions.}
    \label{fig:nexus-system}
\end{figure}

% \paragraph{Boundary.} 
Nexus is the experiment boundary that manages the resources agents use. It is a bridge between agents and compute. An autonomous agent should not be trusted to enforce its own compute budget simply because the prompt tells it to be careful. Nexus makes the budget a property of the system. Each reproduction attempt receives a bounded project: jobs must fit within its resource limits, its concurrent usage cap, and its remaining GPU-hour budget before they are allowed to run. Compute is no longer something the agent can spend freely; every job is counted against the project's budget. The operations view in \figref{fig:nexus-system} shows these exchanges: agents submit or review jobs through Nexus, and Nexus communicates between infrastructures and agents.

\begin{wrapfigure}[9]{r}{0.5\linewidth}
    \centering
    \vspace{-0.92cm}
    \includegraphics[width=\linewidth]{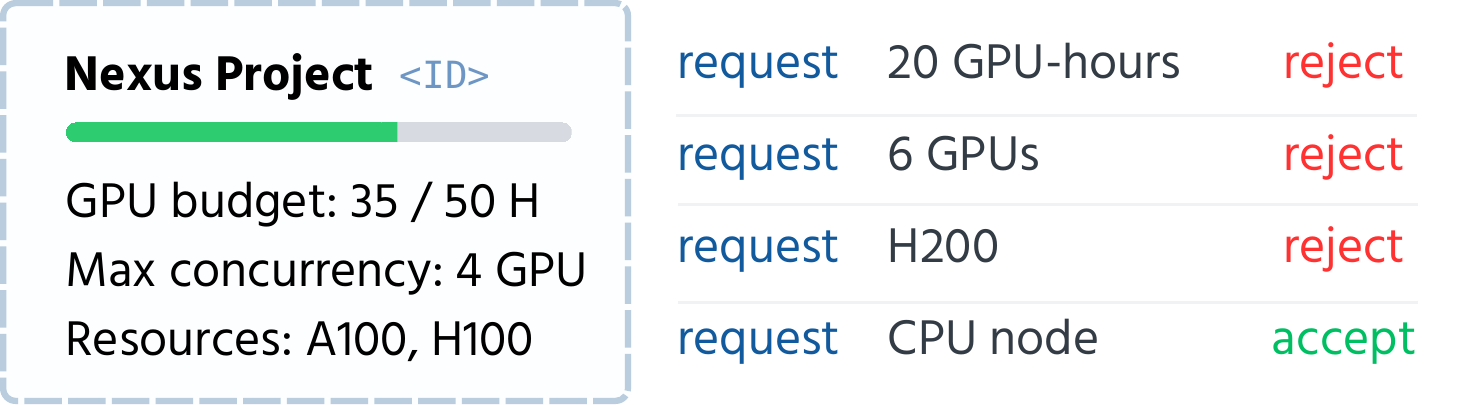}
    \caption{\textbf{A Nexus project turns resource policy into explicit accept/reject decisions.} The example shows how budget, concurrency, and allowed resource pools gate agent requests before they reach Hive.}
    \label{fig:nexus-project}
\end{wrapfigure}

\paragraph{Quota Semantics.} Nexus enforces several constraints that matter for reproducibility: how much compute a paper can spend in total, how much it can hold at once, which kinds of resources it may use, and how long any single job may run before it times out. At the job level, these limits cover CPU, GPU, memory, and time requests. These constraints are checked outside the agent. The project-level example in \figref{fig:nexus-project} makes the gate concrete: a request can be rejected because it exceeds the remaining budget or concurrency capacity, or requests a resource pool outside the given policy.
% The agent can choose an experimental strategy, but the system decides whether that strategy is still within the paper's budget.

\paragraph{Fairness.} This accounting matters for both fairness and interpretability. Without it, one agent can consume resources in ways that alter the conditions for others, and the resulting comparison would conflate scientific difficulty with uncontrolled resource use. With Nexus, the amount of compute available for a single reproduction is explicit and enforceable. The agent can still decide how to reasonably spend its budget, but it cannot silently exceed the given total.

\paragraph{Periodic Review.} Nexus makes long jobs interruptible without requiring the agent to keep an active turn open for the whole execution. At each configured review interval, Nexus snapshots the latest logs, records them in the workspace, and wakes the responsible agent to inspect them. The agent can then continue the run, cancel it, or change strategy before spending the remaining budget. Interval review pays off because many broken reproductions are visible well before natural termination: imports fail after setup, data paths are wrong, losses diverge, or an evaluator writes clearly invalid output. A run that would otherwise consume hours can stop after the first snapshot that shows it is going nowhere. If the agent does not respond within the review window, Nexus continues the job rather than leaving compute blocked by a missing response.

% \paragraph{Evidence.} Nexus also defines what it means for compute evidence to belong to an experiment. Job outputs are mirrored into the workspace, so the agent and later reviewers can inspect what happened without relying on a transient service state. Completed jobs release budget; interrupted jobs can be recovered; budget exhaustion is visible to the agent rather than appearing as an unexplained failure. The goal is not only to avoid lost work, but also to make the accounting auditable after the fact.

\paragraph{Separation.} This layer is intentionally separate from Hive, and can accept any compute infrastructure. Nexus is optimized for accountability. Keeping them separate prevents the compute pool from accumulating task-specific policy and prevents Swarm from having to reason about low-level execution state.

\subsection{Swarm: Builder Coordination}
\label{sec:swarm}

\begin{figure}[h]
    \vspace{-.2cm}
    \centering
    \includegraphics[width=\textwidth]{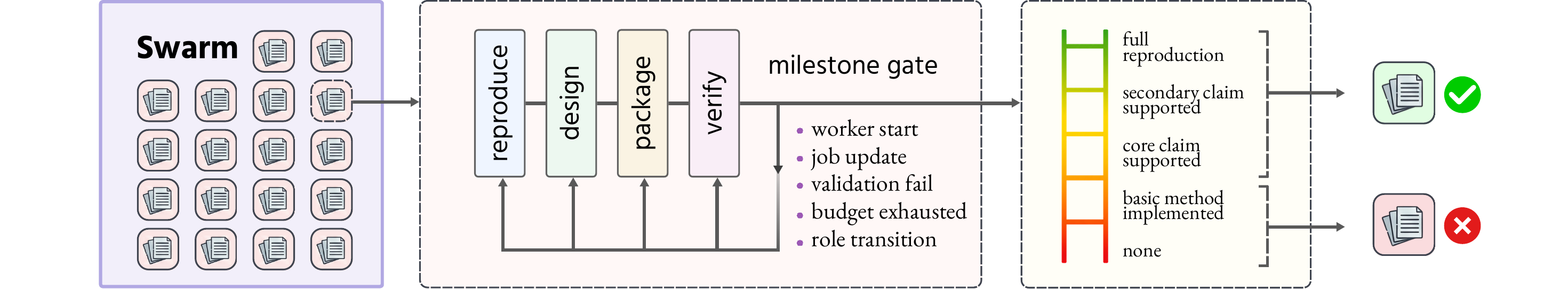}
    \caption{\textbf{Swarm coordination overview.} Swarm drives a pool of paper workspaces in parallel. Each paper moves through an ordered role workflow---reproduce, design, package, verify---whose transitions are driven by wake events: worker start, job submit, job update, validation failure, and role transition. Progress is graded against a milestone ladder, from \texttt{none} up to full reproduction, and the milestone gate decides whether a paper's artifact is accepted.}
    \label{fig:swarm-overview}
    \vspace{-.5cm}
\end{figure}

% \paragraph{Motivation.}
Swarm has two jobs. The first is per paper: it drives one reproduction across many agent turns and several roles, keeping the attempt coherent as \builders hand off, retry, and audit each other's work. The second is across papers: it runs many of these attempts at the same time, each as its own trajectory, without their states blurring together. Both are hard because the work is open-ended. A reproduction can produce useful code without supporting the paper's central claim, support the claim but leave no reusable environment, or produce scores that look plausible until a later audit checks them. If these judgments live only in chat transcripts, the result is a collection of anecdotes. Swarm therefore turns each paper attempt into an auditable trajectory: the assigned agent, budget, workspace state, results, milestone, and next role are explicit at every step. \figref{fig:swarm-overview} shows the end-to-end path: a pool of paper workspaces, each driven through an ordered role workflow under a milestone gate, ending in an artifact that is either accepted or set aside.

\paragraph{Pipeline.} Swarm turns a paper into a finished environment. Its \builders reproduce the paper's main claim, design the reproduction into a usable task, organize its files, and verify the result. Swarm does not run this in one long session. A \builder works in short turns, and ends each turn with an action, usually submitting a compute job or signaling that it is done. Swarm tracks that action and the paper's state, and whenever something changes it wakes the \builder with a prompt written for that exact situation: for example, when a compute job finishes, a prompt that carries the results and asks what to do next; when a build fails, a prompt with the build log and a request to fix it; when one stage is complete, a prompt that briefs the agent for the next. The agent is therefore only ever asked to decide the one thing that just came up. Swarm runs this loop over a pool of many workers, so it advances a large number of papers at the same time.

\paragraph{Coordination.} Swarm answers the question that remains after a single agent turn ends: who should act next, and what do they inherit from the \builder before them? Its unit of state is the paper, not the agent session: Swarm tracks where each paper sits in its workflow and the milestone it has reached, so a new \builder resumes from the right point and can build on a partial result left by the previous one. This is what lets a single paper pass through several \builders, even different models, and still read as one coherent attempt; without it, the final artifact would mix the paper's difficulty with hidden differences in how the agent happened to be run.

\paragraph{Wake Events.} Swarm wakes a \builder only when new information changes what it should decide: a fresh paper or a new role to begin, a running job it may continue or cancel, a finished job, a failed build, an exhausted budget, a validation or milestone-gate failure, or a role transition. The agent returns a structured action, and Swarm records both the action and what prompted it. \tabref{tab:swarm-wake-events} summarizes what is shown.

\begin{table}[h]
    \centering
    \begin{softcolortable}[0.87]
    \setlength{\tabcolsep}{7pt}%
    \setlength{\softTableContentWidth}{\dimexpr\softTableWidth-6\tabcolsep\relax}%
    \begin{tabular}{
        >{\leftskip=9pt \raggedright\arraybackslash}m{0.17\softTableContentWidth}
        >{\raggedright\arraybackslash}m{0.41\softTableContentWidth}
        >{\raggedright\arraybackslash}m{0.42\softTableContentWidth}
        }
    \toprule
    \rowcolor{swarmHeader}
    \textbf{event} & \textbf{shown to agent} & \textbf{kept in trajectory} \\
    \midrule
    \rowcolor{gray!8}
    iteration start & paper, workspace, milestone, budget & model, role, starting checkpoint \\
    \rowcolor{white}
    job running & streaming logs and remaining budget, to continue or cancel & the continue or cancel decision \\
    \rowcolor{gray!8}
    job finished & logs, exit code, and remaining budget & outcome, compute used, next action \\
    \rowcolor{white}
    build failed & the build log, to fix and resubmit & failure cause, fix attempts \\
    \rowcolor{gray!8}
    budget exhausted & budget spent, no further jobs, wrap up current work & final spend, the milestone the agent wraps up at \\
    \rowcolor{white}
    validation fail & validation or milestone-gate failure & resulting status, revert attempts \\
    \rowcolor{gray!8}
    role transition & prior milestone and unresolved issues & next role and inherited state \\
    \bottomrule
    \end{tabular}
    \end{softcolortable}
    \caption{\textbf{Swarm wakes a \builder only at decision-changing events.} Rather than holding a session open for the whole reproduction, Swarm re-engages the \builder only when new information changes what it should do next. Each row gives the triggering event, what the agent is shown, and the state Swarm writes into the paper's trajectory.}
    \label{tab:swarm-wake-events}
    \vspace{-.2cm}
\end{table}

\paragraph{Completion Checks.} When a \builder declares a pass finished, Swarm separates local correctness from progress. The local check confirms the action is parseable and the milestone label is canonical, then runs the workspace validator, which verifies that the score files exist, the evaluator is separated from the method, and the workspace rebuilds from a clean clone (\secrefc{sec:ws-records}). Failures become repair prompts, capped at three validation attempts. The progress check asks whether the claimed milestone clears the bar for the current stage; a run that falls short reverts to an earlier pass or stops (\secrefc{sec:swarm-workflow}).

\paragraph{Worker Pool.} Swarm runs many paper workspaces through the same loop at once. It assigns idle Workers to saved workflow states, resumes the correct role after each turn, and sends experiment declarations to Nexus under that paper's budget. The \builder chooses the scientific next step; Swarm handles scheduling, accounting, and routing the returned evidence back to the right workspace.

\subsubsection{Milestones}
\label{sec:swarm-milestones}
\vspace{-.2cm}

\noindent
\begin{minipage}[t]{0.66\linewidth}
\vspace{0pt}
\paragraph{Progress Ladder.} Swarm measures progress with a canonical milestone ladder, not with a binary success flag. The main target in the standard workflow is \texttt{core\_claim}: a minimum experiment must run the paper's method and support the central claim. \texttt{method\_runs} records weaker progress when the method executes but the claim is not yet supported; \texttt{core\_claim\_plus}, \texttt{secondary\_claims}, \texttt{majority}, \texttt{near\_complete}, and \texttt{full} record broader coverage; and \texttt{not\_possible} records a case that should not be converted into an environment. These labels let Swarm preserve partial progress without recording it as success. The canonical labels are listed in \tabref{tab:swarm-milestones}. The agents self-report this.

\paragraph{Gating.} Milestones also determine whether the workflow should continue. If a \builder declares completion below \texttt{core\_claim}, Swarm gives one or more turns asking for additional evidence. If a required pass still falls short, Swarm can stop the trajectory or revert to an earlier pass. This keeps failed attempts inspectable without letting weak evidence advance.
\end{minipage}\hfill
\begin{minipage}[t]{0.30\linewidth}
\vspace{8pt}
    \centering
    \begin{softcolortable}[1]
    \footnotesize
    \renewcommand{\arraystretch}{1.43}
    \begin{tabular}{
        >{\centering\arraybackslash}m{\softTableContentWidth}
        }
    \toprule
    \rowcolor{roleHeader}
    {\small\textbf{milestone}} \\
    \midrule
    \rowcolor{milestoneNotPossible}
    \texttt{not\_possible} \\
    \rowcolor{milestoneNone}
    \texttt{none} \\
    \rowcolor{milestoneMethod}
    \texttt{method\_runs} \\
    \rowcolor{milestoneCore}
    \texttt{core\_claim} \\
    \rowcolor{milestoneCorePlus}
    \texttt{core\_claim\_plus} \\
    \rowcolor{milestoneSecondary}
    \texttt{secondary\_claims} \\
    \rowcolor{milestoneMajority}
    \texttt{majority} \\
    \rowcolor{milestoneNear}
    \texttt{near\_complete} \\
    \rowcolor{milestoneFull}
    \texttt{full} \\
    \bottomrule
    \end{tabular}
    \end{softcolortable}
    \captionof{table}{\textbf{Milestones make partial progress explicit.} Swarm records one canonical label for each run.}
    \label{tab:swarm-milestones}
\end{minipage}
\par\vspace{0.em}

\subsubsection{Builder Roles}
\label{sec:swarm-roles}
\vspace{-.2cm}

\paragraph{Role Design.} Swarm does not give every agent the same job. It separates reproduction, environment design, packaging, and verification so that later agents can inspect and improve the artifacts left by earlier agents; we refer to these roles collectively as \emph{\builders}, the agents that construct an environment from a paper. Reproducers build and test the paper method; design agents convert a successful reproduction into a scored environment; the packager makes the workspace reusable; and verifiers audit whether the final milestone is supported by evidence. This role split keeps the workflow from depending on a single long agent session while still preserving one shared workspace. \tabref{tab:swarm-workflow} summarizes the role responsibilities and the gate to satisfy.

\paragraph{Milestone Selection.} Each \builder also chooses the milestone that best describes the current workspace after it is done. The choice is a self-evaluation against the canonical ladder in \tabref{tab:swarm-milestones}: the agent claims the milestone supported by runnable code, logs, scores, and notes, rather than simply reporting when the assigned role finishes. The label is a claim, not a verified fact, so it is never taken as final on its own: each later role and iteration inherits the same workspace and re-audits the previous agent's claim, and the verifier lowers any milestone the workspace's logs, scores, and code do not support. The bar for \texttt{core\_claim} in particular is that the supporting run executes the paper's own method on the paper's task, possibly at reduced scale, and reproduces the direction of its central result, not merely that the code runs without error.

\begin{table}[h]
    \centering
    \begin{softcolortable}[0.87]
    \begin{tabular}{
        >{\centering\arraybackslash}m{0.30\softTableContentWidth}
        >{\raggedright\arraybackslash}m{0.40\softTableContentWidth}
        >{\raggedright\arraybackslash}m{0.30\softTableContentWidth}
        }
    \toprule
    \rowcolor{roleHeader}
    \textbf{role} &
    \multicolumn{1}{>{\raggedright\arraybackslash}m{0.40\softTableContentWidth}}{\textbf{main responsibility}} &
    \multicolumn{1}{>{\raggedright\arraybackslash}m{0.30\softTableContentWidth}}{\textbf{gate to continue}} \\
    \midrule
    \rowcolor{roleReproducer}
    \softTableRoleCell{reproducer} &
    \softTableTextCell{implements the paper method and\\ tries to reproduce the central claim.} &
    \softTableTextCell{reach \texttt{core\_claim},\\ or stop or revert.} \\
    \rowcolor{roleDesigner}
    \softTableRoleCell{designer} &
    \softTableTextCell{converts the reproduction into gym\\ targets, constraints, and scoring.} &
    \softTableTextCell{scores must validate} \\
    \rowcolor{rolePackager}
    \softTableRoleCell{packager} &
    \softTableTextCell{consolidates scripts, environment\\ files, and download logic.} &
    \softTableTextCell{workspace needs to\\ be able to rebuild.} \\
    \rowcolor{roleVerifier}
    \softTableRoleCell{verifier} &
    \softTableTextCell{audits whether the final evidence\\ supports the claimed milestone.} &
    \softTableTextCell{unsupported claims have their milestone lowered.} \\
    \bottomrule
    \end{tabular}
    \end{softcolortable}
    \caption{\textbf{Role separation keeps one workspace auditable across \builder handoffs.} The role split separates reproduction, environment design, packaging, and verification while keeping all roles on the same workspace and evidence trail.}
    \label{tab:swarm-workflow}
    \vspace{-.4cm}
\end{table}

% \newpage

\subsubsection{Workflow}
\label{sec:swarm-workflow}

\vspace{-.2cm}

\begin{wrapfigure}[21]{r}{0.5\linewidth}
    \vspace{-.75cm}
    \centering
    \includegraphics[width=\linewidth]{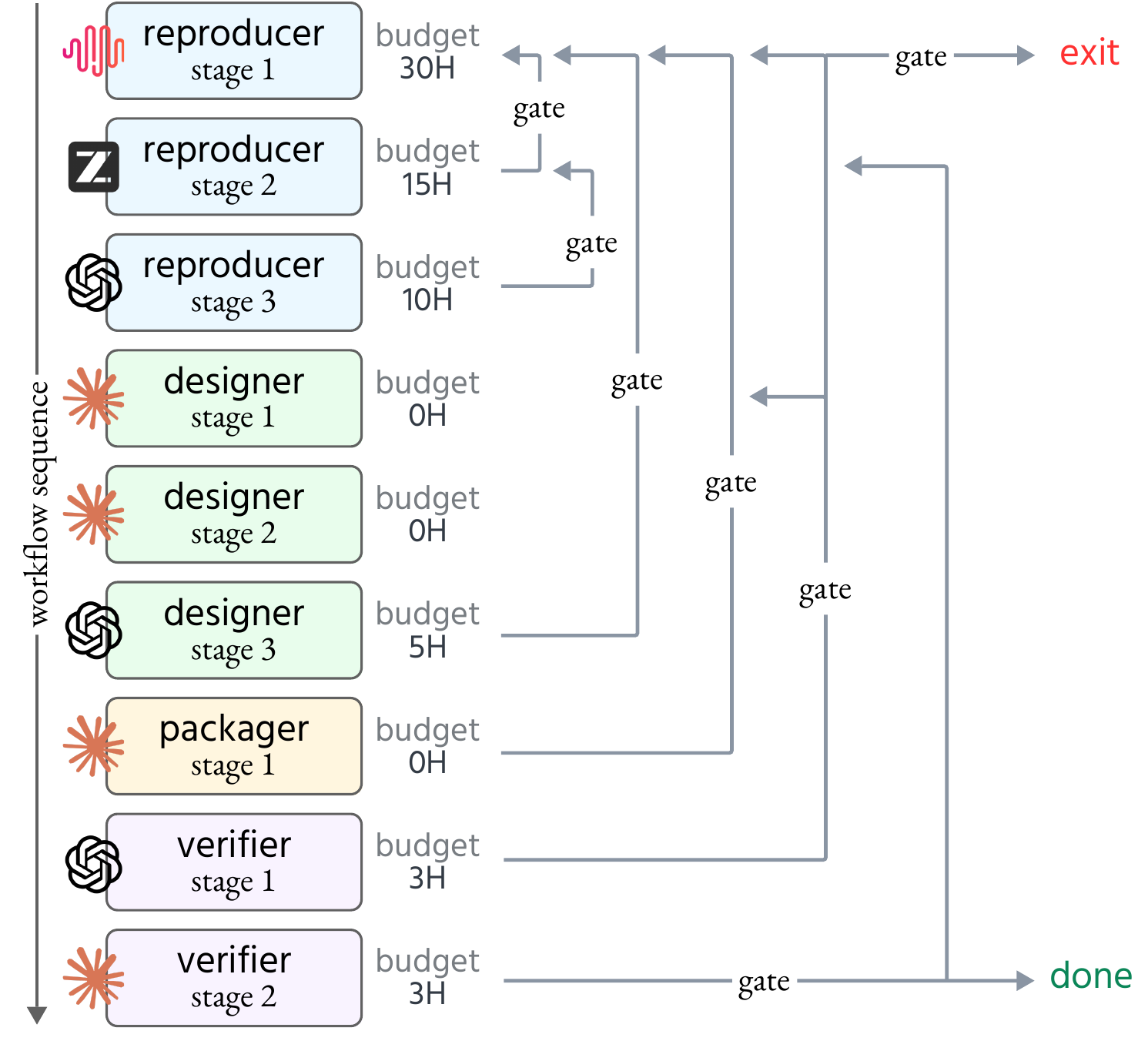}
    \caption{\textbf{Swarm workflow stages.} The standard workflow expands reproduction, design, packaging, and verification into repeated \builder passes; each row carries the model profile, gate, GPU budget, and possible revert target.}
    \label{fig:swarm-workflow-stages}
\end{wrapfigure}

\paragraph{Pass Sequence.} The standard workflow is an ordered list of stage rows (\figref{fig:swarm-workflow-stages}). Each row specifies the model profile, role, GPU-hour budget, milestone cap, minimum milestone, and optional revert target for that pass. When a pass finishes, Swarm compares the claimed milestone with the row's minimum requirement. Meeting the gate advances to the next row; falling below it triggers the configured revert target, or otherwise stops the trajectory as unsuccessful. Because the budget is attached to the row, routine checks stay small, experiment-heavy rows receive more room, and every paper is judged under the same stage-level limits.

\paragraph{Repeated Passes.} A workflow can include repeated rows for the same role. This is deliberate: a single \builder round may leave partial progress, fragile evidence, or an unresolved design choice that should be inspected before the workflow advances. Repeated rows let the next agent continue from the saved workspace, try a different strategy, or audit the previous claim without treating the role as complete. 

\paragraph{Multi-Agent Handoffs.} The handoff structure also makes the workflow a multi-agent process rather than a single long session. A later \builder inherits the workspace, evidence, and gate left by the previous agent, so it can inspect earlier choices before extending them. This gives the system diversity in search strategy and reduces the chance that one mistaken assumption carries through the entire artifact. It also saves cost by placing lower-cost model profiles on routine continuation and reserving stronger profiles for difficult passes. Since the handoff includes the current milestone claim, each agent can revise the milestone selected by previous agents.

\subsection{Worker: Agent Interface}
\label{sec:worker}

\begin{figure}[h]
    \centering
    \includegraphics[width=\textwidth]{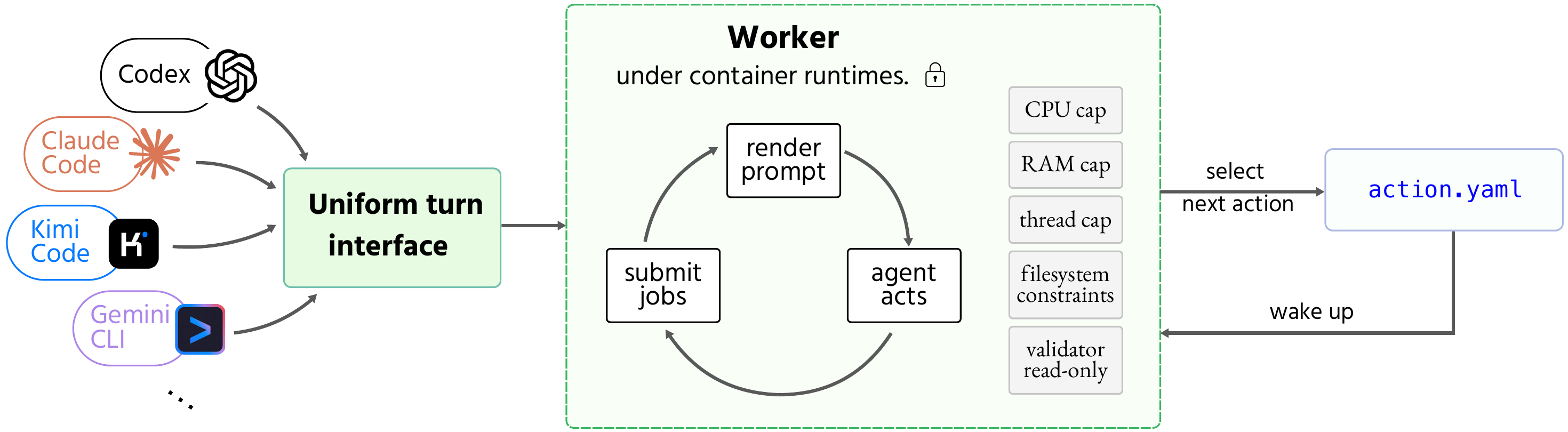}
    \caption{\textbf{Worker runtime overview.} Different agent scaffolds connect through one uniform turn interface. Worker runs the agent under a container runtime with capped resources and a read-only validator, cycling through a turn loop of render prompt, agent acts, and submit jobs. Each turn ends by selecting the next action in \texttt{action.yaml}, which the system reads before waking the agent for the following turn.}
    \label{fig:worker-overview}
\end{figure}

% \paragraph{Adapter.} 
We design Worker as one interface through which every agent scaffold CLI connects to the system. Scaffolds vary in how they are launched, resumed, and billed; Worker absorbs that variation so a model is judged on the research it produces, not on the scaffold that drives it. \figref{fig:worker-overview} shows this path: heterogeneous scaffolds enter through one uniform turn interface, run inside a constrained container, and end each turn by writing an \texttt{action.yaml} that wakes the system for the next turn.

\begin{wrapfigure}[10]{r}{0.3\linewidth}
    \vspace{-.5cm}
    \centering
    \includegraphics[width=\linewidth]{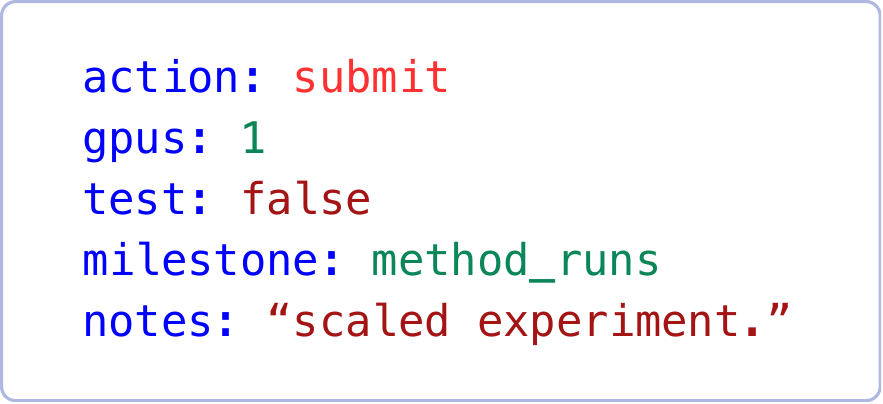}
    \caption{\textbf{The \texttt{action.yaml} that ends a turn.} The agent writes it to request the next job or to stop, and the system reads it after every turn.}
    \label{fig:action-yaml}
\end{wrapfigure}

\paragraph{Turns and actions.} A session is a sequence of turns. The first turn sets the agent up in its workspace; every later turn is short, sent only when something changes, such as a finished job or a failed build. The agent ends a turn by writing one file, \texttt{action.yaml}, that states its decision, shown in \figref{fig:action-yaml}.

The agent makes its request in this fixed format. The system then processes it to decide the next step. \texttt{action} says whether to submit a job or declare the paper done; \texttt{gpus} and \texttt{test} set its size and route it to the full or the short test pool; \texttt{milestone} records the agent's own assessment against the fixed milestone ladder. Because the agent never launches a job itself, the system can check every request, count it against the budget, and schedule it before it runs; a malformed one is handed back to rewrite, while the reasoning behind it survives in the turn record. The loop also removes any need to finish in a single session: the agent is re-engaged as jobs complete and builds fail, and progress accumulates between turns.

\paragraph{Containment.} We run the agent inside an Apptainer container \citep{apptainer} whose home is the paper workspace. It sees only a base image and that workspace: the container hides the cluster's home and temporary directories and drops all administrator bind mounts, so the shared filesystems are out of reach. Inside, it edits code freely, and the GPU jobs it later submits reuse the same layout, so no host path is ever exposed.  Three limits keep one agent from starving its neighbors: a memory cap to bound total RAM, a low scheduler priority to make co-located agents yield CPU, and thread caps on the numeric libraries.

\paragraph{Dependencies.} Installing packages is separated from running jobs. The agent does not install during a job; it declares what it needs in two workspace files: \texttt{environment/container.def} for the base image and OS packages, and \texttt{environment/setup.sh} for Python packages. The system builds these once and keeps the result in a persistent overlay on top of the base image, so the packages are available to every later job without reinstalling or rebuilding the container. Editing either file triggers a rebuild before the next job runs.

\paragraph{Boundary and robustness.} We place a few things outside the agent's reach by construction: the bookkeeping files and the recorded trajectory are hidden, the validator is read-only, and \texttt{git push} is disabled so the system decides when work is published. Turns are stripped of credential patterns before they are written, since the per-paper repositories may be made public. And because a paper can span thousands of unattended turns, Worker recovers on its own: it retries transient errors such as rate limits, switches to another model profile on terminal ones such as an exhausted usage cap, and kills a stalled turn along with its whole process tree.

\subsection{How the Components Connect}
\label{sec:layer-handoffs}

\paragraph{Forward path.} The four components pass work to each other through recorded state, not inside one long agent session. Worker presents the paper and workspace to the agent and reads its action from \texttt{action.yaml}. Swarm owns the paper's trajectory: it tracks the milestone, advances through the workflow of roles, prepares the workspace, and turns a \texttt{submit} action into a job request for Nexus. Nexus checks the project's budget, its concurrency cap, and the requested resource pool, then forwards the job to Hive, which runs it.

\paragraph{Return path.} Results travel back through the same components. Hive reports the job's state, return code, and resource usage; Nexus matches the finished job to its run record, writes the logs and outputs into the workspace, releases the GPUs it was holding, and charges the run against the project's budget. Swarm reads the event, updates the paper's state, and wakes the Worker for the next turn. Because everything lands as files the agent can open, not as hidden service state, a later turn or a different model can look back at what each job did and how it ended.

\paragraph{Why split it this way.} The split gives each component a single, narrow responsibility, and that is what keeps the system auditable. Worker can think and edit but cannot give itself compute; Swarm can coordinate papers but does not run jobs; Nexus enforces resource limits without judging scientific progress; Hive runs jobs without knowing which claim they support. Because the boundaries between them are clean, the four parts are meant to work independently, and any one of them can be swapped for a different implementation, a new scaffold, scheduler, or compute backend, without touching the others. The final artifact is tied to both sides: the agent trajectory that produced it, and the run records that make that trajectory checkable.

\newpage

\section{Builder Design}
\label{sec:agent-pipeline}

\begin{figure}[h]
    \centering
    \includegraphics[width=\textwidth]{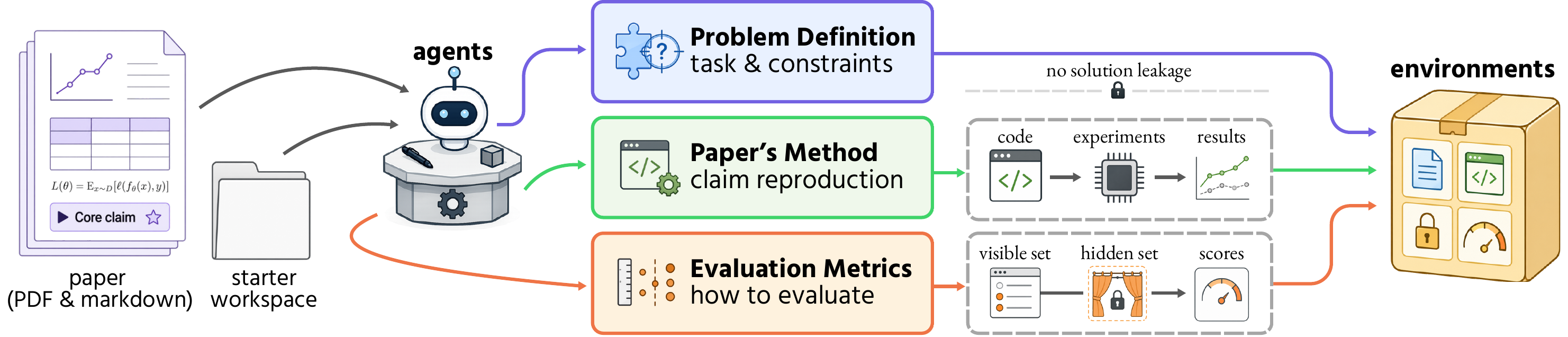}
    \caption{\textbf{Builder overview.} From a paper (in PDF and Markdown) and an empty starter workspace, \builders build a reusable environment along three tracks: a problem definition that states the task and its constraints without leaking a solution, a reproduction of the paper's method whose code and experiments recover the central claim, and an evaluation harness with visible and hidden slices that turns outputs into scores.}
    \label{fig:agent-overview}
\end{figure}

\sw runs entirely on autonomous agents: every environment is produced by \builders that reproduce a paper and shape the result into a task, with no human in the loop. This section describes how those \builders work and the principles that decide whether the environment they leave behind is a good one. \figref{fig:agent-overview} gives the overview: from a paper and an empty workspace, \builders produce a reusable environment across problem definition, method reproduction, and evaluation.

\subsection{What the Builders Do}
\label{sec:agent-workflow}

\paragraph{The task.} \secrefc{sec:infrastructure} described the system that runs the \builders; this section is about the research task it is built for. Given a single published paper, the agents must reproduce the paper's central result and leave behind a reusable environment, an executable setup in which a later scientist can attempt to improve on that result. This is the foundation everything else rests on, since an environment whose method was never really implemented would have nothing to measure. The rest of this subsection traces what the agents actually do, from the paper to the finished environment and how each turn is driven; the design choices behind that environment follow in \secrefc{sec:design-choices}.

\paragraph{From paper to environment.} A \builder's input is the paper itself, in both Markdown and PDF with its figures, together with a fresh, empty workspace. The reproducer starts from this and proceeds in order. It first reads the paper and writes a short briefing, separating the problem and the evaluation protocol from the paper's method, and transcribes the paper's reported numbers into a reference scoring file. It then prepares the environment: downloading the data and any model weights in advance, writing an evaluator that scores outputs without depending on the method, and checking that the scoring file is well formed. With that in place it implements the method and reproduces incrementally, submitting compute jobs to get one experiment that supports the core claim running before expanding to further settings, and recording its measured numbers alongside the paper's. Reproduction usually spans several agents in turn: each later one inherits the saved workspace, audits what the previous agent built, and pushes it further.

\begin{table}[h]
    \centering
    \begin{softcolortable}[0.87]
    \setlength{\tabcolsep}{7pt}%
    \setlength{\softTableContentWidth}{\dimexpr\softTableWidth-4\tabcolsep\relax}%
    \renewcommand{\arraystretch}{1.45}%
    \begin{tabular}{
        >{\leftskip=9pt \raggedright\arraybackslash}m{0.32\softTableContentWidth}
        >{\raggedright\arraybackslash}m{0.68\softTableContentWidth}
        }
    \toprule
    \rowcolor{promptHeader}
    \textbf{condition} & \textbf{what the \builder is asked to do} \\
    \midrule
    \rowcolor{gray!8}
    new paper & read the paper and reproduce its method from a full briefing \\
    \rowcolor{white}
    continued iteration & audit the previous workspace and push it further, starting one milestone below the prior claim \\
    \rowcolor{gray!8}
    job running & read the streaming logs and decide whether to let the job continue or cancel it \\
    \rowcolor{white}
    job finished, budget left & inspect the results, record progress, and plan the next job \\
    \rowcolor{gray!8}
    job finished, budget spent & stop submitting and wrap up the milestone, scoring files, and environment \\
    \rowcolor{white}
    build failed & read the build log, fix the container definition, and resubmit \\
    \rowcolor{gray!8}
    stopped early & keep going; the stop is not yet justified against the target milestone \\
    \rowcolor{white}
    signaled done & finalize and commit, a permanent decision \\
    \rowcolor{gray!8}
    validation failed & repair the flagged scoring or structural problems and re-sign \\
    \bottomrule
    \end{tabular}
    \end{softcolortable}
    \caption{\textbf{Each turn is prompted for its condition.} The first turn is a full role briefing; later turns receive a short prompt for whatever event woke the \builder, keeping it on task across a long unattended run.}
    \label{tab:agent-prompts}
\end{table}

\paragraph{Finishing the environment.} A working reproduction is not yet a usable environment. Once the method runs and supports the claim, a designer adds the visible and hidden evaluation slices and the scripts that produce their scores; a packager makes the workspace rebuild from a clean clone, with every dataset and model fetched by a self-contained download script; and a verifier audits the whole artifact before it is accepted. How these roles are sequenced, what budget each receives, and how progress is graded are handled by the system (\secrefc{sec:swarm}); what each role leaves in the workspace is detailed in \secrefc{sec:agent-workspace}.

\paragraph{How a \builder is prompted.} The \builder works in turns, ending each by requesting a compute job or declaring the work done, and runs fully autonomously with no human in the loop (\secrefc{sec:worker}). Because a paper can span thousands of unattended turns, it is not given one fixed instruction. Where \secrefc{sec:swarm} describes when an agent is woken and what state it is shown, here the point is what it is then asked to do: the first turn carries the full role briefing, and every later turn is a short prompt written for the condition that woke it. \tabref{tab:agent-prompts} lists the main conditions and the instruction each prompt carries.
% ; the prompt texts are reproduced in full in a later section.

\subsection{Design Choices}
\label{sec:design-choices}

A good environment does not come from reproducing a paper alone; it also has to be designed so that a later method can be tested fairly on it, all under a limited compute budget. We follow seven design choices toward this, each stated first and then explained.

\begin{designbox}{dcBlue}
{\small\textsf{\textbf{Design choice 1.} Reproduce the paper's method and core claim, not its exact numbers. Scale down resources when needed, but preserve the experimental procedure and keep the task non-trivial.}}
\end{designbox}

The goal is to make the paper's method run, not to match a table entry. Exact-number reproduction rewards the wrong behavior: a \builder could tune toward the published values, or copy them outright, without implementing the method. Instead, the environment should contain the paper's actual computation and produce reasonable support for the claimed effect. A simple check is whether a reader of the code can recognize the algorithm from the paper.

Most papers cannot be reproduced at full scale under a fixed budget. The \builder may therefore reduce the model size, dataset size, or number of training steps, which also keeps reproductions friendlier to the broader community using the artifact. What it may not do is change the procedure. If the paper trains from scratch, the environment trains from scratch. If the paper uses a particular optimization setup, the environment keeps that setup. Resources can shrink; the experiment cannot become a different experiment.

Scaling down also has a limit. A task made too small becomes trivial, and a trivial task cannot distinguish the paper's method from a weak baseline. When a \builder uses a smaller or synthetic dataset, it must check that a basic baseline does not already solve the problem. If the reproduced method is no better than random guessing, majority class, or another weak baseline within noise, the reproduction has failed. The environment needs a real gap so that future agents have something meaningful to improve.

\begin{designbox}{dcGreen}
{\small\textsf{\textbf{Design choice 2.} Keep the problem separate from the solution: the problem should be stateable without the paper's method, and the evaluator and data should not depend on that method.}}
\end{designbox}

This boundary keeps the problem definition independent of the solution that produced it. The problem, its data, and its evaluator can be stated and run without the paper's method, so the environment poses the problem on its own terms rather than only as a continuation of one particular implementation; the file boundary is described in \secrefc{sec:ws-scoring}. The separation also protects the score. If the evaluator secretly depended on the reproduced method, then changing the method might not change what is measured, and the environment would look functional while no longer testing the intended problem.

The separation must hold at runtime, not only in the file layout. The evaluator must read the outputs a method produced, never call the method's code: if deleting \texttt{method/} would change the evaluation scores, the boundary is broken. Likewise, the data loaders belong to the problem, not the solution, and must not import from \texttt{method/}. The evaluate scripts must also discover methods dynamically, enumerating whatever methods exist on disk rather than hardcoding a name, so that the evaluator works for a future method without editing. How the environment is then presented to a future agent is a matter for evaluation (\secrefc{sec:eval-modes}).

\begin{designbox}{dcPurple}
{\small\textsf{\textbf{Design choice 3.} Provide a visible evaluation slice for development, and reserve a matched hidden slice for final evaluation to detect shortcuts and overfitting.}}
\end{designbox}

The finished environment includes an evaluation suite that a future scientist can see and optimize against. If that suite is only one metric on one dataset, the easiest path may be to game that number rather than improve the underlying idea. To reduce this risk, the designer builds two slices: a visible slice for development and a hidden slice that is run only after done. A large gap between the two suggests overfitting or shortcuts.

The split can be drawn in different ways, depending on the paper: held-out samples, held-out subgroups, harder inputs, or several related metrics that measure the same claim from different angles. Using multiple metrics from different angles is stronger than a single held-out number, because a single metric can be gamed while triangulation from several directions catches shortcuts that any one metric would miss.

The hidden slice must be calibrated against the visible one in difficulty: both should probe the method at roughly equal rigor, using matched data scales, evaluation protocols, and sample counts. An honest method should score similarly on both; if the hidden slice is simply harder, then every method will look worse there and the gap will be meaningless. The two slices should also be independent, so reading or importing the visible evaluator does not reveal the hidden one. And the test slice should be lightweight: it re-scores existing artifacts or slices them differently, rather than retraining or re-running expensive inference.

\begin{designbox}{dcAmber}
{\small\textsf{\textbf{Design choice 4.} Score future methods on the quantity the paper cares about, not on how closely they match the paper. Fix the constraints needed to make gains meaningful.}}
\end{designbox}

An environment should allow future methods to improve on the paper. For that reason, the score must not measure distance from the paper's published number. Such a metric would make the paper's own method the target and turn the environment into a reproduction-fidelity test. The paper's reported numbers should be recorded as context, but future methods should be scored on the underlying quantity: accuracy, loss, robustness, sample efficiency, parameter count, or whatever the claim is actually about.

This principle is enforced in the scoring schema: metric names that encode closeness to the paper, such as \texttt{deviation}, \texttt{vs\_paper}, or \texttt{agreement\_with}, are blocked by the validator, so the score must name the underlying scientific quantity.

The environment must also define what stays fixed. A future scientist should not win by using more compute, a larger model, more data, or a different training paradigm unless that is the allowed direction of improvement. For papers that claim a benefit without a cost, the evaluator should measure both sides: the benefit that should improve and the constraint that should not break. The scoring contract expresses this through metric coefficients: a positive coefficient marks a benefit to maximize, a negative one marks a constraint that must not degrade, and the experiment weights reflect how central each result is to the paper's claim.

\begin{designbox}{dcRose}
{\small\textsf{\textbf{Design choice 5.} Score on quantities that do not depend on the machine. Replace hardware-sensitive measures such as latency or wall-clock time with stable proxies.}}
\end{designbox}

Hardware varies from run to run, so a latency or wall-clock measurement can swing severalfold with no change to the method, and an environment that scored it would reward landing on a faster GPU rather than finding a better idea. Scores therefore use quantities that are stable across machines: parameter count, FLOPs, training steps, examples processed, or final task performance. When the paper's own claim is about speed or efficiency, the environment keeps the claim but measures it through a machine-independent proxy rather than the clock.

\begin{designbox}{dcTeal}
{\small\textsf{\textbf{Design choice 6.} Reproduce enough of the paper to give future scientists a real surface to improve, preferably a small matrix over the paper's own axes rather than one isolated number.}}
\end{designbox}

A one-number reproduction is too thin a target. A future scientist may match it and still have little room to learn from or improve. Once the core claim runs, the environment should expand along the axes the paper itself studies: datasets, input difficulty, model scale, ablations, data regimes, or other relevant settings. A small matrix over two meaningful axes is usually more useful than repeated runs of the same setting.

This matrix does not need to be large. It needs to expose the structure of the claim. Combined with the non-triviality check above, it turns the environment from a single point estimate into one with headroom: weak baselines leave a gap, the reproduced method occupies a recognizable starting point, and future methods have several genuine ways to improve.

\begin{designbox}{dcIndigo}
{\small\textsf{\textbf{Design choice 7.} The environment must rebuild and rerun from a fresh clone, with every external resource fetched by a self-contained script and every dependency declared.}}
\end{designbox}

An environment that only works on the machine that built it is not reusable. The download script must fetch every dataset, model weight, and external resource from a public source using executable commands, not comments or placeholder URLs. It must be idempotent and must not reference local shared paths that exist only on the original cluster. The container definition and setup script must declare every system and Python dependency so that the software rebuilds without manual intervention. And every local import in the evaluation, method, and baseline code must resolve to a file tracked by git, so that a fresh clone contains everything the experiments need to run.

The validator enforces this: it scans committed code for references to machine-local shared directories and checks that every import target is a tracked file. Together with the reproducible-execution scripts (\secrefc{sec:ws-execution}), this makes the environment a portable artifact rather than a snapshot of one session's state.

\newpage

\section{Agent Workspace}
\label{sec:agent-workspace}

The workspace is the single directory the \builders work in and the artifact they leave behind. Its layout is what makes the result a reusable environment rather than the residue of one run. We describe its structure (\secrefc{sec:ws-structure}), its problem statement and scoring (\secrefc{sec:ws-scoring}), its execution (\secrefc{sec:ws-execution}), and its records and checks (\secrefc{sec:ws-records}).

\subsection{Structure}
\label{sec:ws-structure}

\paragraph{Overview.} Each paper is reproduced inside a single working directory, and that directory is also the artifact the \builders leave behind. During the run it holds the code, data, experiments, and notes; once the run ends, it is designed to stand on its own, so that another party can clone it, rebuild the environment, rerun the method, and obtain the same scores without access to the original session. This is what distinguishes a reusable research environment from the residue of a single successful run, in which the method is entangled with the evaluator, the data exists only in a local cache, and the scores carry no well-defined meaning. The layout shown in \figref{fig:workspace-structure} is designed to prevent these failures.

\begin{center}
    \includegraphics[width=0.7\linewidth]{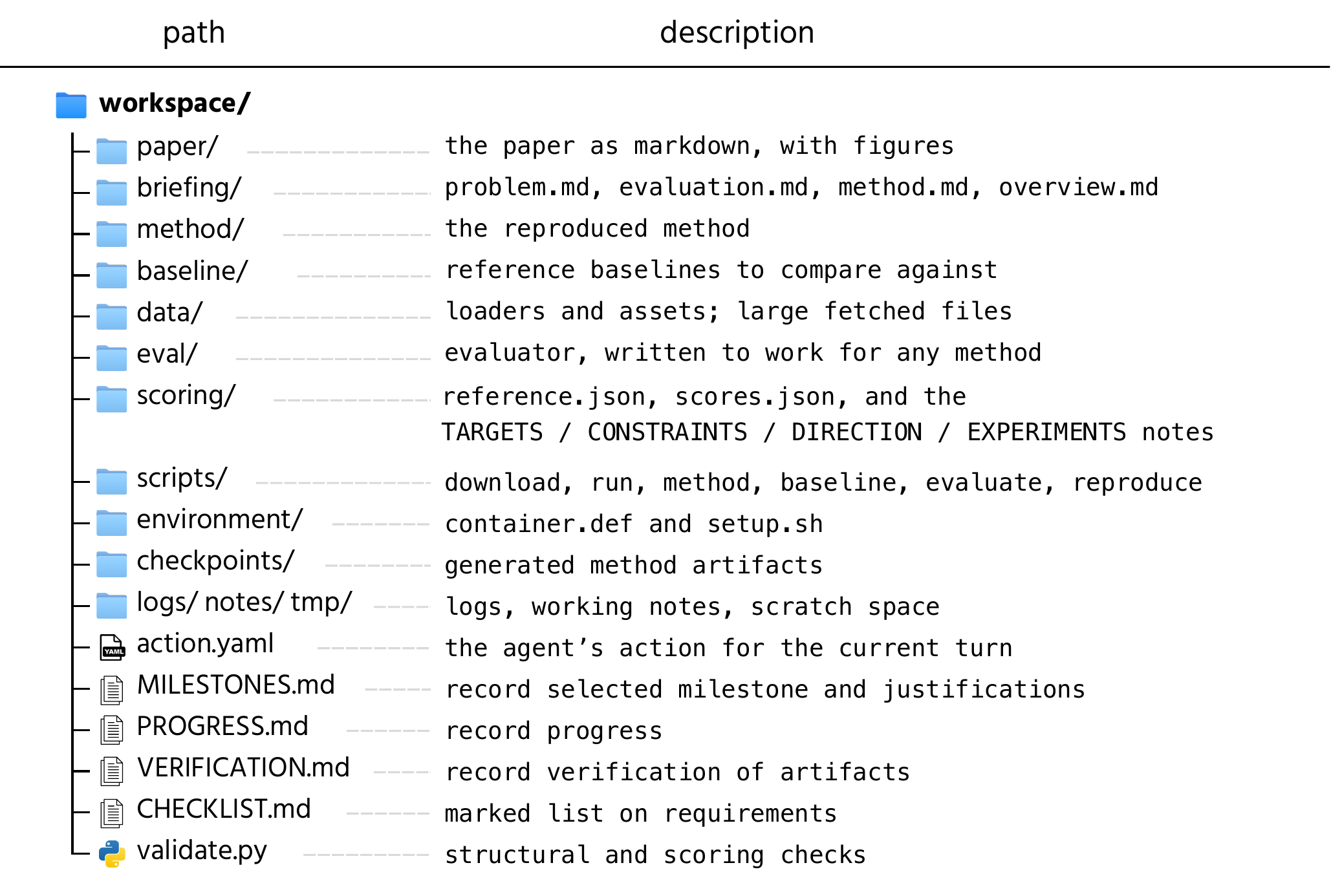}
    \captionof{figure}{\textbf{The agent workspace.} A single directory holds the paper and briefing, the reproduced method and baselines, the data and evaluator, the scoring contract, the run scripts and environment, and the root-level record and validation files. Each directory has a single responsibility, which keeps the problem separated from the solution, lets the workspace rebuild and rerun from a fresh clone, and preserves a record of how the reproduction was produced.}
    \label{fig:workspace-structure}
\end{center}

\paragraph{File Roles.} The layout is a deliberate design choice rather than a storage convention: each directory has a single responsibility. The briefing defines what a future scientist may know; \texttt{method/} and \texttt{baseline/} hold what the reproducer built; \texttt{data/} and \texttt{eval/} define the problem interface that any method must satisfy; \texttt{scoring/} defines how a method is judged; and \texttt{scripts/} and \texttt{environment/} determine whether the workspace runs outside the original session. \tabref{tab:agent-workspace-files} summarizes the groups.

\begin{table}[h]
    \centering
    \begin{softcolortable}[0.87]
    \setlength{\tabcolsep}{7pt}%
    \setlength{\softTableContentWidth}{\dimexpr\softTableWidth-4\tabcolsep\relax}%
    \renewcommand{\arraystretch}{1.5}%
    \begin{tabular}{
        >{\leftskip=9pt \raggedright\arraybackslash}m{0.34\softTableContentWidth}
        >{\raggedright\arraybackslash}m{0.66\softTableContentWidth}
        }
    \toprule
    \rowcolor{roleHeader}
    \textbf{group} & \textbf{purpose} \\
    \midrule
    \rowcolor{gray!8}
    \texttt{briefing/} & states the task and how success is measured, while keeping the paper's method out of the problem files \\
    \rowcolor{white}
    \texttt{method/}, \texttt{baseline/} & the reproduced implementation, with the baselines used to interpret its score \\
    \rowcolor{gray!8}
    \texttt{data/}, \texttt{eval/} & the data interface and a method-agnostic evaluator that any future method must run against \\
    \rowcolor{white}
    \texttt{scoring/} & the scoring contract: how each metric counts, what an improvement means, and what must stay fixed \\
    \rowcolor{gray!8}
    \texttt{scripts/}, \texttt{environment/} & everything needed to rebuild the environment and rerun the method from a fresh clone \\
    \rowcolor{white}
    root files & the current action, the claimed milestone, progress and deviation notes, and the automatic checks \\
    \bottomrule
    \end{tabular}
    \end{softcolortable}
    \caption{\textbf{Each directory has one responsibility.} The workspace keeps task context, method code, evaluation, scoring, execution, and record state apart, so a later scientist can reuse it without the original conversation.}
    \label{tab:agent-workspace-files}
\end{table}

\subsection{Problem and Scoring}
\label{sec:ws-scoring}

\paragraph{Problem and Solution Boundary.} The workspace deliberately separates the problem from the solution so that the problem is method-agnostic: it states what must be achieved and how it is measured, without revealing the paper's approach. This is what allows the workspace to serve as a fair task for any future method rather than a recreation of one particular solution. The boundary appears in two places. In the briefing, \texttt{problem.md} gives the research question and success criteria and \texttt{evaluation.md} the metrics and protocol, neither naming the method; only \texttt{method.md} describes the paper's approach, and \texttt{overview.md} summarizes the finished workspace. In the code, \texttt{method/} contains the reproduced algorithm, while \texttt{eval/} and \texttt{data/} are written to work for any method, so a score never depends on the specific implementation. Because the problem stands on its own, an innovate scientist can attempt it from the briefing and evaluator alone, whereas an improve scientist additionally receives \texttt{method/} as a starting point (\secrefc{sec:evaluation-protocol}).

\begin{figure}[h]
\centering
\includegraphics[width=\textwidth]{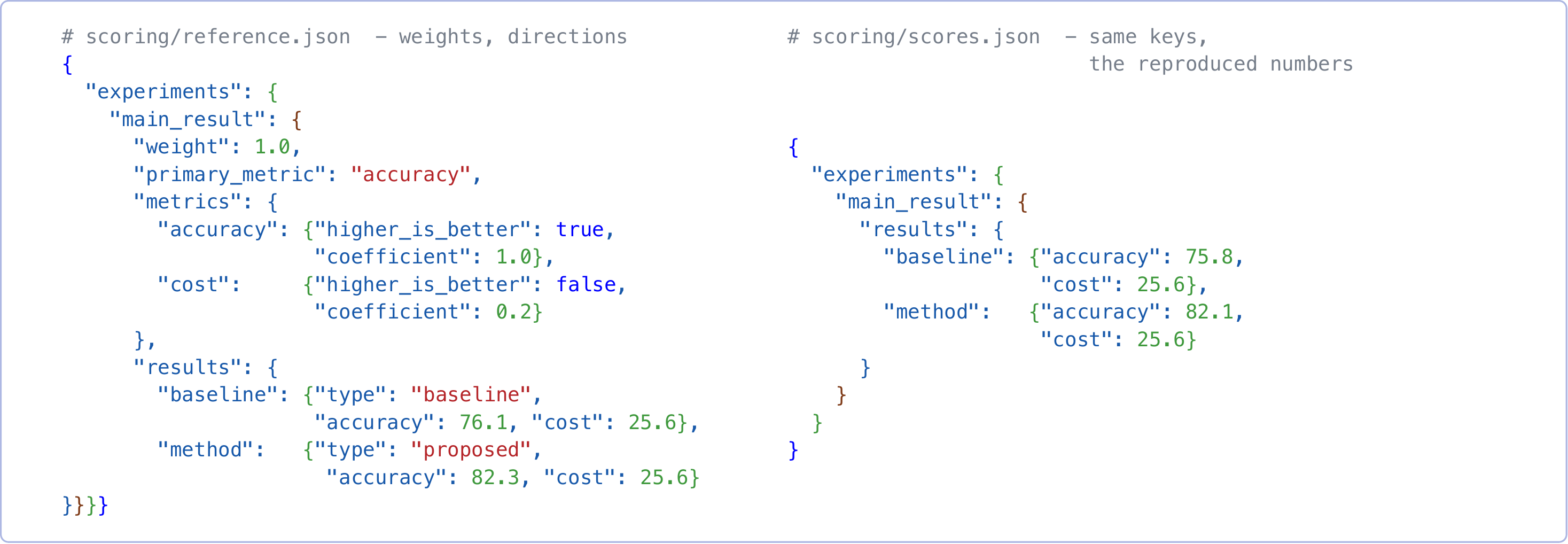}
\caption{The scoring contract, held in two files. \texttt{scoring/reference.json} fixes the weights, metric directions, coefficients, and the paper's reported numbers; \texttt{scoring/scores.json} holds the numbers the reproduction actually produced, under the same keys.}
\label{fig:scoring-json-example}
\end{figure}

\paragraph{Scoring Contract.} Scoring is held in two files. \texttt{scoring/reference.json} records the paper's reported numbers for each baseline and proposed method, along with the scoring rules: a \emph{weight} per experiment, set by the agents to reflect how central that result is to the paper's core claim (weights sum to one), the direction of each metric, and a label marking each method as a baseline or a proposed method. \texttt{scoring/scores.json} holds the numbers the reproduction actually produced. A later method is scored by its improvement over those reproduced numbers, in each metric's direction and combined by the weights, so the score measures progress over the reproduction rather than closeness to the paper. \figref{fig:scoring-json-example} shows the two files side by side.

\paragraph{Targets and Constraints.} The JSON files fix the numbers, but a scientist also requires the intent behind them. Four short markdown files provide it: \texttt{TARGETS.md} states what to optimize, \texttt{CONSTRAINTS.md} states what must remain fixed, \texttt{DIRECTION.md} states which kinds of change are permitted, and \texttt{EXPERIMENTS.md} describes each scored experiment without disclosing the method. Together they exclude trivial gains, such as substituting a larger model, adding unrelated data, or altering the problem itself, while preserving room for a genuine methodological improvement. For example, an agent may improve performance on a Mamba architecture optimization problem by replacing the architecture with a Transformer. While this could yield better results, such a change provides no scientific value, as it sidesteps the original problem rather than advancing our understanding of how to improve Mamba models themselves.
When a paper makes a trade-off claim, both sides must be scored: the benefit and the constraint it must not violate. Otherwise the easiest route to a high score is to break the very condition the paper was designed to satisfy.

\subsection{Execution}
\label{sec:ws-execution}

\paragraph{Reproducible Execution.} The workspace must be able to rebuild from source, not only on the machine where it was created. \texttt{scripts/download.sh} fetches external data, weights, and other public resources; it must be idempotent and must not reference local shared paths. \texttt{environment/container.def} captures the system dependencies and \texttt{environment/setup.sh} the packages. \texttt{scripts/method.sh} then runs the reproduced method, \texttt{scripts/baseline.sh} runs the comparisons, \texttt{scripts/evaluate.sh} scores the outputs through \texttt{eval/}, and \texttt{scripts/reproduce.sh} runs the full pipeline end to end. The workspace is therefore an executable artifact rather than a static record of results: a fresh clone rebuilds the environment, reruns the method, and regenerates the score files.

\paragraph{Visible and Held-Out Slices.} A finished workspace also includes a train/test split for the scientist who reuses it. The visible slice, \texttt{eval/train/}, is what a future agent develops against; the held-out slice, \texttt{eval/test/}, lives in the environment's pristine source copy and is materialized into a run only for the final score, so it is never present in the agent's working tree during development. The paired \texttt{scripts/evaluate\_train.sh} and \texttt{scripts/evaluate\_test.sh} write \texttt{scores\_train.json} and \texttt{scores\_test.json}. The purpose of the split is not to make the paper harder to reproduce but to detect a future agent that has overfit the visible evaluator. 
% A well-formed split uses the same interface on both sides, keeps the two sides comparable in difficulty, and scores every method present in the visible slice.

\subsection{Records and Validation}
\label{sec:ws-records}

\begin{figure}[h]
    \centering
    \includegraphics[width=\linewidth]{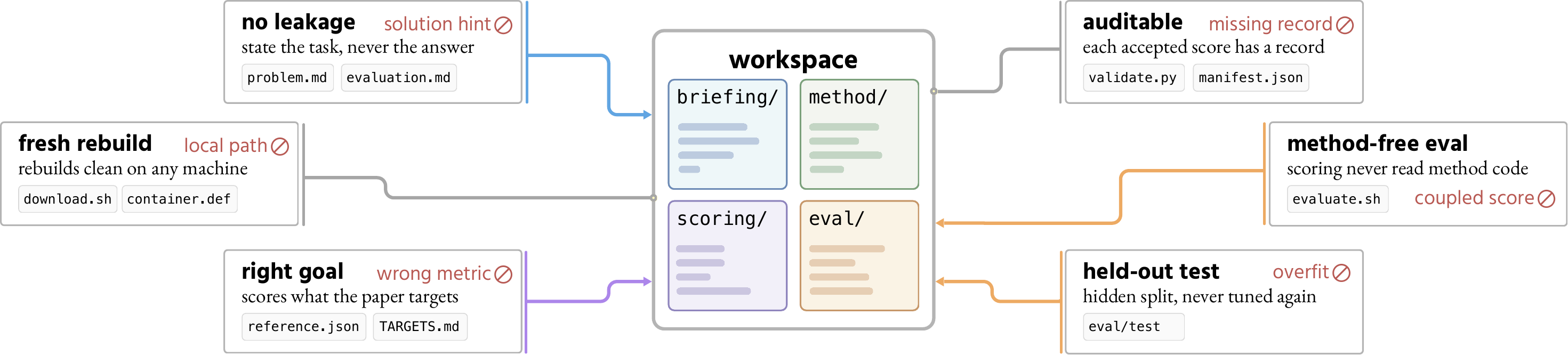}
    \caption{\textbf{Each design choice targets a concrete failure mode.} The workspace is structured to avoid solution leakage, method-coupled scoring, unrebuildable dependencies, evaluator overfitting, and unsupported claims.}
    \label{fig:workspace-design-choices}
\end{figure}

\paragraph{Design Summary.} Each choice in the workspace exists to rule out a specific way it could fail to transfer to a new scientist. \figref{fig:workspace-design-choices} collects them: the design decision, the failure it prevents, and where it lives.

\paragraph{Reproduction Record.} When a run ends, the system writes a \texttt{manifest.json} into the workspace and compiles a \texttt{README.md} containing a per-iteration reproduction log, alongside the git history that checkpoints every turn. Together these tie the result to recorded state rather than to the conversation, allowing a later agent or reviewer to trace how it was reached. \tabref{tab:workspace-trace} groups the fields the manifest records.

\begin{table}[h]
    \centering
    \begin{softcolortable}[0.95]
    \setlength{\tabcolsep}{7pt}%
    \setlength{\softTableContentWidth}{\dimexpr\softTableWidth-6\tabcolsep\relax}%
    \renewcommand{\arraystretch}{1.45}%
    \begin{tabular}{
        >{\leftskip=9pt \raggedright\arraybackslash}m{0.22\softTableContentWidth}
        >{\raggedright\arraybackslash}m{0.46\softTableContentWidth}
        >{\raggedright\arraybackslash}m{0.32\softTableContentWidth}
        }
    \toprule
    \rowcolor{roleHeader}
    \textbf{group} & \textbf{fields recorded} & \textbf{use} \\
    \midrule
    \rowcolor{gray!8}
    reproduction summary & final milestone and status, scaffold and model, workflow role and step, and turn, run, and iteration counts & attributes the result to the \builders and role that produced it \\
    \rowcolor{white}
    compute and cost & GPU and CPU hours, the GPU budget, jobs run, wall-clock duration, tokens, and dollar cost & separates a scientific failure from an execution failure, and compares spend \\
    \rowcolor{gray!8}
    components present & whether the container, method, evaluator, scripts, reference, and data are all in place & confirms the workspace is complete enough to rerun \\
    \rowcolor{white}
    per-builder records & one entry per \builder, with its turns, its jobs, and why it stopped & reconstructs the result across handoffs \\
    \bottomrule
    \end{tabular}
    \end{softcolortable}
    \caption{\textbf{The finished workspace records how it was produced.} \texttt{manifest.json} ties each accepted artifact to the \builders, compute, and components behind it, so it stays auditable after the session ends.}
    \label{tab:workspace-trace}
\end{table}

\paragraph{Acceptance Checks.} A workspace is checked before it is accepted. \texttt{validate.py} runs automatically and enforces the scoring schema, blocks the obvious paper-agreement metric names, and checks the structural rules that keep the workspace independent of its own method and rebuildable from a fresh clone; \tabref{tab:acceptance-checks} lists the checks in full. The markdown audit files then cover what code cannot decide: \texttt{MILESTONES.md} states the claimed milestone, \texttt{PROGRESS.md} and \texttt{VERIFICATION.md} record why it is justified and what was changed from the paper, and \texttt{CHECKLIST.md} confirms that the workspace is ready for both improve and innovate mode reuse. A workspace is accepted only when the automatic checks pass and the audit files are complete.

\begin{table}[h]
    \centering
    \begin{softcolortable}[0.95]
    \setlength{\tabcolsep}{7pt}%
    \setlength{\softTableContentWidth}{\dimexpr\softTableWidth-4\tabcolsep\relax}%
    \renewcommand{\arraystretch}{1.45}%
    \begin{tabular}{
        >{\leftskip=9pt \raggedright\arraybackslash}m{0.30\softTableContentWidth}
        >{\raggedright\arraybackslash}m{0.70\softTableContentWidth}
        }
    \toprule
    \rowcolor{roleHeader}
    \textbf{check} & \textbf{what it verifies} \\
    \midrule
    \rowcolor{gray!8}
    scoring schema & \texttt{reference.json} parses, experiment weights sum to one, and each experiment declares a primary metric, its metrics, and its results; every metric declares a direction and a coefficient; and every method is typed as a baseline or a proposed method, with at least one proposed \\
    \rowcolor{white}
    no paper-agreement metrics & metric names may not encode closeness to the paper (\texttt{deviation}, \texttt{error\_vs\_}, \texttt{agreement\_with}, and similar); the score must measure the scientific quantity, not fidelity to the publication \\
    \rowcolor{gray!8}
    scores match the schema & \texttt{scores.json} uses the same experiments, methods, and metric names as \texttt{reference.json}, and every value is numeric \\
    \rowcolor{white}
    reproduction sanity & each reproduced primary-metric value stays close to the paper's: no sign flip, under a $2\times$ magnitude gap, and under a $0.3$ absolute gap for bounded metrics \\
    \rowcolor{gray!8}
    evaluator independence & \texttt{eval/} and \texttt{data/} never import from \texttt{method/}, so the evaluator works for any method, not only the reproduced one \\
    \rowcolor{white}
    fresh-clone integrity & every local import in \texttt{eval/}, \texttt{method/}, and \texttt{baseline/} resolves to a git-tracked file, and no committed code references the machine-local \texttt{shared/} cache \\
    \rowcolor{gray!8}
    train/test split & when a split is present: the two slices are independent and wired to the correct evaluator and output file, both score the same methods, the slices and the primary metric actually differ, and the primary metric separates methods on the train slice \\
    \bottomrule
    \end{tabular}
    \end{softcolortable}
    \caption{\textbf{What \texttt{validate.py} checks before a workspace is accepted.} The validator enforces the scoring schema, blocks scores that measure agreement with the paper, and verifies that the workspace is independent of its own method, rebuildable from a fresh clone, and free of train/test shortcuts.}
    \label{tab:acceptance-checks}
\end{table}

\paragraph{Version Control.}\label{sec:version-control} Every workspace is a git repository, and the system commits to it continuously, so the full trajectory can be replayed from the repository alone, without the session that produced it. Each agent edit is committed as it is made, and each job submission is committed together with the exact code it ran, so the state behind any result is pinned to a specific revision. When a job finishes, its logs are written back into the workspace and committed as tracked files, so what ran and what it produced live in the same history as the code. The \builder cannot publish: \texttt{git push} is disabled, and the system decides when it is released (\secrefc{sec:worker}).

Evaluation reuses the same git-backed design with one branch per attempt. Each time a model is run on an environment, that attempt is isolated on its own branch, so attempts never interfere, and each one can be inspected, rescored, or rerun on its own from the committed state it started from.

\newpage

\section{Selecting Papers}
\label{sec:paper-selection}

Not every paper can or should become a suitable environment. A useful task has to be both worth reproducing and feasible for an autonomous \builder to attempt under a bounded budget. Before any reproduction begins, we therefore narrow a large pool of recent papers down to a smaller, category-balanced set through a staged pipeline: cheap model screens drop low-quality and off-type papers, a deeper analysis scores each remaining paper for quality and reproduction difficulty, and a final ranking and balancing step chooses the papers we attempt. The rest of this section walks through that pipeline and reports how many papers survive each stage.

\begin{figure}[h]
    \centering
    \includegraphics[width=1.0\linewidth]{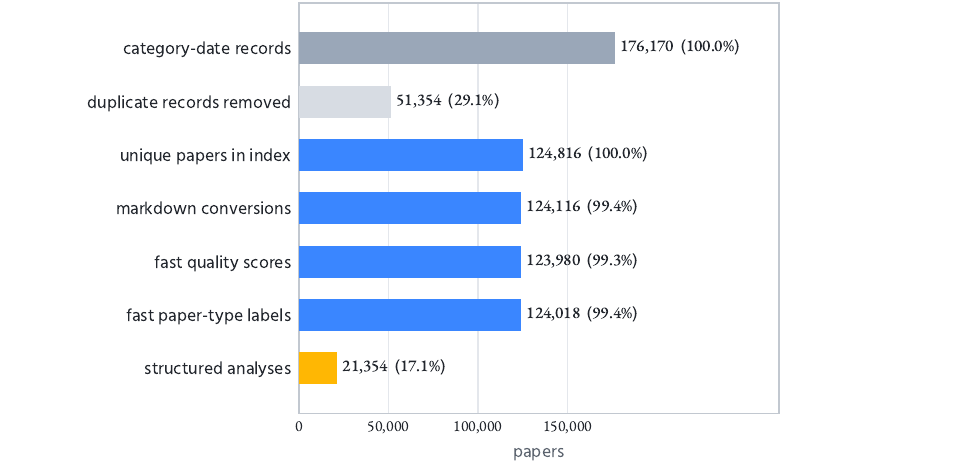}
    \caption{\textbf{Screening preserves nearly all unique papers before deeper analysis.} Counts are computed from the selection snapshot used by the export. Shares after the unique-index row use 124,816 as the denominator.}
    \label{fig:paperscope-pool}
\end{figure}

\begin{figure}[h]
    \centering
    \includegraphics[width=1.0\linewidth]{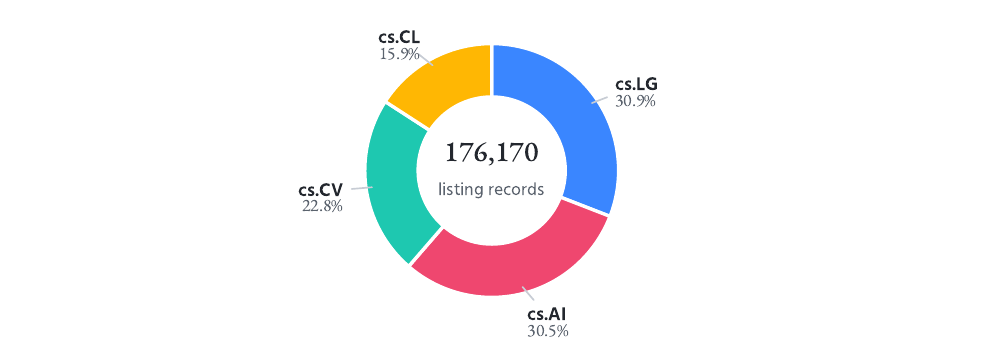}
    \vspace{-.6cm}
    \caption{\textbf{The source pool is largest in cs.AI and cs.LG before deduplication.} These are category-listing records before ID-level deduplication; a single paper can contribute to multiple rows.}
    \label{fig:paperscope-source-categories}
\end{figure}

\paragraph{Candidate Pool.} We assemble the candidate pool before any \builder is launched. Our selection snapshot is exported from a global arXiv index of 124,816 unique papers. The index is built from daily category listings between January 1, 2025 and March 1, 2026, and exported on March 28, 2026. We restrict the source categories to cs.CL, cs.CV, cs.AI, and cs.LG (\figref{fig:paperscope-source-categories}), then deduplicate by arXiv ID because the same paper can appear under more than one category. We also drop papers whose primary category is statistics (the stat.* group), which are dominated by theoretical work that offers little to execute or reproduce. This gives the candidate pool a temporal and topical boundary before any model is asked to judge feasibility. \figref{fig:paperscope-pool} reports how many papers survive each screening stage from this pool onward.

\begin{figure}[h]
    \centering
    \includegraphics[width=1.0\linewidth]{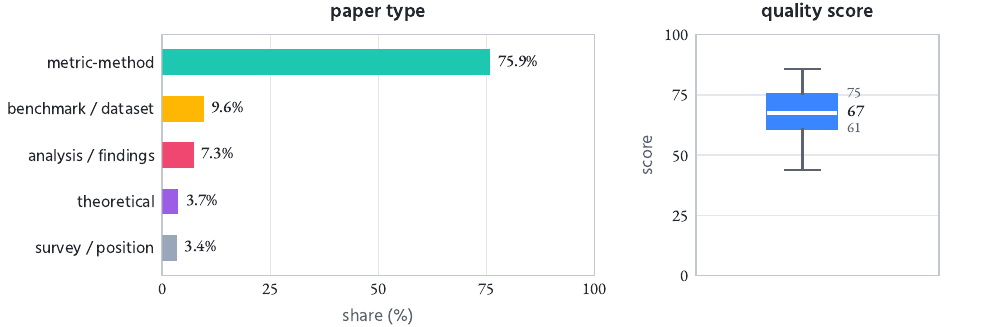}
    \caption{\textbf{Fast screens keep broad coverage while identifying metric-method papers as the dominant type.} The left side summarizes 123,980 fast quality scores; the right side summarizes 124,018 fast paper-type labels.}
    \label{fig:paperscope-fast-screens}
\end{figure}

\begin{figure}[h]
    \centering
    \includegraphics[width=1.0\linewidth]{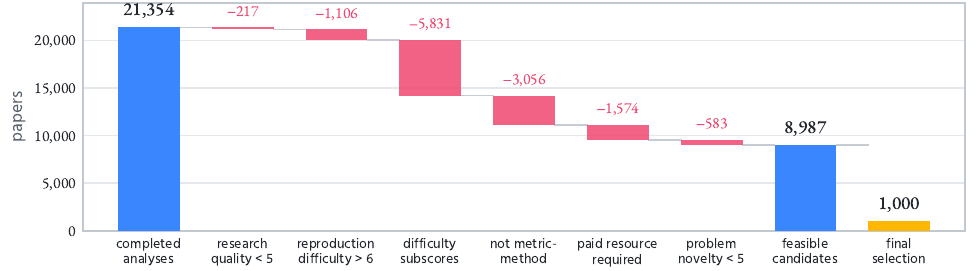}
    \caption{\textbf{Feasibility filters reduce 21,354 analyzed papers to 8,987 candidates.} The feasible candidates are ranked by quality-feasibility utility before the final category-balanced selection.}
    \label{fig:paperscope-filter-funnel}
\end{figure}

\paragraph{Fast Screening.} Before any model reads a paper, we convert its PDF into Markdown with MinerU~\citep{mineru}, which extracts the text, tables, and figures into a clean form that the screening and analysis models can read directly. The first two passes over this text are cheap, high-recall screens rather than final selection rules. Each pass sees only the opening 15,000 characters of the converted paper, which typically span the abstract and introduction, so that screening cost stays low across the full index. The first pass returns a quality score: \model{MiniMax M2.7}~\citep{minimax} reads the excerpt and emits a short justification together with an integer rating from 0 to 100. A second pass assigns each paper one of five types: metric-method, benchmark/dataset, analysis/findings, theoretical, or survey/position. The quality screen is broad: 94.3\% of scored papers are at least 50, 40.5\% are at least 70, and only 12.0\% reach 80 or higher. The type screen is more selective in the direction \sw needs: 94,161 papers, or 75.9\%, are labeled metric-method. \figref{fig:paperscope-fast-screens} reports the full quality-score and paper-type distributions. The quality screen is the gate into deep analysis: structured analysis runs only within the top quality band, papers scoring 75 or higher (33{,}393 papers, the top 26.9\%), and 21{,}354 of them complete the analysis under the analysis budget. The fast type label guides this stage but does not filter it; the decisive type filter uses the structured analysis's own classification and appears as the first drops in \figref{fig:paperscope-filter-funnel}.

\paragraph{Structured Analysis.} Papers carried forward from the fast screens go through a deeper analysis with \model{MiniMax M2.7}~\citep{minimax}, which classifies each paper and records a reproduction-oriented catalog entry. The first and most decisive field is the paper type. We keep only method papers, whose contribution is a method or technique evaluated with measurable results, and set aside the other four types: benchmark or dataset papers, where the contribution is the evaluation rather than a method; analysis or findings papers, which report observations rather than improvements; theoretical papers, whose contribution is a formal result; and survey or position papers, which carry no original experiments. Only method papers fit a reproduce-then-improve task, where an agent has to implement a method and then improve on a measured number.

\begin{figure}[h]
    \centering
    \includegraphics[width=1.0\linewidth]{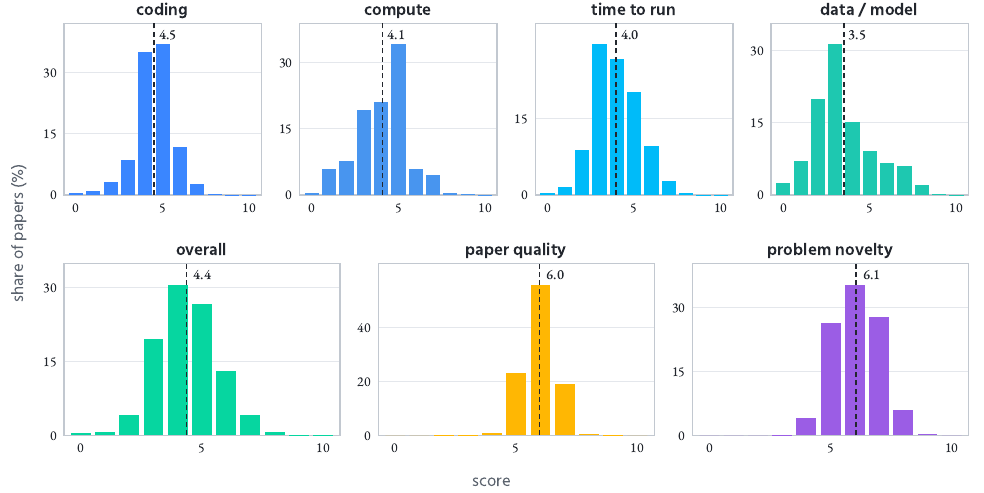}
    \caption{\textbf{Structured-analysis score distributions across the 21{,}354 analyzed papers.} Each panel gives the share of papers at each integer score from 0 to 10: the five reproduction-difficulty sub-scores (coding, compute, time to run, data or model, and overall), and the paper-level quality and problem-novelty scores. Dashed lines mark the means. Difficulty scores concentrate in the low-to-middle range, while quality and problem novelty center higher.}
    \label{fig:paperscope-analysis-scores}
\end{figure}

The remaining fields describe each surviving paper in more detail: contribution novelty, taxonomy, compute and paid-resource requirements, benchmarks, datasets, baselines, concrete dependencies on other papers, code availability, reproduction difficulty, research quality, and problem novelty, with every scalar judgment given as an integer score from 0 to 10. Reproduction difficulty is split into coding, compute, time to run, and data or model difficulty, plus an overall score, because a paper should not be chosen on scientific appeal alone; a useful task also has to be feasible for an autonomous agent under a bounded budget. \figref{fig:paperscope-analysis-scores} shows how these scores are distributed across the analyzed papers.

\paragraph{Filtering and Ranking.} The final selection balances importance against feasibility. Let \(q\) be the 0-10 paper quality score, and let \(r\) be the aggregate reproduction difficulty. When all four reproduction subscores are present, we compute
\[
r = \frac{5r_{\mathrm{overall}} + 1.25(r_{\mathrm{code}} + r_{\mathrm{compute}} + r_{\mathrm{speed}} + r_{\mathrm{data}})}{10}.
\]
It then ranks eligible papers by \(u=\sqrt{(q/10)(1-r/10)}\), so a paper must be both scientifically strong and reasonably reproducible to rank highly. Code availability is recorded but is not a hard requirement, which allows \sw to include papers where agents must reconstruct missing pieces. The filtering counts in \figref{fig:paperscope-filter-funnel} are sequential counts from the exported snapshot, each paper attributed to its first rejecting rule.

% \vspace{-.6cm}
\begin{figure}[h]
    \centering
    \includegraphics[width=1.0\linewidth]{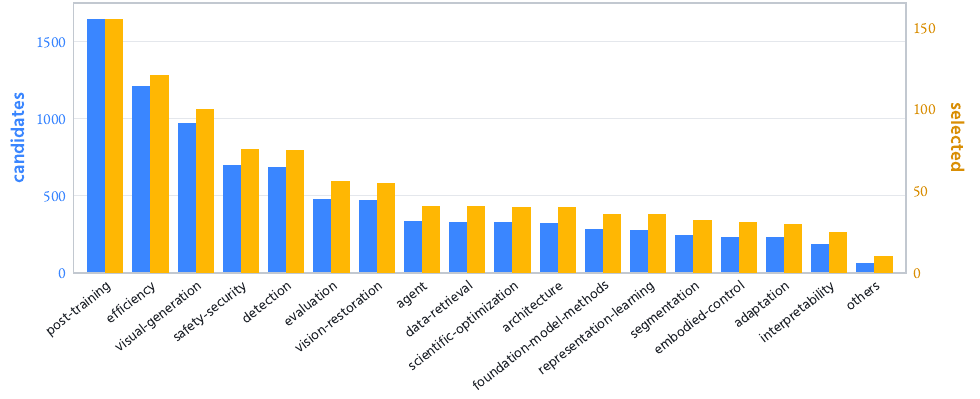}
    \vspace{-.6cm}\caption{\textbf{Temperature-balanced sampling keeps large areas prominent while preserving smaller categories.} The final 1,000-paper selection is sampled from 8,987 feasible candidates across 18 canonical categories.}
    \label{fig:paperscope-category-balance}
\end{figure}

\begin{figure}[!t]
    \centering
    \includegraphics[width=0.8\linewidth]{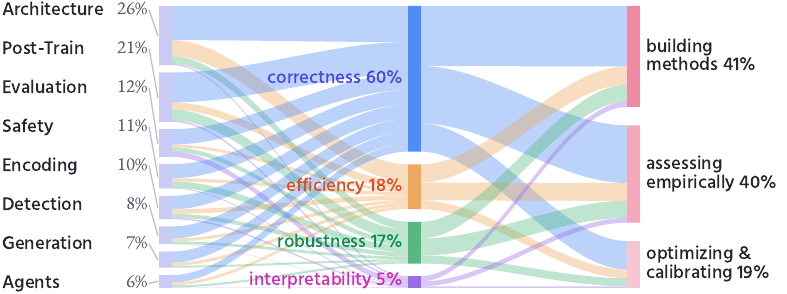}
    \caption{\textbf{What environments ask, and the work it takes.} Each environment states what should get better and what work that takes; phrases are grouped into families that emerge from data; ribbon mass is the share of environments.}
    \label{fig:ask-flow}
\end{figure}

\paragraph{Category Balancing.} The feasible pool is still highly uneven across topics, so we do not simply take the top 1,000 utility scores. We first map fine-grained taxonomy labels into 18 canonical categories, then allocate final slots using temperature-smoothed category shares with temperature \(1.2\). The largest categories remain large, but smaller categories are not squeezed out entirely. \figref{fig:paperscope-category-balance} shows the candidate pool and selected count for each canonical category. \figref{fig:ask-flow} reads the resulting environments in their own words: what each asks to make better, and the work delivering it requires.

\newpage

\section{Evaluation Protocol}
\label{sec:evaluation-protocol}

Evaluation is the scoring harness that turns the reproduced environments into comparable scores across models. Where Swarm is built for open-ended \builder coordination, Evaluation is deliberately flatter: it runs one controlled trial for each combination of paper, model, and mode. Each trial starts from a fresh workspace, receives a bounded compute budget through Nexus, drives one agent session through reason-run-feedback turns, and records the resulting train and held-out test scores. The harness splits this into auditable stages, from scaffolding through normalization (\tabref{tab:evaluation-stages}). The unit of comparison is therefore explicit: a model is judged by what its agent can build under the same problem statement, workspace boundary, scoring contract, and compute budget. We call this evaluated agent the \emph{scientist agent}, to distinguish it from the \builders that construct an environment (\secrefc{sec:swarm-roles}).

% \vspace{-.97cm}

\subsection{Evaluation Modes}
\label{sec:eval-modes}

Each environment can be run in two modes that ask different questions of a model. In \emph{improve} mode, the scientist is given the reproduced method and the train-phase scores it already achieves, and must push those numbers higher without changing the evaluation protocol; the question is whether the model can advance an existing result. In \emph{innovate} mode, the agent is given only the problem statement, the constraints and targets, the visible evaluator, and the data interface, with an empty method directory, and must build a method from first principles; the question is whether the model can solve the problem on its own.

The two modes are the problem-and-solution separation of \secrefc{sec:design-choices} made operational: the same environment, presented once with the reproduced method and once without it. This separation is built into the workspace layout, not just the prompt. The reproduced method lives in its own \texttt{method/} directory, while the problem statement, scoring targets, visible evaluator, and data interface sit outside it, and the evaluator and data never import from \texttt{method/} (\secrefc{sec:agent-workspace}). Because the rest of the environment does not depend on the method, a single change switches modes: \texttt{method/} is handed over intact for improve mode and emptied for innovate mode. They are kept separate in both the run matrix and the normalization, because a score earned by inventing a method from scratch is not the same experimental object as one earned by improving an existing method.
% that was handed over.

\subsection{Running a Trial}
\label{sec:eval-trial}

\paragraph{Workspace Boundary.} Evaluation controls what the scientist can see and modify. The problem briefing, data, baseline artifacts, and visible training evaluator are read-only. The method directory, run scripts, scratch directory, and scoring directory are writable. Held-out test files, test scripts, reference scores, and the paper text are not present during the main phase. This boundary prevents agents from tuning against the hidden test split while still giving them enough information to conduct real research: they can read the task, implement a method, run the visible evaluator, inspect failures, and iterate. In improve mode, the starting method's train score is copied into the briefing as a target to beat, but it is not used as an oracle during scoring.

\begin{table}[h]
    \centering
    \begin{softcolortable}[0.95]
    \setlength{\tabcolsep}{7pt}%
    \setlength{\softTableContentWidth}{\dimexpr\softTableWidth-4\tabcolsep\relax}%
    \renewcommand{\arraystretch}{1.45}%
    \begin{tabular}{
        >{\leftskip=9pt \raggedright\arraybackslash}m{0.26\softTableContentWidth}
        >{\raggedright\arraybackslash}m{0.74\softTableContentWidth}
        }
    \toprule
    \rowcolor{evalHeader}
    \textbf{evaluation stage} & \textbf{contract enforced by the harness} \\
    \midrule
    \rowcolor{gray!8}
    scaffold & create a fresh per-run workspace and remove files that would leak hidden tests or reference scores \\
    \rowcolor{white}
    main phase & let the scientist edit the method and submit compute jobs until it writes train-phase scores and declares completion \\
    \rowcolor{gray!8}
    validation & check that every required proposed-method metric is present; missing cells trigger bounded repair turns \\
    \rowcolor{white}
    held-out test & materialize the hidden evaluator only after the method is frozen, then run it through Nexus \\
    \rowcolor{gray!8}
    normalization & convert heterogeneous metric improvements into per-paper scores before averaging into a leaderboard \\
    \bottomrule
    \end{tabular}
    \end{softcolortable}
    \caption{\textbf{Evaluation separates iteration, validation, hidden testing, and normalization into auditable stages.} The staged harness keeps scientist behavior inspectable and comparable.}
    \label{tab:evaluation-stages}
\end{table}

\paragraph{Scientist Loop.} At the end of each turn, the scientist declares whether it wants to submit a compute job or finish. A submitted job runs the workspace's compute entrypoint through Nexus, using the requested CPU or GPU resources and optionally the short test capacity for smoke tests. Evaluation does not implement its own execution loop; it consumes the job results that Nexus exposes and resumes the same agent session with them. Current runs use a 200-turn ceiling, and a trial that fails for a transient reason before producing any score file is retried up to three times. Once scoring files exist, the workspace is kept for review rather than overwritten.

\paragraph{Train and Test Scores.} The main phase produces \texttt{scores\_train.json} using the visible evaluator. A host-side validator then checks the output against the source scoring shape: for every required proposed method, every numeric metric present in the reference reproduction must also be present in the agent's score file. 
Metrics marked as baselines are ignored because they are not the scientist's responsibility. 
% Missing train cells trigger up to five repair rounds.

After train validation passes, the method the scientist built is frozen: its directory is mounted read-only before the held-out evaluator runs, so the test score reflects exactly the method that produced the train score and cannot be tuned against the hidden split. Evaluation then materializes the held-out evaluator from a pristine clone, so the scientist cannot influence what the test runs, and executes
% \texttt{evaluate\_test.sh} 
against the frozen method.

If the test job fails, the default recovery path keeps the method and the hidden evaluator read-only for five repair turns, so the scientist can repair missing packages or regenerate artifacts without modifying the frozen method or peeking at held-out data. If those turns do not succeed, a final three-turn path makes \texttt{method/} writable again under a stricter honesty prompt, so a genuine interface mismatch can be repaired; even then the held-out evaluator and test data stay read-only and re-materialized from the pristine clone, so the method can be fixed but the hidden split still cannot be read nor the scorer altered, and these unlock turns are flagged in the trace. A run is complete only when both train and test score files validate.

\subsection{Scoring}
\label{sec:eval-scoring}

Evaluation reduces each model to one comparable number per paper, its \metric, then averages over the ten into a leaderboard. The raw measurements are heterogeneous at three levels. Within a paper, the result is split across experiments and their metrics, which sit on different scales, so one in the hundreds must not swamp one in hundredths, and carry different importance, so a metric central to the claim should count more than an incidental one. Across papers, the natural improvement ranges from a fraction of a percent to tens of points, yet every paper should contribute equal magnitude rather than let large-swing papers dominate. Across models, the comparison must stay fixed: a later model is scored against numbers frozen at publication rather than re-baselined, with the sign preserved and no value clamped. The scoring function below reconciles these levels, one cell at a time.

\paragraph{Definitions.} The scoring is built on the following objects, all fixed before any model is scored.
\begin{itemize}
\setlength{\itemsep}{2pt}
\item \underline{Reference cohort} \(\mathcal{M}=\{m_1,\dots,m_K\}\): a fixed set of \(K\) models, chosen once and frozen at publication. Each \(m_k\) is one model, and every model in \(\mathcal{M}\) is run on every paper.
\item \underline{Cell} \(c\): one metric of one method in one experiment. A single paper has many cells, spread across its experiments and the several metrics each one reports; baseline methods are excluded, leaving only the proposed methods.
\item \underline{Source value} \(s_c\): the value the paper's own method reaches on cell \(c\) once reproduced; it is the reference every model is measured against.
\item \underline{Model value} \(x_{m,c}\): the value reached on cell \(c\) by the method that model \(m\) builds.
\item \underline{Direction} \(d_c\in\{+1,-1\}\): \(+1\) when a higher value is better, \(-1\) when lower is better.
\item \underline{Prior weight} \(p_c=w_{e(c)}\,\lvert\kappa_c\rvert\): how central the cell is to the paper's claim, set by the \builder when it writes the reference scoring file, not inferred from any model's results. The agent records two numbers there: a weight \(w_e\) for each experiment, normalized to sum to one across the paper, and a coefficient \(\kappa_c\) for each metric that sets the metrics within an experiment against one another. The prior weight is their product, taking the coefficient's magnitude so that a metric's direction (already captured by \(d_c\)) does not affect its weight.
\end{itemize}
The cohort is fixed because it calibrates every paper: adding a model to it would move every paper's scale and baseline and invalidate all stored coefficients, so a later model is scored against the cohort, never added to it.

\paragraph{Cell Improvement.} The signed improvement of model \(m\) on cell \(c\) is
\[
\delta_{m,c} = d_c(x_{m,c}-s_c) =
\begin{cases}
x_{m,c} - s_c, & \text{if higher is better},\\
s_c - x_{m,c}, & \text{if lower is better},
\end{cases}
\]
positive for an improvement over the source and negative for a regression, on every metric. Every scale below is built from these improvements: how widely they spread across the cohort sets the unit for each cell, and the cohort's typical paper score sets the baseline that makes papers comparable.

\paragraph{Cohort Scale.} For each paper, mode, and phase, the cohort fixes the scale that each cell is divided by. The first quantity is the per-cell spread,
\[
\sigma_c = \operatorname{sd}_{m\in\mathcal{M}}(\delta_{m,c}) = \sqrt{\frac{1}{K-1}\sum_{m\in\mathcal{M}}\big(\delta_{m,c}-\bar{\delta}_c\big)^2},
\qquad
\bar{\delta}_c = \frac{1}{K}\sum_{m\in\mathcal{M}}\delta_{m,c},
\]
the sample standard deviation of the cohort's improvements on cell \(c\), where \(\bar{\delta}_c\) is the cohort's mean improvement on that cell. We use \(\sigma_c\) as the cell's unit because it is expressed in the metric's own scale: a metric reported in the hundreds gives a \(\sigma_c\) in the hundreds and one reported in hundredths gives a \(\sigma_c\) in hundredths, so dividing an improvement by \(\sigma_c\) removes the metric's scale and lets cells be compared. Some cells have \(\sigma_c=0\), where the cohort returns the same value and the cell carries no spread of its own; we want neither to drop these cells nor to divide by zero, so we also compute a fallback, the paper's typical spread,
\[
\tau = \operatorname*{median}_{j\,:\,\sigma_j>0}\ \sigma_j,
\]
the median of the nonzero per-cell spreads. A median, not a mean, keeps one unusually noisy cell from inflating this typical scale and shrinking every other cell. The per-cell normalizer then uses each cell's own spread when it has one and \(\tau\) otherwise,
\[
n_c =
\begin{cases}
\sigma_c, & \sigma_c>0,\\
\tau, & \sigma_c=0 \text{ and } \tau>0,
\end{cases}
\]
so a tied cell still contributes, scaled by the paper's typical spread. If even \(\tau=0\), no cell on the paper carries any spread, the paper has no usable scale, and it is excluded rather than scored with an arbitrary fallback.

\paragraph{Scale from the Cohort, not the Source.} For a cell the cohort does not tie on, the normalizer from the previous paragraph is simply the cohort spread, \(n_c=\sigma_c\). Dividing by this spread, rather than by the source value, is what places metrics of different scales on a common footing. The more obvious choice would divide each improvement by its source value, \(\delta_{m,c}/|s_c|\); but \(\sigma_c\) is computed from those same improvements, so dividing them by \(|s_c|\) scales the cell and its spread alike, and the source value cancels:
\[
\frac{\delta_{m,c}/|s_c|}{\operatorname{sd}_{m}(\delta_{m,c}/|s_c|)}
= \frac{\delta_{m,c}/|s_c|}{\sigma_c/|s_c|}
= \frac{\delta_{m,c}}{\sigma_c},
\qquad s_c \ne 0.
\]
The cohort spread, not the source value, carries the scale. Writing the standardized cell score with the general normalizer,
\[
z_{m,c} = \frac{\delta_{m,c}}{n_c},
\]
and \(s_c\) sets only the zero point of the improvement, not its scale. The score also stays well defined when \(s_c\) is zero or near zero (for example, when a metric counts the number of bugs in a code snippet), where dividing by the source value would be undefined or blow a tiny gap into an enormous ratio.

\paragraph{Effective Weights.} The prior weight \(p_c\) records how important a metric is to the paper, but not how much it tells one model apart from another. A cell on which the whole cohort lands at nearly the same value carries little signal about which model is better, while a cell on which the cohort spreads out widely both separates models and is statistically reliable. We therefore scale each prior by the cell's spread \(\sigma_c\) relative to the paper's typical spread \(\tau\), and renormalize so the weights sum to one:
\[
\tilde{w}_c
= p_c\left(\beta+\frac{\sigma_c}{\sigma_c+\tau}\right),
\qquad
w_c = \frac{\tilde{w}_c}{\sum_j \tilde{w}_j}.
\]
Here \(\beta\) is a floor we set to \(0.5\). The weight \(w_c\) this produces is each cell's share of the paper score, so scaling it by the cell's spread makes the score lean on the cells that actually separate models and discount those that do not. When the cohort ties on a cell (\(\sigma_c=0\)), every model scores it the same, so the cell cannot tell a better model from a worse one; counting it at full weight would only dilute the score with a cell that carries no signal, so the factor drops to its floor \(\beta=0.5\), halving its influence without dropping it. When a cell spreads the cohort by the paper's typical amount (\(\sigma_c=\tau\)), it carries a representative amount of signal and the factor stays at \(1.0\), leaving its prior unchanged. When the cohort disagrees on a cell far more than usual (\(\sigma_c\gg\tau\)), the cell sharply separates models and is the most informative, so the factor rises toward \(1.5\) and the cell counts for up to half again its prior. The factor saturates smoothly toward \(\beta+1\) rather than being clipped at a hard cap.

This does a different job from the normalizer: \(n_c\) puts every cell on a common scale, while \(w_c\) sets how much each cell then counts. Together they keep any single cell, an extreme-scale metric or a tied, uninformative one, from deciding a paper's score by itself. As a final backstop against a near-tied cell whose tiny spread inflates its coefficient, any cell whose contribution both outweighs all of a paper's other cells combined and exceeds one full paper score in magnitude is pruned, uniformly for every model, so no saturated metric can swing a paper's score on its own.

\paragraph{Paper Coefficients.} The cohort model's score on a paper is the weighted sum of its standardized cell scores,
\[
R_m = \sum_c w_c\, z_{m,c}.
\]
The size of \(R_m\) still varies across papers: a paper whose metrics swing widely gives every model a large \(R_m\), and would dominate a leaderboard that averaged raw paper scores. To put papers on equal footing, we divide by the cohort's typical paper-score magnitude on that paper,
\[
b = \frac{1}{|\mathcal{M}|}\sum_{m\in\mathcal{M}} |R_m|,
\]
the mean absolute \(R_m\) over the cohort. After dividing by \(b\), the cohort's typical paper score has magnitude one on every paper, so each paper contributes equally to a model's average instead of in proportion to how large its improvements happen to be. Weight, normalizer, and baseline then fold into a single stored coefficient per cell,
\[
\alpha_c = \frac{w_c}{n_c b},
\]
so a future model is scored by the frozen sum
\[
S = \sum_c \alpha_c \delta_c.
\]
All spread estimates, weight adjustments, and paper baselines are absorbed into \(\alpha_c\) and never re-estimated.

\paragraph{Leaderboard.} Train and held-out test phases, and innovate and improve modes, keep separate coefficients because they measure different experimental objects. Within each published setting, the leaderboard reports the mean paper score over published papers. The scale comes from the reference cohort, and \metric is this score scaled by 100: a paper's typical score across the cohort has magnitude 100, so a model that matches the cohort scores about 100, a model with about half again as much improvement scores near \(150\), and a model at half the cohort's gains scores near \(50\); the score is positive when the model improves over the source reproduction and negative when it regresses. Because each paper's spread across the cohort is about one, averaging over many published papers shrinks the noise in a model's mean score, so models are separated by the average of many per-paper differences rather than by any single paper; we therefore read only sustained gaps as real and treat very close scores as ties. A paper is excluded from publication uniformly for every model, never per submission, when the cohort gives no usable scale (every cell tied, or a baseline that cancels to zero) or is too small to estimate spread (fewer than three cohort models present).

\newpage

\section{Evaluation Analysis}
\label{sec:analysis}

The main text reads these environments as a leaderboard. Here we examine the evaluation behind it, asking three questions: how often a scientist actually improves on the reproduced method, whether a score reflects a real gain rather than a metric optimized during the run, and whether the number is largely a proxy for how much compute or interaction a run was given. In short: usually, yes, and no.

\subsection{How Often Scientists Beat the Paper}
\label{sec:success-rate}

\begin{figure}[t]
    \centering
    \includegraphics[width=\linewidth]{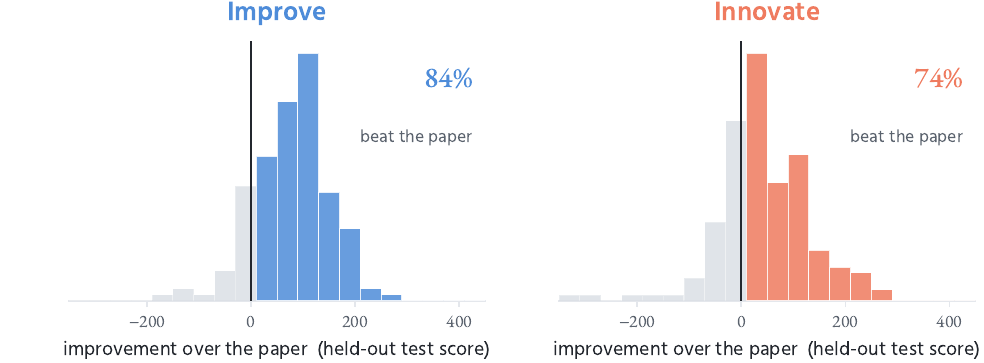}
    \caption{\textbf{How often scientists beat the paper.} Distribution of held-out improvement over the paper's source method, per mode. Colored bars (right of zero) beat the reproduced method; the grey tail regressed. The percentage is the share of runs that improved.}
    \label{fig:success-rate}
\end{figure}

Before asking which scientist is best, it is worth knowing whether any of them do real work. They do. \figref{fig:success-rate} shows the distribution of held-out improvement over the paper's source method, one panel per mode. Most runs land to the right of zero: 84\% beat the reproduced method when improving it, and 74\% do when designing a new one from scratch. Innovate carries the heavier left tail, since a fresh design more often fails outright, yet even there the typical attempt still moves the metric forward.

\subsection{Do the Gains Generalize}
\label{sec:generalization}

\begin{figure}[h]
    \centering
    \includegraphics[width=\linewidth]{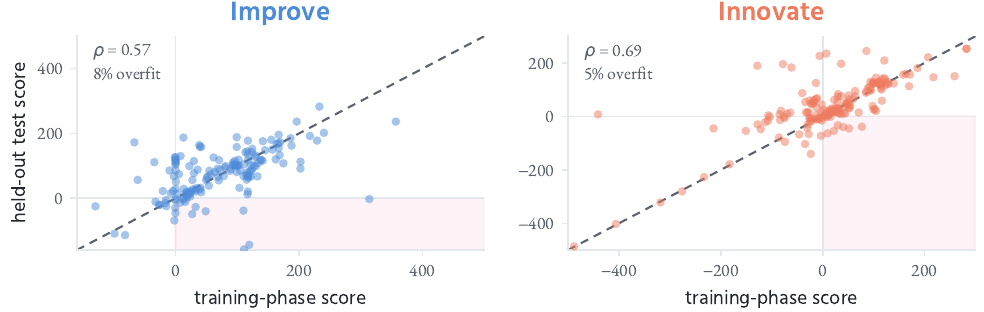}
    \caption{\textbf{Training gains carry over to held-out test.} Training-phase score against held-out test score, per mode, with the \(y=x\) line. Points hug the diagonal; the shaded quadrant holds the few runs that gained on training but regressed on test.}
    \label{fig:generalization}
\end{figure}

A score is only meaningful if it reflects a real gain rather than a metric gamed during the run. The held-out split separates the two: a scientist optimizes against a training-phase metric but is scored on a test slice it never sees. \figref{fig:generalization} plots one against the other. The two track closely (Spearman \(0.57\) in improve and \(0.69\) in innovate), so training gains usually survive to the held-out metric, and only a small share of runs fall in the overfit quadrant where a training gain reverses on test (8\% in improve and 5\% in innovate). The split rarely needs to intervene, but when it does it removes exactly the runs that would otherwise inflate a score.

\subsection{Effort and Outcomes}
\label{sec:effort}

\begin{figure}[t]
    \centering
    \includegraphics[width=\linewidth]{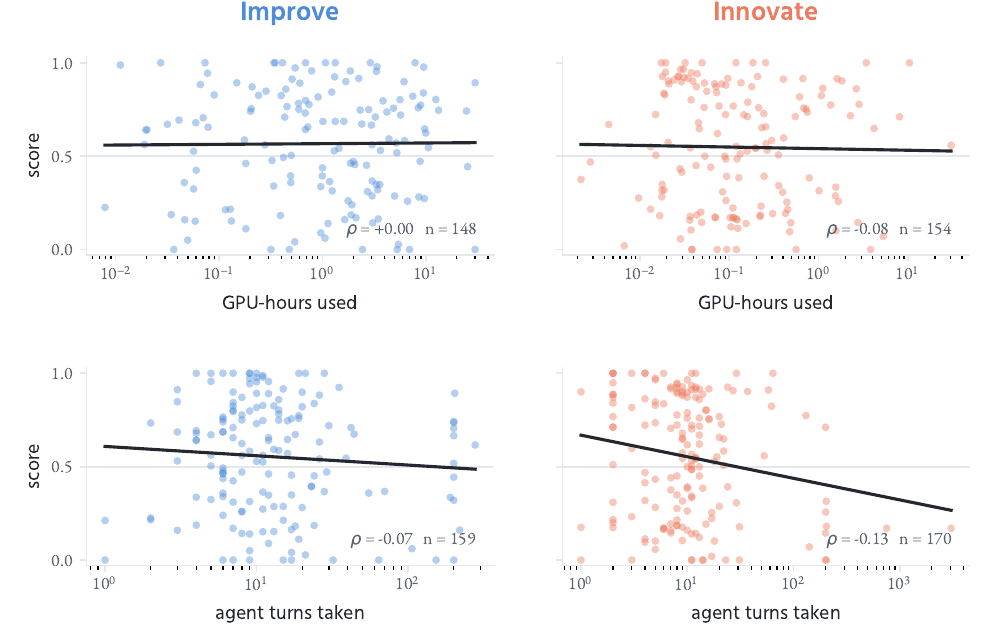}
    \caption{\textbf{Effort versus outcome.} Within-paper normalized score against GPU-hours (top row) and agent turns (bottom row), shown separately for the Improve and Innovate modes. Scores are min-max normalized within each (paper, mode); each line is a least-squares fit in log-effort. Every panel is flat (\(|\rho|\le 0.22\)).}
    \label{fig:effort}
\end{figure}

A natural worry is that a score mostly tracks how much compute or interaction a run spent, rewarding brute force over insight. It does not. \figref{fig:effort} plots each run's within-paper normalized score against the GPU-hours it consumed and the number of turns it took, with the Improve and Innovate modes kept apart. Every trend is flat: the rank correlation stays weak throughout and is slightly negative where it moves at all, at most \(0.22\) in magnitude. Spending more does not produce a better result; a better method does.

\newpage

\section{Experimental Setup}
\label{sec:experimental-setup}

This section records the concrete setup behind our experiments: the paper set the scientists attempt, the compute pool they run on, the agent scaffolds and models we compare, and how trials are run and scored.
% turned into comparable scores.

\subsection{Data}
\label{sec:exp-data}

All papers come from arXiv. We work from a global index of 124{,}816 unique papers in the categories cs.CL, cs.CV, cs.AI, and cs.LG, built from daily listings between January 1, 2025 and March 1, 2026 and exported on March 28, 2026. From this index, the selection procedure of \secrefc{sec:paper-selection} chooses a category-balanced set of 1{,}000 papers to attempt; the per-stage counts are reported there.

\subsection{Compute}
\label{sec:exp-compute}

\sw's experiments run on a heterogeneous setup provisioned through Hive and accounted by Nexus. The compute pool contains 16 A100 GPU nodes, 16 H100 GPU nodes, 16 H200 GPU nodes, and 4 CPU nodes (\tabref{tab:experimental-setup-compute}). The GPU nodes support accelerator-heavy reproduction jobs, while the CPU nodes support CPU-only experiments and non-accelerated setup or validation work.

\begin{table}[h]
    \centering
    \begin{softcolortable}[0.95]
    \setlength{\tabcolsep}{7pt}%
    \setlength{\softTableContentWidth}{\dimexpr\softTableWidth-6\tabcolsep\relax}%
    \renewcommand{\arraystretch}{1.45}%
    \begin{tabular}{
        >{\leftskip=9pt \raggedright\arraybackslash}m{0.28\softTableContentWidth}
        >{\raggedleft\arraybackslash}m{0.10\softTableContentWidth}
        >{\raggedright\arraybackslash}m{0.62\softTableContentWidth}
        }
    \toprule
    \rowcolor{computeHeader}
    \textbf{resource} & \textbf{count} & \textbf{role} \\
    \midrule
    \rowcolor{gray!8}
    A100 GPU nodes & 16 & general accelerator workloads and pilot experiments \\
    \rowcolor{white}
    H100 GPU nodes & 16 & high-throughput training and evaluation jobs \\
    \rowcolor{gray!8}
    H200 GPU nodes & 16 & memory-intensive or largest-scale reproduction runs \\
    \rowcolor{white}
    CPU nodes & 4 & low-compute, CPU-only setup and validation jobs \\
    \bottomrule
    \end{tabular}
    \end{softcolortable}
    \caption{\textbf{Hive exposes heterogeneous capacity for accelerator-heavy and CPU-only experiments.} The counts describe the compute capacity exposed in Hive.}
    \label{tab:experimental-setup-compute}
\end{table}

\subsection{Agent Scaffolds}
\label{sec:exp-scaffolds}

Each scientist reaches the workspace through a scaffold, the layer that drives a coding-agent command-line tool behind one uniform turn interface (\secrefc{sec:worker}). Our experiments center on two scaffolds: Claude Code, which runs Anthropic models, and Codex, which runs OpenAI models. The same harness also supports other agents, including Gemini, Kimi, and Grok, so a single task can be run across a range of models. We prioritize a provider's official agent scaffold when one is released; otherwise we use Claude Code as the default. Because a model and its scaffold are evaluated together and not every model runs through the same scaffold, the unit being compared is a model together with its scaffold, and a scaffold difference is part of what the comparison measures rather than something controlled away.

\subsection{Models}
\label{sec:exp-models}

We evaluate scientists built on models from seven providers: Anthropic (\model{Fable 5.1}, \model{Fable 5}, \model{Opus 4.8}, and \model{Opus 4.7}), OpenAI (\model{GPT 6 Astra}, \model{GPT 5.6 Sol}, \model{GPT 5.5}, and \model{GPT 5.4}), Zhipu (\model{GLM 5.2} and \model{GLM 5.1}), Moonshot (\model{Kimi K2.7} and \model{Kimi K2.6}), MiniMax (\model{MiniMax M3} and \model{MiniMax M2.7}), Xiaomi (\model{MiMo V2.5 Pro}), and DeepSeek (\model{DeepSeek V4}). Anthropic and OpenAI models run through their official scaffolds; the rest run through Claude Code against an Anthropic-compatible gateway (\secrefc{sec:exp-scaffolds}). Every model is run at the extra-high (xhigh) reasoning-effort setting, or at its highest available setting if it does not support xhigh, so models are judged at full strength rather than at a throttled setting.

\subsection{Evaluation Trials}
\label{sec:exp-trials}

Every trial hands the scientist the same resources, so a difference in outcome reflects the model and not the budget behind it. A trial runs under a ceiling of 200 turns and a compute budget of 30 GPU-hours, with at most 2 GPUs held at once, accounted by Nexus (\secrefc{sec:nexus}). A trial that aborts for a transient infrastructure reason, such as a gateway timeout or a compute-pool blip, is retried up to three times before it is recorded as a failure; retries cover infrastructure, never an agent that simply solved the task poorly, so no model is re-run for doing badly. We cap in-flight trials at 30 and split that capacity evenly across the active models, so a fast API does not crowd a slower one out of the queue. Each model is run in both evaluation modes of \secrefc{sec:eval-modes}. GPU-hours are counted as wall-clock time times the number of GPUs held, without weighting by accelerator class, so a trial scheduled onto faster cards can fit more experimentation into the same nominal budget; we treat this as a known limitation of expressing the budget in raw GPU-hours.

\subsection{Provider Failures}
\label{sec:exp-failures}

A run spans thousands of unattended turns over hours or days, during which a model provider can rate-limit, throttle, or go down. The two settings absorb these interruptions differently, because they optimize for different things. When building an environment, the goal is to preserve work, so a run that fails resumes from the start of the current workflow iteration, reusing the saved workspace; a provider outage then costs at most the iteration in progress, not the whole reproduction. During evaluation, the goal is fairness, so a trial that fails before producing any score is discarded and restarted from a fresh workspace rather than resumed: no model is scored on a run that was continued mid-trajectory or warm-started from partial state, and every model meets the same clean-room conditions. A failure that occurs after scores are written is set aside for review rather than silently retried.

\subsection{Cohort and Normalization}
\label{sec:exp-cohort}

Running every model on all 1{,}000 selected papers would be expensive to repeat, so the headline evaluation uses a smaller fixed cohort drawn from the lower-compute environments, chosen so that a full pass stays cheap enough to run across many models and to rerun over time. Scores are normalized against a reference set of models rather than reported in absolute terms (\secrefc{sec:eval-scoring}): the per-paper normalization coefficients are computed once from that reference set and then frozen, so a newly added model is scored against the existing board without shifting the numbers already reported for other models. The reference set is the current generation of each model family; older generations are scored against that frozen reference but kept out of it.

\subsection{Reproducibility}
\label{sec:exp-reproducibility}

Every trial starts from the same pinned container image and the same workspace template (\secrefc{sec:agent-workspace}), so the software environment and the starting state are identical across models and stay fixed as the evaluation is rerun over months. Scoring depends only on the numbers a trial records, not on wall-clock time or the hardware it ran on (\secrefc{sec:design-choices}), so a trial can be rescored later without rerunning the agent. We retain the full record of each trial, the agent log, the final workspace, and the jobs it submitted, so any reported score can be traced back to what the agent actually did.

\newpage

\section{Dataset and Training Details}
\label{sec:training-details}

This appendix documents the two artifacts behind \secrefc{sec:scientists}: the released trace corpus (\secrefc{sec:dataset}) and the scientist trained on it (\secrefc{sec:gym}). It records how the corpus is counted, how the training data and configuration are derived from it, and the held-out protocol behind \figref{fig:gym-training}.

\subsection{Trace Corpus}
\label{sec:trace-corpus}

\paragraph{What Is Preserved.} Every evaluation attempt is preserved as its own git repository, one per environment, model, and run mode. The repository is the final workspace of \secrefc{sec:agent-workspace} with its full commit history: the system commits after every turn (\secrefc{sec:version-control}), so the corpus holds the development history of each attempt rather than a cleaned final state. A trace records the research idea and its revisions, the method code as it changed, and the per-experiment logs and scores, tied together by the turn-by-turn trajectory that produced them. Attempts that regress on the source method are kept alongside those that improve on it; a corpus of research being done must include the failures.

\paragraph{Accounting.} The totals in \secrefc{sec:dataset} are computed per attempt and summed. Hours of work is the scientist's end-to-end running wall-clock time, from the first turn of a trial to its final recorded score; summed over every preserved attempt, across all models, effort tiers, and both modes, it gives the roughly 6{,}400 hours reported. The per-attempt averages of \tabref{tab:dataset} come from the same records.

\subsection{Training Data}
\label{sec:training-data}

\paragraph{Provenance.} All training data derives from finished evaluation traces. Each example is rooted in a task specification exactly as a scientist receives it: the problem, the constraints, and the metric to move (\secrefc{sec:agent-workspace}). The ten held-out papers of \secrefc{sec:idea-ablation} appear in no training example.

\paragraph{Supervised Set.} The SFT set has 1{,}000 examples. The prompt is the task specification; the target is the attempt's reasoning inside \texttt{<think>} tags followed by its final idea. Reasoning is cleaned before training: fabricated run results and job identifiers are removed, and empirical claims are reframed as conceptual, with nothing invented or added. The set combines 925 authentic completions with 75 faithful rephrasings of the strongest completions, which round the set to 1{,}000 without introducing new content.

\paragraph{Reward Set.} The GRPO set has 1{,}500 already-scored ideas, rewarded by measured outcomes only: 777 examples by the held-out test score (weight 1.0), and 723 by generalization, the test score minus the train-test overfit gap (weight 0.5). Rewards are z-normalized within each paper and mode group, so an idea is reinforced or suppressed relative to the other ideas for the same task. We deliberately exclude rewards judged by an LLM: every reward in the set is a measured result.

\subsection{Training Procedure}
\label{sec:training-procedure}

\paragraph{Setup.} The base model is \model{Qwen3-8B} in its thinking variant. Both stages train a LoRA adapter of rank 32 through Tinker~\citep{tinker}, and the GRPO stage continues from the SFT checkpoint.

\paragraph{SFT.} Cross-entropy on the assistant message only, learning rate \(10^{-4}\), batch size 8, 3 epochs, maximum sequence length 8{,}192. Training NLL decreases from 2.25 to 2.16 over the three epochs.

\paragraph{Offline GRPO.} The objective is a clipped PPO loss. The SFT policy's log-probabilities are computed with a no-gradient forward pass and used as the reference, so the clip bounds every update to a trust region around the SFT policy. Learning rate \(10^{-5}\), batch size 16, one epoch, gradient-norm clip 1.0. Training stays bounded: the completion NLL oscillates in a stable band around the SFT baseline and never diverges. This stability is not free. An earlier variant that weighted cross-entropy by the signed advantage collapsed, because tokens with negative advantage are pushed toward unboundedly low log-probability; the clipped objective with an explicit reference policy is what makes offline training on scored ideas stable.

\subsection{Held-Out Evaluation}
\label{sec:idea-ablation}

\paragraph{Protocol.} The gym result of \figref{fig:gym-training} scores ideas, not the idea generator's own coding. For each of the ten held-out papers, one idea is sampled from each of the three checkpoints, base, SFT, and GRPO, with the same prompt format, temperature 0.7, and top-p 0.8. A single fixed implementer, \model{GPT 5.6 Sol} at the xhigh effort tier, then executes every idea in innovate mode under identical conditions: the idea is injected into the briefing with the instruction to implement it faithfully and to optimize within its framing rather than invent a different approach. Only the idea's origin varies across the thirty runs.

\paragraph{Scoring and Isolation.} Scoring is identical to the benchmark: the held-out \metric, computed with each paper's frozen cohort coefficients (\secrefc{sec:exp-cohort}). The ablation is isolated from the leaderboard: it runs in a separate workspace namespace, shifts no normalization, and publishes nothing.

\begin{figure}[h]
    \centering
    \begin{minipage}[t]{0.46\textwidth}
        \begin{minipage}[c][2.16in][c]{\linewidth}
            \centering
            \includegraphics[width=0.8748\linewidth]{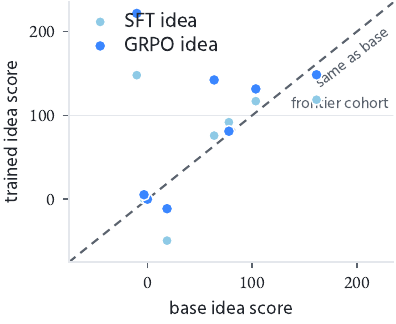}
        \end{minipage}\\
        \captionof{figure}{\textbf{Trained ideas against the base.} Each of the ten held-out papers contributes one SFT and one GRPO dot, plotted against the same paper's base idea score (displayed scaled by 100). A dot above the dashed diagonal means the trained idea beat the untrained idea on that paper.}
        \label{fig:gym-appendix}
    \end{minipage}\hfill
    \begin{minipage}[t]{0.46\textwidth}
        \begin{minipage}[c][2.16in][c]{\linewidth}
        \centering
        \begin{softcolortable}[1]
        \setlength{\tabcolsep}{7pt}%
        \setlength{\softTableContentWidth}{\dimexpr\softTableWidth-8\tabcolsep\relax}%
        \begin{tabular}{
            >{\leftskip=9pt \raggedright\arraybackslash}m{0.28\softTableContentWidth}
            >{\raggedleft\arraybackslash}m{0.22\softTableContentWidth}
            >{\raggedleft\arraybackslash}m{0.22\softTableContentWidth}
            >{\raggedleft\arraybackslash}m{0.22\softTableContentWidth}
            }
        \toprule
        \rowcolor{datasetHeader}
        \textbf{statistic} & \textbf{base} & \textbf{SFT} & \textbf{GRPO} \\
        \midrule
        \rowcolor{gray!8}  median & $41.2$ & $83.9$ & $106.3$ \\
        \rowcolor{white}   mean & $51.5$ & $62.8$ & $89.9$ \\
        % \rowcolor{gray!8}  std.\ dev. & $57.2$ & $65.9$ & $79.7$ \\
        \bottomrule
        \end{tabular}
        \end{softcolortable}
        \end{minipage}\\
        \captionof{table}{\textbf{Ablation summary statistics.} Median, mean, and standard deviation of the held-out score over the ten held-out papers, displayed scaled by 100.}
        \label{tab:gym-summary}
    \end{minipage}
\end{figure}

\paragraph{Results.} \figref{fig:gym-appendix} plots each trained idea's score against the same paper's base idea score, and \tabref{tab:gym-summary} summarizes the distributions over the ten held-out papers. Both the mean and the median rise monotonically from base to SFT to GRPO. The spread grows as well, so the gain is a shift in the typical outcome rather than a uniform improvement on every paper.

\newpage 

\input{example_env.tex}

\newpage

\input{prompts/prompts.tex}

\end{document}

%% file: example_env.tex
\section{Example Produced Environment}
\label{sec:example-env}

This section uses one finished environment as a lens on what the \builders actually do, and why. For each choice a \builder makes while turning a paper into a reusable task, we ask why it makes that choice and whether the behavior is one the system is built to elicit or the \builder's own judgment. The paper is only background: it is the reproduction of \emph{Transformers without Normalization} (arXiv:2503.10622), whose Dynamic Tanh (DyT) replaces Layer Normalization with a learnable element-wise operation. The run used \model{MiniMax M2.7}, \model{GLM 5.1}, and \model{GPT 5.4} across 25 agent sessions over the reproduce, design, package, and verify roles, and about 19 GPU-hours, ending at the \texttt{core\_claim} milestone. The few excerpts below come from the produced workspace; layout follows \figref{fig:workspace-structure}.

\paragraph{Implement the real method, not a lookalike.} The reproducer writes the paper's actual computation rather than a model tuned to emit its published numbers. It does this because the milestone rewards a method that runs: a surrogate fit to the reported accuracy would clear a naive number-matching test but fails ours by construction, and the verifier re-checks that the code is the algorithm itself. This is designed in --- the milestone ladder and the verify role reward a runnable method (\secrefc{sec:swarm}).

\begin{filebox}{accBlue}{method/dyt.py}
def forward(self, x):                  # DyT(x) = gamma * tanh(alpha*x) + beta
    return self.gamma * torch.tanh(self.alpha * x) + self.beta

def replace_ln_with_dyt(model, init_alpha=0.5):
    for name, m in list(model.named_modules()):
        if isinstance(m, (nn.LayerNorm, nn.RMSNorm)):   # swap every norm layer
            set_submodule(model, name, DyT(m.normalized_shape[-1], init_alpha))
    return model
\end{filebox}

\paragraph{Write the briefing so it cannot leak the answer.} The \builder states the problem in \emph{method-agnostic} terms: it fixes the question and the success criteria but never names DyT or normalization-free training. Why withhold what it just implemented? Because the same workspace is later handed to a from-scratch scientist, and a brief that named the solution would let that scientist copy it instead of doing research. By design, briefings must be method-agnostic, and the verifier and \texttt{validate.py} flag leaks; one is caught below.

\begin{mdbox}{accPurple}{briefing/problem.md}
## Research Question
Can deep transformer-style classifiers be trained stably and accurately when
their usual feature-normalization components are removed or replaced, without
winning by increasing model size, changing the data, or using pretrained weights?

## Success Criteria
- Train without numerical divergence; match or improve the reference accuracy.
- Keep parameter count and model scale matched within the comparison.
- Same data split and preprocessing; no pretrained weights, extra data, or ensembles.
\end{mdbox}

\paragraph{Decide what counts as an improvement.} The scoring contract is where the \builder exercises judgment the system asks for but does not dictate. It makes top-1 accuracy the primary metric, keeps top-5 and balanced accuracy as diagnostics, and gives parameter count a \emph{negative} coefficient --- so beating the baseline by scaling the model up does not register as a win. The format invites such a constraint metric; which metrics matter, and by how much, is the \builder's call. It records the paper's reported numbers as the ranking reference and its own measured numbers as the baseline a future method must beat.

\paragraph{Scale the experiment down without trivializing it.} The full paper trains ViT-B/L for 300 epochs, beyond a per-paper budget, so the \builder scales to a 50-class slice and a ViT-S/16. The choice is bounded by a rule the system cares about: model and data scale may shrink, but the algorithm and the matched comparison must be preserved, and the task must stay hard enough to discriminate. The \builder checks this --- the LayerNorm baseline lands below saturation, so comparison still has room, and DyT edges it at equal parameter count.

\paragraph{Pin the surface a future scientist competes on.} Because the environment is graded by comparison, the \builder writes down what a future scientist may optimize, what it must hold fixed, and where it may explore. It splits this into three documents: a \emph{target} (what improvement means and how it is measured), \emph{constraints} (what must stay matched across compared systems), and a \emph{direction} (the mechanism to work on, and what is out of bounds). These are required and method-safe: a from-scratch reader sees the rules, not the solution.

\begin{mdbox}{accIndigo}{scoring/TARGETS.md}
## Primary: supervised_classification
Improve accuracy on the matched ImageNet class slice without increasing model
scale or changing the data.
- Benefit: top1_accuracy (top5_accuracy, balanced_accuracy as diagnostics)
- Constraint: param_count_M within 5% of the matched reference
- Success: top-1 within 1 pt of the reference, with no parameter increase
\end{mdbox}

\begin{mdbox}{accRose}{scoring/CONSTRAINTS.md}
## Model Scale
Candidate and reference must have matched scale: param_count_M within 5%.
A larger candidate is an extra experiment, not an improvement.

## Training Budget
Same data, epoch budget, optimizer, augmentation, and preprocessing. A recipe
change counts only when it is applied to every system in the comparison.
\end{mdbox}

\begin{mdbox}{accTeal}{scoring/DIRECTION.md}
## Direction Scope
Work on normalization as a mechanism: stabilize activations, gradients, or
optimization while keeping the compared architecture and data fixed.

## Out of Bounds
- Resource scaling as the only change   - Pretrained weights or extra data
- Ensembles   - Switching to an easier or different dataset
\end{mdbox}

\paragraph{Keep evaluation independent of any one method.} The \builder makes the evaluator score prediction files and own the labels, rather than load the reproduced model. Why give up the convenience of calling its own checkpoint? Because a future method, however different from DyT, must score through the same interface; an evaluator coupled to this run's model would break for everyone else. The boundary is designed --- and, as the next step shows, enforced rather than assumed.

\paragraph{Build a held-out test, and say where it is weak.} The designer adds a hidden test slice so a later scientist cannot quietly overfit the visible scores, and --- notably --- writes down why it chose this split and where it falls short. The choice of split is the \builder's own judgment; that it records the trade-off is the behavior the design-notes step is meant to produce.

\begin{mdbox}{accAmber}{DESIGN\_NOTES.md}
### Split style: per-class alternating validation parity
Within each class, even-position images go to train and odd-position to test.

Why this fits: both slices have comparable difficulty, so an honest method
scores similarly on both, while a method that games the visible slice shows a gap.
Known limitation: the parity split is moderate. A harder split (held-out classes
or OOD) gives a stronger shortcut signal but needs a larger dataset.
\end{mdbox}

\paragraph{Catch the system's own mistakes.} The last role is where design meets behavior. The verifier audited the artifact and found that an earlier \builder had let the evaluator load the reproduced checkpoint --- a method leak that would silently break from-scratch mode. It moved inference into \texttt{method/} and rewrote \texttt{eval/} as a generic prediction-file scorer. The system is built to catch exactly this class of error (the verify role together with \texttt{validate.py}), and here it did.

\begin{mdbox}{accGreen}{VERIFICATION.md}
## Issues found
- Critical, fixed: eval/ contained solution-specific checkpoint loading. Even
  without importing from method/, this leaked the reproduced algorithm into the
  reusable evaluation path and would compromise from-scratch mode.

## Fixes applied
- Moved checkpoint inference into method/; rewrote eval/ as a generic evaluator
  that scores JSONL predictions from any method without knowing the algorithm.
\end{mdbox}

Only after this check passed did the workspace settle at \texttt{core\_claim}. Every choice above is committed: the run writes a \texttt{manifest.json} and a \texttt{README.md} reproduction log, and the git history checkpoints each turn, so each decision traces back to the \builder, the jobs, and the code behind it (\secrefc{sec:agent-workspace}).

%% file: prompts/prompts.tex
% AUTO-GENERATED by prompts/generate_prompts.py -- do not edit by hand.
\section{Agent Prompts}
\label{app:prompts}

This section shows representative prompts from each stage of the system, to illustrate how agents are instructed: the screening and analysis prompts that grade and select papers, the role prompts that drive a Swarm reproduction, and the prompts that run a single evaluation trial. The prompts are excerpted; operational scaffolding repeated across prompts (job submission, progress logging, version control, the workspace layout) is omitted and described in sections above. Placeholders in braces are filled at run time.

\input{prompts/sec_selection.tex}

\input{prompts/sec_swarm.tex}

\input{prompts/sec_eval.tex}

%% file: prompts/sec_selection.tex
% Generated by generate_prompts.py (coarse pass) then HAND-CURATED.
% Re-running the generator reverts manual trims.
\subsection{Prompts for Paper Grading and Selection}
\label{app:prompts-selection}

\begin{promptbox}{promptC1}{Prefilter}
\psec{Paper Pre-Filter}

Read the paper excerpt below and rate its \textbf{research quality} on a scale of 0-100.

\begin{itemize}
\tightlist
\item
  \textbf{100}: Landmark paper. Defines a new field or sets a new paradigm. Must-read for anyone in ML/AI.
\item
  \textbf{80}: Top-venue quality (ICLR, NeurIPS, ICML, ACL, EMNLP, CVPR). Novel idea, strong experiments, clear results. Worth reading and citing.
\item
  \textbf{60}: Second-tier venue quality. Solid work with experiments, but less novelty or narrower scope. Worth reading in your area.
\item
  \textbf{40}: Has a method and some experiments, but limited novelty or weak evaluation. You'd skim it.
\item
  \textbf{20}: Minimal contribution. Shallow, incremental, or poorly executed. Not worth reading.
\item
  \textbf{0}: Not a research paper, or no method and no experiments.
\end{itemize}

These anchors are reference points, not buckets. Use the full range freely.
Scores like 7, 23, 41, 58, 73, 84, and 97 are all valid when appropriate.
Do not restrict yourself to only 0, 20, 40, 60, 80, or 100.

Note: The excerpt below is the first \textasciitilde15,000 characters of the paper, converted to markdown, intentionally truncated. Formatting artifacts (broken text, missing figures, odd symbols) are from automatic PDF-to-markdown conversion --- ignore them and judge only the research content.

Think briefly about the paper's contribution, novelty, and experimental rigor.
Give a short explanation of the main strengths and weaknesses.
Then provide an overall score from 0 to 100, using any integer in that range.
The score should reflect the paper's actual research quality, not just which
anchor it is closest to.

\psec{Paper Excerpt}
\end{promptbox}

\begin{promptbox}{promptC3}{Analyst}
\psec{PaperScope --- Paper Cataloging Prompt}

You are a research paper analyst. Given a paper, produce a \textbf{structured catalog entry} of 13 fields whose purpose is to assess whether and how hard the paper can be reproduced. The fields cover paper type, novelty, taxonomy, compute requirements, paid resources, benchmarks, datasets, baselines, dependencies, code availability, reproduction difficulty, paper quality, and problem novelty. Every score is an integer on a \textbf{0--10} scale, grounded in specific paper content; when uncertain, say so rather than guess.

\begin{center}\rule{0.5\linewidth}{0.5pt}\end{center}

\psec{Field 1: Paper Type}

Classify into one primary type, with justification:

\begin{itemize}
\tightlist
\item \textbf{metric\_method} --- introduces or improves a method, evaluated with measurable results.
\item \textbf{benchmark\_dataset} --- the contribution is a benchmark, dataset, or evaluation suite.
\item \textbf{analysis\_findings} --- an empirical discovery or insight rather than an improvement.
\item \textbf{theoretical} --- a formal result: proofs, bounds, complexity analysis.
\item \textbf{survey\_position} --- aggregates or opines on existing work; no original experiments.
\end{itemize}

\promptellipsis

\psec{Scoring: Reproduction Difficulty, Paper Quality, Problem Novelty}

The decisive fields. \textbf{Reproduction difficulty} --- split into coding, compute, runtime, and data sub-scores plus an overall --- estimates the cost to implement the method, run it, and evaluate the main claims, judged as the AI agent that would reproduce it rather than a human. \textbf{Paper quality} rates the contribution as a reviewer would. \textbf{Problem novelty} rates how interesting the question is, independent of the method. Each is a 0--10 integer with a short justification.

\promptellipsis
\end{promptbox}

%% file: prompts/sec_swarm.tex
% Generated by generate_prompts.py (coarse pass) then HAND-CURATED.
% Re-running the generator reverts manual trims.
\subsection{Prompts for Swarm}
\label{app:prompts-swarm}

\begin{promptbox}{promptC4}{Reproduce Start}
\psec{Reproducer Agent}

You are a Reproducer agent. Your job is to read a research paper, reproduce its core results, and package the reproduction as a reusable environment.

Your reproduction does not need to match the paper's exact numbers, but you must implement the paper's actual method and demonstrate that it works as claimed. Fitting surrogates to published data, hardcoding paper values, or building shortcuts that bypass the method are not reproduction --- they are fabrication.

\textbf{You are fully autonomous. There is no human in the loop. Do not ask for confirmation, do not wait for direction, do not say "let me know" or "please confirm". Analyze the situation, make your own decisions, and act immediately. Every turn you spend not making concrete progress (editing code, creating scripts, fixing bugs) is wasted.}

You should launch multi- or sub-agents to accelerate the implementation.

\textbf{Your minimum goal is \texttt{core\_claim}. Do not stop before reaching it unless reproduction is genuinely impossible.}

You may receive a newly initialized workspace, or one that already has partial or substantial implementations. In either case, your job is to push the reproduction forward --- not just audit what exists.

\begin{center}\rule{0.5\linewidth}{0.5pt}\end{center}

\promptellipsis

\psec{Workflow}

Work through these phases. You decide when to move between them --- they are not hard boundaries.

\psecsub{Phase A: Paper Analysis (no GPU needed)}

\begin{enumerate}
\def\labelenumi{\arabic{enumi}.}
\tightlist
\item
  Read the paper thoroughly (both markdown and PDF)
\item
  Identify: dataset, method, evaluation metric, baseline numbers, hyperparameters
\item
  Check if the paper links to a code repository
\item
  Check if the paper has a code release --- if so, clone it into \texttt{method/} and study it
\item
  Plan your implementation --- whether building on existing code or writing from scratch, the paper's method goes in \texttt{method/}
\item
  Write an initial plan as comments or a brief README
\item
  Fill in the \texttt{briefing/} folder --- \texttt{problem.md} and \texttt{evaluation.md} first (method-agnostic), then \texttt{method.md}, and finally \texttt{overview.md} (short summary)
\item
  Fill in \texttt{scoring/reference.json} with the paper's reported numbers --- experiments, metrics, baselines, and the proposed method (see Deliverables for format)
\end{enumerate}

\psecsub{Phase B: Environment Setup (light compute)}

\begin{enumerate}
\def\labelenumi{\arabic{enumi}.}
\tightlist
\item
  Acquire the dataset: check the shared datasets directory first, download if missing (use shared for general datasets, workspace \texttt{data/} for paper-specific ones)
\item
  Download pretrained models if needed (use shared models directory for general weights, workspace for paper-specific ones)
\item
  Write data loading and preprocessing code
\item
  Write the evaluation script (compute the paper's metric)
\item
  Validate \texttt{scoring/reference.json} by running \texttt{python\ validate.py}
\item
  Verify on a trivial baseline (e.g., random predictions) that the eval pipeline works
\item
  Start drafting the container definition and task description for the environment
\end{enumerate}

\psecsub{Phase C: Reproduction (GPU, iterative)}

Plan the reproduction incrementally --- start from the minimum, expand if budget allows:

\begin{enumerate}
\def\labelenumi{\arabic{enumi}.}
\tightlist
\item
  \textbf{Plan first.} Break the reproduction into ordered steps from minimum to complete.
\item
  \textbf{Reproduce the minimum that supports the core claim.} A single experiment on one setting --- just enough to show the method works.
\item
  \textbf{Expand incrementally.} Each step adds more coverage --- more settings, more comparisons, more of the paper's results. But only expand after the minimum is confirmed working --- validating that the idea works matters far more than matching the paper's full scale.
\item
  \textbf{Mark progress.} Report milestones (see below) after each turn.
\end{enumerate}

When expanding, prefer adding a new axis of variation over running more of the same single-setting experiment. The relevant axes depend on the paper: dataset, problem instance, input difficulty, environment, backbone, hyperparameter regime, task type --- anything the paper itself varies. A small matrix across two axes is far more informative than two runs at the same condition, and it gives downstream agents a real surface to improve on.

\begin{center}\rule{0.5\linewidth}{0.5pt}\end{center}

\promptellipsis
\end{promptbox}

\begin{promptbox}{promptC5}{On Designer}
\psec{Designer --- Build the Train/Test Evaluation}

You are the \textbf{designer}. Previous agents reproduced a paper and produced \texttt{scoring/scores.json}. Your job is to design a two-part evaluation --- a train slice and a held-out test slice --- that future agents will be tested against.

You might be resuming from a previous designer's work, or a repo not yet started on this part.

\textbf{Training and test should be roughly equal in scale.}

\begin{center}\rule{0.5\linewidth}{0.5pt}\end{center}

\psec{What this workspace is becoming}

This workspace will ship as a \textbf{reusable benchmark and gym environment}. Future AI scientist agents will operate on it in two modes:

\begin{itemize}
\tightlist
\item
  \textbf{Improve mode}: the agent gets everything (problem + method + code + results + data + documentation) and must \textbf{beat the current scores}.
\item
  \textbf{From-scratch mode}: the agent gets only the problem definition (problem + eval + data + constraints + direction) and must invent their own method.
\end{itemize}

In both modes, the agent optimizes against an evaluation suite (what in \texttt{eval/}). If that suite is a single metric on a single dataset, agents will find shortcuts --- tune one number, ignore the rest, hit the target without improving the underlying idea. We want to measure \emph{real} improvement, not metric-hacking.

Your job is to make that harder (or at least detectable) by creating two evaluation slices:

\begin{itemize}
\tightlist
\item
  \texttt{eval/train/} --- the slice future agents see during development. They optimize against this.
\item
  \texttt{eval/test/} --- held out from future agents. They never see it. When they signal "done", we run \texttt{eval/test/} on their final method. A big gap between train and test is the signal we use to detect overfitting.
\end{itemize}

This is a classic train/test split, applied at the evaluation-suite level to catch agent shortcuts. \textbf{You are evaluating future agents' work, not the paper's claim.} You are not trying to break the paper. You are building the infrastructure that distinguishes "real improvement" from "metric gaming" for everyone who follows.

\begin{center}\rule{0.5\linewidth}{0.5pt}\end{center}

\psec{What to produce}

Six new files / directories total --- two new eval subfolders, two new score files, two new shell entry-points (existing files are read-only, except \texttt{reference.json}, which you may extend with new metric entries only):

\begin{lstlisting}[style=promptcode]
eval/
|-- <existing eval files  DO NOT MODIFY>
|-- train/                    <- new, self-contained, what future agents see
`-- test/                     <- new, self-contained, held out from future agents

scoring/
|-- reference.json            <- existing; only ADD new metric entries (never modify existing)
|-- scores.json               <- existing, DO NOT MODIFY
|-- scores_train.json         <- new, produced by scripts/evaluate_train.sh
`-- scores_test.json          <- new, produced by scripts/evaluate_test.sh

scripts/
|-- evaluate.sh               <- existing, DO NOT MODIFY
|-- evaluate_train.sh         <- new, mirrors evaluate.sh's contract
`-- evaluate_test.sh          <- new, mirrors evaluate.sh's contract
\end{lstlisting}

\textbf{THESE SCORES MUST BE PRODUCED BY A REAL RUN. YOU CANNOT GUESS OR HARDCODE ANY NUMBERS IN THEM.}

Format follows the existing \texttt{scoring/scores.json} and \texttt{scoring/reference.json} conventions. Each metric entry should have at least a \texttt{value:} field. If \texttt{reference.json} uses \texttt{coefficient} to weight metrics, carry those coefficients into both new files consistently so the aggregates are comparable.

Plus two new shell entry-points alongside the existing \texttt{scripts/evaluate.sh} (do NOT modify \texttt{evaluate.sh}):

\begin{itemize}
\tightlist
\item
  \texttt{scripts/evaluate\_train.sh} --- runs \texttt{eval/train/} on a method's pre-generated outputs and writes \texttt{scoring/scores\_train.json}
\item
  \texttt{scripts/evaluate\_test.sh} --- runs \texttt{eval/test/} on the same outputs and writes \texttt{scoring/scores\_test.json}
\end{itemize}

These follow the \textbf{same contract as the existing \texttt{scripts/evaluate.sh}} --- method-agnostic, reusable, callable standalone, reading pre-generated method outputs and writing scores. Mirror \texttt{evaluate.sh}'s argument shape and input/output conventions exactly --- a future agent should be able to swap in a different method and run \texttt{scripts/evaluate\_train.sh} / \texttt{scripts/evaluate\_test.sh} without code changes on their side.

\textbf{\texttt{evaluate\_test.sh} must score every method present in \texttt{scores\_train.json} (never one hardcoded name), write to \texttt{scores\_test.json}, and invoke the \texttt{eval/test/} path.} A copy of \texttt{evaluate\_train.sh} that still points at \texttt{eval/train/} silently destroys the anti-shortcut signal.

\promptellipsis

\promptellipsis

\psec{Design principles}

\psecsub{Metrics measure the scientific quantity, NEVER agreement with the paper}

\textbf{This is the single most important rule.} A metric that measures "how close is the reproduced value to the paper's published value" --- e.g., \texttt{mean\_abs\_rel\_error\_author\_X}, \texttt{deviation\_from\_paper\_X}, \texttt{\textbar{}value\ -\ 0.87\textbar{}}, \texttt{fit\_to\_published\_Y} --- is categorically forbidden as a scored metric. The gym is built FROM the paper, but we are \textbf{never} trying to match the paper.

The paper is \textbf{context}, not a target. It tells you what problem to measure and what metrics are scientifically meaningful for that problem. It does NOT tell you what numbers the score should reach. Future agents may produce a method that scores higher than the paper (real improvement), lower than the paper (regression), or that invents an entirely different approach --- every one of those outcomes must be visible in the score.

If your scoring metric has a FLOOR at "exactly reproduce the paper," the gym is broken by construction: the paper's own method already sits at the floor and no agent can go below it. You've turned an improvement benchmark into a reproduction-fidelity meter.

What TO score:

\begin{itemize}
\tightlist
\item
  The scientific quantity itself (accuracy, loss, F1, calibration error, similarity, detection rate, reward, etc.), computed on the paper's evaluation protocol. Avoid wall-clock-dependent quantities such as latency or runtime; use a machine-independent substitute (FLOPs, parameter count, training steps, samples processed) instead, since the verifier rejects scored wall-clock-speed metrics.
\item
  Continuous metrics whose range spans both directions of the paper's claim --- an agent that beats the paper's reported X=0.87 with X=0.95 must score higher; an agent that scores X=0.80 must score lower.
\item
  Ratios, directional differences, or any quantity where "the paper's number" is simply one point on a continuum that future methods can move along.
\end{itemize}

What NOT to score:

\begin{itemize}
\tightlist
\item
  \texttt{\textbar{}reproduced\ -\ paper\_published\textbar{}} (absolute error vs paper)
\item
  \texttt{mean\_abs\_rel\_error\_author\_*} or any metric whose name includes \texttt{author}, \texttt{paper}, \texttt{published}, \texttt{vs\_paper}, \texttt{deviation\_from\_paper}, \texttt{reproduction\_error}, or equivalent wording.
\item
  Any metric whose theoretical floor equals "the paper's published value." If matching the paper gives 0 and lower-is-better, you have built the forbidden thing --- even if the metric name hides it.
\end{itemize}

The paper's numbers belong in \texttt{scoring/reference.json} under \texttt{results.\textless{}method\_name\textgreater{}} as the baseline against which future agents are ranked. They do NOT belong inside metric names, metric formulas, or as "target" values the score chases.

\psecsub{Anti-shortcut, not anti-paper}

A well-reproduced method should score \textbf{similarly on train and test}. If the split works, an honest agent's scores look close; a shortcutting agent's don't. Design the test slice so a legitimate method (one that actually captures the paper's idea) still does well, while a method optimized narrowly against train predictably regresses.

\textbf{Do not hardcode which method to evaluate.} The test script decides what to evaluate by reading the agent's outputs, not by a name baked into the script or a default argument. If your evaluator only evaluates one name, a future agent's improvements will never reach it and every improve-mode run looks identical on test --- silently.

\psecsub{Train and test must be calibrated to comparable difficulty}

The split only works if train and test probe the method at \textbf{roughly the same level of rigor}. If test is drastically harder than train by construction, every method will score worse on test --- that gap comes from your design, not from overfitting, and it tells you nothing. If test is drastically easier, everyone ceilings out and there's nothing to compare.

The calibration target: on the \textbf{existing final reproduction}, train and test should produce scores that are meaningfully comparable side-by-side. A small, expected gap is fine; a huge collapse means your split is miscalibrated, not that the reproduction is bad.

When using a split style that intrinsically shifts difficulty (difficulty stratification, OOD/perturbed, held-out subgroup), keep the shift \textbf{moderate} --- enough to surface methods that don't generalize, not so much that even a legitimate method breaks. Sanity-check by running the existing final checkpoint on both slices before you declare the split final: if train scores are strong and test scores collapse, back off and re-design.

Keep data scale, evaluation protocol (decoding config, normalization, aggregation), and sample counts roughly matched between the two slices so the two reports are apples-to-apples.

\promptellipsis

\psecsub{Self-contained (critical)}

\textbf{Core idea: once a future agent has optimized against \texttt{eval/train/} and signaled done, running \texttt{eval/test/} on their method must be a zero-effort step. No retraining. No re-running expensive experiments. No code changes on the agent's side. Just point \texttt{eval/test/} at the method (and result) they already produced, through the same interface \texttt{eval/train/} used, and score.} As if we got a score by running the train set.

\texttt{evaluate\_test.sh} should be \textbf{lightweight} given that the method has already been run for \texttt{evaluate\_train.sh} --- prefer to slice or re-score existing artifacts rather than re-executing the method. If you do need to call the method again (e.g. for a stochastic-seed split), make sure the extra cost is small.

This means \texttt{eval/test/} has to go through the \textbf{same pre-agreed method interface} that \texttt{eval/train/} uses --- the contract that already exists here between \texttt{method/} and \texttt{eval/} (a function signature, a CLI like \texttt{scripts/method.sh}, a predictions file at a fixed path, etc.). The agent writes their method to satisfy that interface using \texttt{eval/train/} as the visible spec. \texttt{eval/test/} then calls the same interface and scores --- nothing else required from the agent.

If \texttt{eval/test/} couples to anything specific to the current \texttt{method/} (an internal class, a particular checkpoint format, a file that only this training run happens to write), a future different method won't satisfy it --- and the test breaks through no fault of the agent.

The set of keys under \texttt{experiments.\textless{}exp\textgreater{}.results} (or the equivalent method-name axis) must match between \texttt{scores\_train.json} and \texttt{scores\_test.json}. When an agent adds a new method variant on train, the test script must automatically score it on test --- not by re-running \texttt{evaluate\_test.sh} with a different name, but because the script discovers and iterates over whatever methods it finds. The method-name contract is enforced by the script's enumeration logic, not by a caller remembering to pass the right argument.

\psecsub{No leakage between train and test}

\texttt{eval/train/} and \texttt{eval/test/} must be \textbf{independent}. \texttt{eval/train/} must not import from \texttt{eval/test/}, and vice versa. Do not put shared helpers inside either folder that the other reads --- if you need shared code, duplicate it (the two folders are small) or keep it outside both. A future agent reading \texttt{eval/train/} to optimize must not be able to learn the test set's metrics, data split, or scoring details by following an import.

\texttt{scoring/reference.json} is fine --- it may document both slices; that's expected. Top-level \texttt{eval/} files (outside \texttt{train/} and \texttt{test/}) should \textbf{preferably} not reveal test-specific metric/data shape either --- if you can avoid it, do.

\psecsub{Do NOT retrain. Do NOT modify the method.}

The method is already built by prior agents. You are only \textbf{re-running inference + scoring} on the existing final checkpoint. If there is no usable checkpoint or \texttt{scoring/scores.json} is missing, run the minimal inference path to produce scores --- but do not retrain, change hyperparameters, or tweak the method.

You are encouraged to use as little computing resource as possible. In most cases, you do not need to rerun massive experiments. For example, the previous result might be already saved, and you only need to re-evaluate some rubric-based scores or the model is already trained, and you only need to do some lightweight evaluation.

\begin{center}\rule{0.5\linewidth}{0.5pt}\end{center}

\promptellipsis
\end{promptbox}

\begin{promptbox}{promptC6}{On Packager}
\psec{Packager --- Make This Repo Reproducible From a Fresh Clone}

You are the \textbf{packager} for this paper reproduction. Previous agents built the method, ran experiments, and produced scores. Your job is a targeted audit: ensure anyone can reproduce this work by cloning the repo and running \texttt{scripts/download.sh} followed by the method.

You do NOT redesign the method. You do NOT change algorithms, training loops, loss functions, eval metrics, or scoring code. You fix \textbf{distribution} issues so a fresh clone actually works.

This is a focused task: you check the paths and the download command lines so the environment rebuilds from a clean clone. Do not read files you do not need, to avoid wasting time.

You may use the command line for light checks --- \texttt{ls}, \texttt{grep}, \texttt{head}, \texttt{curl\ -I} (HEAD request only), \texttt{huggingface-cli\ download\ -\/-include\ config.json}, and similar. \textbf{Never run heavy commands. Never actually download datasets or model weights.} You are a checker, not a fetcher. A HEAD request that returns 200 is enough to validate a URL --- pulling the full file is unnecessary. If something you run takes more than a few seconds or produces more than a few KB of output, you are doing the wrong thing.

\begin{center}\rule{0.5\linewidth}{0.5pt}\end{center}

\psec{What to check}

\begin{enumerate}
\def\labelenumi{\arabic{enumi}.}
\item
  \textbf{Every needed \emph{pre-existing} artifact has a REAL, EXECUTABLE fetch command in \texttt{scripts/download.sh}.} Walk the repo. For every path the code reads from that lives inside a \texttt{.gitignore}'d directory (e.g. \texttt{data/downloads/}, pretrained-model dirs), \texttt{download.sh} must include an actual shell command (not a comment, not an echo message) that fetches it from a public source. Few times exceptions occurs, like it uses synthetic data, where other part of the repo handles it.

  \textbf{Only artifacts that must exist before the method runs are your responsibility}: datasets, pretrained base models, paper-released public weights, config bundles. \textbf{Training outputs are NOT your responsibility}: files under \texttt{checkpoints/} (or similar) that this paper's own training scripts produce are regenerated when a fresh clone runs the method. Do not attempt to fetch agent-trained checkpoints --- they don't exist on a public host. Skip \texttt{checkpoints/} entirely when auditing unless the paper explicitly ships a pretrained checkpoint to a public URL.

  Patterns that are BROKEN and you must fix:

  \begin{itemize}
  \tightlist
  \item
    A comment describing the fetch instead of running it: \texttt{\#\ curl\ https://\ldots{}\ dataset.tar.gz}, \texttt{\#\ \ \ wget\ -nc\ https://example.com/dataset.tar.gz}, \texttt{\#\ huggingface-cli\ download\ Qwen/Qwen3-1.7B\ \ldots{}}. A \texttt{\#} at the start of the line means the shell never runs it. This is unacceptable, since the download script in the future will run automatically. No comment will be accepted. If detected, you need to fix it.
  \item
    Placeholder URLs: \texttt{https://example.com/\ldots{}}, \texttt{https://TODO}, \texttt{https://placeholder}, or any domain that is obviously a stub. Replace with the real public URL for this paper's actual dataset/model.
  \item
    "Already present" skips: \texttt{echo\ "dataset\ already\ in\ shared/..."}, \texttt{if\ {[}\ -d\ shared/...\ {]};\ then\ echo\ present;\ fi}, or any branch that assumes a file is pre-placed on this specific machine and omits the fetch. A fresh clone on a new node has no \texttt{shared/} and no pre-placed files --- write the fetch command so it works from empty.
  \item
    Missing fetches entirely for a needed path. If \texttt{method/} loads \texttt{data/downloads/foo} but \texttt{download.sh} doesn't mention \texttt{foo}, add it.
  \end{itemize}

  IN SUCH CASE, YOU MUST REPLACE IT WITH A REAL, EXECUTABLE COMMAND.

  It is fine to wrap a fetch in \texttt{if\ {[}\ !\ -f\ \textless{}path\textgreater{}\ {]};\ then\ \textless{}fetch\textgreater{};\ fi} so running \texttt{download.sh} twice doesn't re-download what already exists --- as long as the fetch inside is a real command, not a comment.
\item
  \textbf{\texttt{scripts/download.sh} fetches only from public sources.} No \texttt{cp\ /home/user/shared/\ldots{}}, no local mount paths, no \texttt{shared/} references of any kind in this file. Allowed tools: \texttt{curl}, \texttt{wget}, \texttt{git\ clone}, \texttt{huggingface-cli\ download}, \texttt{aria2c}, \texttt{kaggle}, \texttt{gsutil}, \texttt{aws\ s3}. If a prior agent wrote \texttt{cp\ shared/\ldots{}} or \texttt{huggingface-cli\ download\ \ldots{}\ -\/-local-dir\ shared/\ldots{}}, replace it: fetch from the public origin and place the file under \texttt{data/downloads/} (or the path the method reads from).
\item
  \textbf{URLs resolve.} Spot-check with \texttt{curl\ -I\ \textless{}url\textgreater{}} --- do not download gigabytes. For HF models, a small \texttt{huggingface-cli\ download\ \textless{}repo\textgreater{}\ -\/-include\ config.json} confirms the repo exists. If a URL is 404, a domain is unreachable, or a resource was removed, find the correct public source and update.
\item
  \textbf{No \texttt{shared/} references in committed code (outside download.sh).} Search \texttt{method/}, \texttt{eval/}, \texttt{baseline/}, \texttt{scripts/}, configs for \texttt{shared/(datasets\textbar{}models\textbar{}hf\_cache)}. Replace any match with the corresponding \texttt{data/downloads/\textless{}name\textgreater{}} path. (\texttt{validate.py} also enforces this.)
\item
  \textbf{Paths line up.} If \texttt{download.sh} fetches to \texttt{data/downloads/foo/} but the method reads from \texttt{data/foo/} (or worse, \texttt{shared/datasets/foo/}), fix the mismatch --- prefer changing path strings in \texttt{method/} over contorting \texttt{download.sh} into a weird destination.
\item
  \textbf{Writable-paths rule for new files.}

  \begin{itemize}
  \tightlist
  \item
    \textbf{Public fetched data} (datasets, pretrained weights downloaded by \texttt{download.sh}) -\textgreater{} \texttt{data/downloads/\textless{}name\textgreater{}/}. Gitignored. Repopulated by \texttt{download.sh} on fresh clone.
  \item
    \textbf{First-party Python modules and small data files the code imports} -\textgreater{} directly under \texttt{data/} (e.g. \texttt{data/foo.py}, \texttt{data/instances.json}). \textbf{Tracked in git} so fresh clones can import them.
  \item
    \textbf{Never} use \texttt{/home/user/shared/...} --- it doesn't exist on a fresh clone and is banned by \texttt{validate.py}.
  \item
    If you find a blanket \texttt{data/} line in \texttt{.gitignore}, remove it. Only specific subdirectories like \texttt{data/downloads/} should be ignored, never the whole \texttt{data/} directory --- the paper's first-party modules live there.
  \end{itemize}
\end{enumerate}

\begin{center}\rule{0.5\linewidth}{0.5pt}\end{center}

\promptellipsis
\end{promptbox}

\begin{promptbox}{promptC7}{On Verify}
\psec{Final Verification}

You are the final step in a multi-agent workflow. The above prompt is given to previous agents (some still apply to you). Previous agents have reproduced a paper and iterated on the results. Your job is a \textbf{quality check} --- make sure the workspace is clean, correct, and ready to serve as a research gym for future AI scientist agents.

The previous agent reached milestone \texttt{\{previous\_milestone\}}. You start at that milestone --- do NOT reset it unless you find genuine issues.

Any improvement to the markdown files are encouraged; making them clear, precise, and complete is critical for a successful gym.

\begin{center}\rule{0.5\linewidth}{0.5pt}\end{center}

\promptellipsis

\psec{\texorpdfstring{\textbf{Method-Agnostic Files Are Critical}}{Method-Agnostic Files Are Critical}}

Future scientist agents must be able to understand the problem and evaluation without being told the paper's method.
If we leak the method, the environment is not useful.

Treat the following files as \textbf{strictly method-agnostic}:

\begin{itemize}
\tightlist
\item
  \texttt{briefing/problem.md}
\item
  \texttt{briefing/evaluation.md}
\item
  \texttt{scoring/EXPERIMENTS.md}
\end{itemize}

Treat the following files as \textbf{high-level and method-safe}:

\begin{itemize}
\tightlist
\item
  \texttt{scoring/TARGETS.md}
\item
  \texttt{scoring/CONSTRAINTS.md}
\item
  \texttt{scoring/DIRECTION.md}
\end{itemize}

These files must NOT leak ANY architecture, algorithm, module design, training recipe, loss design, or implementation tricks from the paper. They should describe:

\begin{itemize}
\tightlist
\item
  what problem is being solved
\item
  what each experiment is trying to establish
\item
  what must stay fixed
\item
  what research direction is allowed
\end{itemize}

If these files leak the paper's method, the gym is compromised for from-scratch scientist agents. Treat this as a real issue, not a cosmetic one. Fix if needed.

\begin{center}\rule{0.5\linewidth}{0.5pt}\end{center}

\psec{Verification Checklist}

Go through these in order. Fix issues you find, but \textbf{do not rewrite working code or waste compute re-running experiments that already succeeded.}

\psecsub{1. Folder Structure and correctness}

\begin{itemize}
\tightlist
\item
  \textbf{Clean root and self-contained \texttt{data/}} --- no stray scripts or data in the root, and a future agent can load the data without importing from \texttt{method/}.
\item
  \textbf{\texttt{eval/} is independent of \texttt{method/}} --- a from-scratch method must be able to use \texttt{eval/} and \texttt{scripts/evaluate.sh} without touching \texttt{method/}.
\item
  \textbf{\texttt{briefing/problem.md} and \texttt{evaluation.md} are method-agnostic} --- they state what to solve and how it is scored, never the paper's approach.
\item
  \textbf{\texttt{scripts/download.sh} is self-contained and \texttt{shared/}-free} --- a real, verified fetch for every external resource, runnable from an empty clone.
\end{itemize}

\promptellipsis

\psecsub{2. Scoring Accuracy}

\begin{quote}
\textbf{Reject any scored wall-clock-speed metric.} Hardware varies across runs (different GPU models), so identical code can clock 2--5x faster or slower with no algorithmic change. Any metric whose value depends on wall-clock time / throughput / runtime is unreliable here. The fix is a machine-independent substitute: FLOPs, parameter count, training-step count, or samples processed.
\end{quote}

\begin{itemize}
\tightlist
\item
  Does \texttt{scoring/reference.json} exist and pass \texttt{python\ validate.py}?
\item
  Do experiment names in \texttt{scoring/scores.json} match those in \texttt{scoring/reference.json}?
\item
  Does \texttt{scoring/EXPERIMENTS.md} match the experiment names in \texttt{scoring/scores.json}, and does it describe each experiment at a high level without leaking method details?
\item
  Are the \texttt{coefficient} values in \texttt{reference.json} reasonable? Positive for benefits, negative for constraints. Do the magnitudes reflect how much the paper emphasizes each metric relative to others (not absolute value, but think of if the scientist achieves a better score, is the coefficient multiple that relative different an appropriate scale)?
\item
  Are experiment \texttt{weight} values reasonable? Does the primary experiment get the most weight?
\item
  Are there constraint metrics where the paper has trade-offs? If the paper claims "X without sacrificing Y," is Y measured?
\item
  \textbf{\texttt{reference.json} values should match what the paper reports, OR carry a \texttt{notes} field explaining a deliberate change.} If a value in \texttt{reference.json} is noticeably lower / different from the paper's published number and there's no explanation (e.g., \texttt{"notes":\ "scaled\ to\ 10\%\ data;\ paper\ reports\ 0.97\ on\ full"}), that's suspicious --- the prior agent may have edited the reference downward to mask a broken reproduction rather than fix it. Spot-check by skimming the paper's results table for the headline numbers and comparing. Flag undocumented drift as a finding; ask the next iteration to either restore the paper's value (and re-run the method) or document the scaling choice properly.
\end{itemize}

\psecsub{3. Gym Scoring Review}

Think of yourself as a \textbf{judge designing the evaluation rubric} for this gym. Future AI scientist agents will compete on this workspace --- you are deciding how their work gets scored.

The gym works as follows: scientists are scored ONLY on experiments that appear in \texttt{scores.json}. The values in \texttt{scores.json} serve as the baseline. The weights from \texttt{reference.json} are renormalized over just those experiments. For example, two AI scientists each propose their own method; we compare the weighted relative changes in scores among them, of what they produced in \texttt{scores.json}.

Review with this in mind:

\begin{itemize}
\tightlist
\item
  \textbf{Are the renormalized weight proportions reasonable?} Scientists are scored only on the experiments in \texttt{scores.json}, with \texttt{reference.json} weights renormalized over them --- check that ratio reflects what matters for this paper.
\item
  \textbf{Are the coefficient scales appropriate?} A small, easy gain on one metric should not dominate a hard-won gain on the primary metric; scale by how easily each metric moves, not by its absolute value.
\item
  \textbf{CRITICAL: no scored metric may measure deviation from the paper's published values.} The paper is context, never an optimization target; a metric whose floor is ``exactly match the paper'' leaves the paper's own method at the ceiling and breaks the gym by construction.
\end{itemize}

\promptellipsis

\psecsub{4. Completeness}

\begin{itemize}
\tightlist
\item
  Are \texttt{scoring/TARGETS.md}, \texttt{scoring/EXPERIMENTS.md}, \texttt{scoring/CONSTRAINTS.md}, \texttt{scoring/DIRECTION.md} filled in and reasonable (not just template comments)? These documents guide future scientist agents --- vague or generic language wastes their context window. \texttt{EXPERIMENTS.md} must stay method-agnostic and correspond exactly to the experiment structure in \texttt{scores.json}. If you see room for improvement, consider refining them.
\item
  Does \texttt{scripts/evaluate.sh} work standalone?
\item
  Does \texttt{scripts/reproduce.sh} exist and describe the end-to-end flow?
\item
  Is \texttt{environment/container.def} + \texttt{environment/setup.sh} reasonable?
\item
  Is \texttt{MILESTONES.md} filled in with an accurate progress log and stop justification?
\end{itemize}

\promptellipsis

\psecsub{6. Evaluation Difficulty}

Agents often scale down or substitute datasets to work around resource constraints. This is acceptable when necessary, but it can silently make the problem trivial --- destroying the evaluation's ability to distinguish methods. A workspace where every approach achieves near-perfect scores is useless as a benchmark, regardless of how well the code is written. An evaluation reduced from the scale of thousands to tens usually could not show its judgement.

\textbf{Critical: Experiment procedure must match the paper.} Scoping down the model size, dataset size, or epoch count is acceptable. Changing the training method is not. If the paper trains from scratch and the agent fine-tunes from pretrained weights (or vice versa), or if the agent swaps the optimization regime (supervised vs self-supervised, RLHF vs SFT, etc.), the experiment cannot demonstrate the paper's claim --- it produces numbers that look reasonable but are structurally invalid. Read \texttt{briefing/method.md} and the agent's \texttt{scripts/reproduce.sh} / training code; if the training protocol differs from the paper, flag it and downgrade the milestone accordingly.

\begin{itemize}
\tightlist
\item
  \textbf{Compare reproduced scores against the paper's numbers.} If the reproduction scores are dramatically higher than the paper's, the agent likely used a simplified setup that does not preserve the original difficulty. This is a critical issue.
\item
  \textbf{Check whether the agent used the paper's original data or a substitute.} If a substitute, verify it preserves the problem's difficulty. Look at baseline performance --- if a simple baseline with no special method already achieves near-perfect results, the substitute is inadequate.
\item
  \textbf{Check ablation results for sanity.} If removing the paper's key module has no effect on scores, either the ablation is broken or the evaluation is too easy to detect the difference.
\item
  Do the numbers in \texttt{scores.json} look plausible for this domain? \textbf{If any reproduced score is more than 10x off from the corresponding value in \texttt{reference.json}, flag it as a likely implementation bug --- not just an imperfect reproduction.}
\end{itemize}

\begin{center}\rule{0.5\linewidth}{0.5pt}\end{center}

\psec{What to Do}

\begin{itemize}
\tightlist
\item
  \textbf{If everything looks good}: update \texttt{MILESTONES.md} if needed, commit, and signal done. Do not make changes for the sake of making changes.
\item
  \textbf{If you find issues}: fix them. Prioritize scoring accuracy and folder structure --- these directly affect whether the gym is usable. You may submit a GPU job if a quick re-evaluation is needed, but avoid long training runs.
\item
  \textbf{If the reproduction is completely broken} --- the code is broken, the method is fundamentally wrong, results are fabricated, or the workspace is unrecoverable to be used as a gym --- set the milestone to \texttt{not\_possible} and explain why in \texttt{VERIFICATION.md}. This tells the system the paper needs a full redo.
\item
  \textbf{Do not}: refactor working code unnecessarily, rename files for style, add features that weren't asked for, or re-run experiments that already have valid results.
\end{itemize}

\begin{center}\rule{0.5\linewidth}{0.5pt}\end{center}

\promptellipsis
\end{promptbox}

\begin{promptbox}{promptC8}{On Job Completion}
\psec{Job Completed}

A compute job has finished.

\begin{tabularx}{\linewidth}{@{}lX@{}}
\toprule
Field & Value \\
\midrule

\textbf{Job ID} & \texttt{\{run\_id\}} \\
\textbf{Exit code} & \texttt{\{returncode\}} \\
\textbf{stdout} & \texttt{\{stdout\_path\}} \\
\textbf{stderr} & \texttt{\{stderr\_path\}} \\
\bottomrule
\end{tabularx}

Read the log files to review the job's output.

\begin{center}\rule{0.5\linewidth}{0.5pt}\end{center}

\psec{What to Do}

\textbf{You are fully autonomous. There is no human in the loop. Do not ask for confirmation, do not wait for direction, do not say "let me know" or "please confirm". Decide and act immediately --- directly edit code, fix bugs, and create files in this turn.}

\begin{enumerate}
\def\labelenumi{\arabic{enumi}.}
\tightlist
\item
  \textbf{Review the logs.} Understand what happened --- success, failure, or partial results.
\item
  \textbf{Assess progress.} How does this move you toward your reproduction goals?
\item
  \textbf{Edit code in this turn.} Based on the logs, directly edit scripts, fix bugs, update configs, create new files --- whatever is needed. Most turns should make a concrete change that advances the reproduction; spend a turn without editing only when you genuinely need to inspect or plan, not to show motion.
\item
  \textbf{Submit the next job} with your updated code, or signal done if finished.
\item
  \textbf{Update your milestone} if progress was made.
\end{enumerate}

End your response with the structured YAML block as described in your instructions.
\end{promptbox}

%% file: prompts/sec_eval.tex
% Generated by generate_prompts.py (coarse pass) then HAND-CURATED.
% Re-running the generator reverts manual trims.
\subsection{Prompts for Evaluation}
\label{app:prompts-eval}

\begin{promptbox}{promptC1}{Protocol}
\psec{Action Protocol}

You are fully autonomous --- there is no human in the loop. Before your
turn ends, edit \texttt{action.yaml} at the workspace root with one of the
blocks below, then stop.

\textbf{To submit a compute job:}

\begin{lstlisting}[style=promptcode]
action: submit
gpus: 1
test: false
notes: "what this job does"
\end{lstlisting}

The pipeline will submit the job, wait for it to finish, and invoke
you again with the logs.

\begin{itemize}
\item
  \texttt{gpus}: number of GPUs (set to \texttt{0} for a CPU-only job).
\item
  \texttt{test}: set to \texttt{true} for quick validation jobs on test nodes
  (max 1 hour wall-clock), \texttt{false} for normal jobs.

  \begin{itemize}
  \tightlist
  \item
    Use \texttt{test:\ true} to smoke-test your setup (does the container
    work? does training start? does eval run?) before committing to a
    long job.
  \item
    Test nodes have a \textbf{1-hour time limit} --- only use them for quick
    checks, never full training runs.
  \item
    Works for both GPU and CPU jobs: \texttt{gpus:\ 1,\ test:\ true} -\textgreater{} GPU test
    node; \texttt{gpus:\ 0,\ test:\ true} -\textgreater{} CPU test node.
  \end{itemize}
\item
  Anything beyond a few minutes of compute must use \texttt{action:\ submit}.
  Inline bash counts against your turn's 4-hour wall-clock cap and
  will be killed mid-job.
\end{itemize}

\textbf{To finish the run:}

\begin{lstlisting}[style=promptcode]
action: done
notes: "why you are done"
\end{lstlisting}

\texttt{action:\ done} is PERMANENT --- no follow-up session, no successor
agent, no next turn. Do not mark \texttt{done} because of transient
infrastructure problems (container build failures, Nexus/Hive errors,
GPU job timeouts); those are fixable. Iterate on the environment,
resubmit, or describe the issue in your next \texttt{submit} notes, but stay
in the workflow.
\end{promptbox}